\documentclass[journal]{IEEEtran}
\usepackage{amsmath}
\usepackage{amsmath,amssymb}
\usepackage{stmaryrd}
\usepackage{graphicx,graphics,color,psfrag}
\usepackage{cite,balance}
 \usepackage{subfigure}
 \usepackage{booktabs}
\usepackage{caption}   % set the caption size
\allowdisplaybreaks
\usepackage{algorithm}
\usepackage{bm}
\usepackage{eufrak}
\usepackage{url}
\usepackage[english]{babel}
\usepackage{algpseudocode}
\usepackage{multirow}
\usepackage{enumerate}
\usepackage{cases}

\newtheorem{property}{Property}
\begin{document}
 \pdfminorversion=4
\graphicspath{{figure/}}
\title{Towards Flexible Sparsity-Aware Modeling: Automatic Tensor Rank Learning Using \\ The  Generalized Hyperbolic Prior}
\author{Lei Cheng, Zhongtao Chen, Qingjiang Shi, \\ Yik-Chung Wu, {\it Senior Member, IEEE}, and Sergios Theodoridis, {\it Life Fellow, IEEE} 

\thanks{Lei Cheng is with the College of Information Science and Electronic Engineering, Zhejiang University, Hangzhou, 310027, China, and also with  Zhejiang Provincial Key Laboratory of Info. Proc., Commun. \& Netw. (IPCAN), Hangzhou 310027, China (e-mail: lei\_cheng@zju.edu.cn).}   \thanks{Zhongtao Chen and Yik-Chung Wu are with the Department of Electrical and Electronic Engineering, The University of Hong Kong, Hong Kong (e-mail: ztchen@connect.hku.hk, ycwu@eee.hku.hk).} 
\thanks{Qingjiang Shi is with the School of Software Engineering at Tongji University,
Shanghai 201804, China. He is also with the Shenzhen Research Institute
of Big Data, Shenzhen 518172, China (e-mail: shiqj@tongji.edu.cn).}
\thanks{Sergios Theodoridis is with the National and Kapodistrian University of Athens, Greece and with the Department of Electronic Systems, Aalborg University, Denmark (email: stheodor@di.uoa.gr).}}

\maketitle

\begin{abstract}
Tensor rank learning for canonical polyadic decomposition (CPD) has long been deemed as an essential yet challenging problem. In particular,  since the tensor rank controls the complexity of the CPD model, its inaccurate learning would cause overfitting to noise or underfitting to the signal sources, and even destroy the interpretability of model parameters. However, the optimal determination of a  tensor rank is known to be a non-deterministic polynomial-time hard (NP-hard) task. Rather than exhaustively searching for the best tensor rank via trial-and-error experiments,  Bayesian inference under the Gaussian-gamma prior was introduced in the context of  probabilistic CPD modeling, and it was shown to be an effective strategy for automatic tensor rank determination. This triggered flourishing research on other structured tensor CPDs with automatic tensor rank learning.  On the other side of the coin, these research works also reveal that the Gaussian-gamma model does not perform well for  high-rank tensors and/or low signal-to-noise ratios (SNRs). To overcome these drawbacks, in this paper, we introduce a more advanced generalized hyperbolic (GH) prior to the probabilistic CPD model, which not only includes the Gaussian-gamma model as  a special case, but  also is more flexible  to adapt to different levels of sparsity. Based on this novel probabilistic model, an algorithm is developed under the framework of variational inference, where each update is obtained in  a closed-form. Extensive numerical results, using synthetic data and real-world datasets, demonstrate the significantly improved performance of the proposed method in learning both low as well as high tensor ranks even for low SNR cases.
\end{abstract}


\section{Introduction}

In the Big Data era, tensor decomposition has become one of the most important tools in both theoretical studies of machine learning  \cite{theory1, theory2} and a variety of real-world applications \cite{nature1, nw1,  bmc1, app1, app2, app3, app4, app5}. Among all the tensor decompositions, canonical polyadic decomposition (CPD) is the most fundamental format. {\color{black} It not only provides a faithful representation of a lot of real-world multidimensional data, see, e.g., [14]-[17],} but it also  allows a unique factor matrix recovery up to trivial scaling and permutation ambiguities \cite{tensor1}. This uniqueness bolsters the uncovering of the interpretable knowledge from the tensor data, which endows tensor  CPD an advantage  over other learning approaches (e.g., deep models),  in various data analytic tasks including  social group mining \cite{papa1}, drug discovery  \cite{J1}, biomedical data analytics \cite{J3} and functional Magnetic Resonance Imaging (fMRI) \cite{Sergios}.

\begin{figure}[!t]
\centering
\includegraphics[width= 3.2 in]{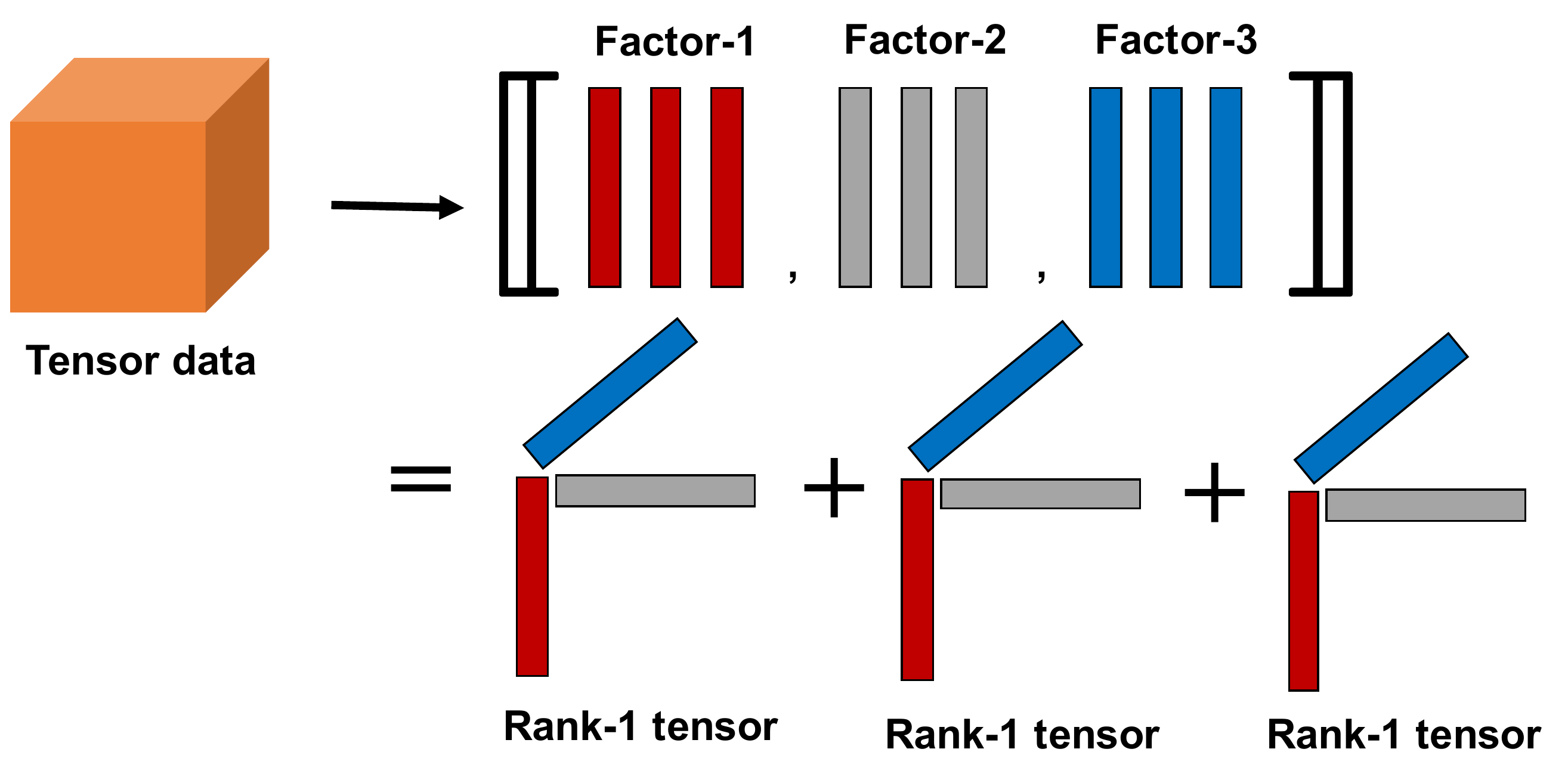}
\caption{Illustration of tensor CPD.}
\label{fig_topology}
\end{figure}

In tensor CPD, given an $N$ dimensional (N-D) data tensor $\mathcal Y \in \mathbb R^{J_1 \times \cdots \times J_N}$, a set of factor matrices $\{\boldsymbol U^{(n)} \in \mathbb R^{J_n \times R} \}$ are sought via solving the following problem \cite{tensor1}:
\begin{align}
&\min_{ \{ \boldsymbol U ^{(n)}\}_{n=1}^N}  \parallel \mathcal Y  - \underbrace{\sum_{r=1}^R   \boldsymbol U^{(1)}_{:,r} \circ  \boldsymbol U^{(2)}_{:,r} \circ \cdots \circ  \boldsymbol U^{(N)}_{:,r}}_{\triangleq \llbracket  \boldsymbol U^{(1)}, \boldsymbol U^{(2)}, \cdots,  \boldsymbol U^{(N)}   \rrbracket } \parallel_F^2,  \label{eq1}
\end{align}
where the symbol $\circ$ denotes vector outer product and the shorthand notation $\llbracket  \cdots \rrbracket$ is termed as the Kruskal operator. As illustrated in Figure 1, the tensor CPD aims at decomposing {\color{black} an N-D tensor} into a summation of $R$ rank-1 tensors, with the $r^{th}$ component constructed as the vector outer product of the $r^{th}$ columns from all the factor matrices,  i.e.,  $\{ \boldsymbol U^{(n)}_{:,r} \}_{n=1}^N$.  In problem \eqref{eq1}, the number of columns $R$ of each factor matrix, also known as tensor rank \cite{tensor1}, determines the number of unknown model parameters and equivalently the model complexity. In practice, it needs to be carefully selected to achieve the best performance in both recovering the noise-free signals (e.g., image denoising \cite{denoising1}) and unveiling the underlying components (e.g., social group clustering \cite{papa1}).

If the value of the tensor rank is known, problem \eqref{eq1} can be solved via nonlinear programming methods \cite{NP}. In particular,  it has been found that  problem \eqref{eq1} enjoys a nice block multi-convexity property, in the sense that after fixing all but one 
factor matrix, the problem is convex with respect to that matrix. This property motivates the use of block coordinate descent (BCD) methods (or {\color{black} alternating optimization methods}) to devise fast and accurate algorithms for tensor CPD and its structured variants \cite{bcd1, bcd2, bcd3}.  From a nonlinear programming perspective,  these solutions need the knowledge of the tensor rank $R$, which, however,  is unknown and, in general, it is  non-deterministic polynomial-time hard (NP-hard) to obtain \cite{tensor1}. To acquire the optimal tensor rank (or equivalently the optimal model complexity), trial-and-error parameter tuning has been employed in previous works  \cite{denoising1, denoising2, papa1, J1, J3}, which,  unfortuantely, is computationally costly.

To tackle the challenge of automatic tensor rank learning, a novel heuristic test method based on the core consistency diagnostic was proposed in \cite{C1,C2} and later efficiently implemented in \cite{C3}, which, however, still needs to compute CPD multiple times. Recently, relying on a regularization scheme, model selection for block-term tensor decomposition, which includes CPD as its special case, was investigated in \cite{C4}, in which hyper-parameter tuning is inevitable. 

A breakthrough to the challenge of automatic tensor rank learning has been achieved under the framework of Bayesian modeling and inference. In the early work \cite{C5}, sparsity-promoting priors (including Gaussian-gamma prior  and Laplacian prior) were adopted, based on which the Tucker/CPD model parameters were estimated via the maximum-a-posterior (MAP) approach (also called ``poor-man'' Bayesian inference). Fully Bayesian treatments for probabilistic tensor CPD have been achieved in \cite{C6, new1} using sampling schemes, and the variational inference (VI) framework \cite{PI1, C7, C8}, which is more scalable to massive data \cite{beal,VI1,VI2,  Sergios2}. 

Unlike \cite{C7, C8} which focus on count data, continuous tensor data was considered in [22] and the follow-up works \cite{PI2, CL2, CL3,CL5, C9, C10,  C12}. These two data types result in different likelihood functions (or cost functions) and prior distribution design principles. In particular, the likelihood functions in  \cite{C7, C8} are poisson  and  binomial distributions, respectively, while \cite{PI1, PI2, CL2, CL3, CL5, C9, C10,  C12} adopt Gaussian distributions (and its variants) as likelihood functions. 

Since the inception of  \cite{PI1}, the development of VI-based structured continuous-valued tensor decompositions with automatic tensor rank learning  \cite{PI2, CL2, CL3,CL5, C9, C10,  C12} has been flourishing. Among these works, the key idea lies in the adoption of sparsity-promoting Gaussian-gamma prior and its variants for modeling the powers of columns in all the factor matrices, so that most columns in the factor matrices will be driven to zero during inference. Then, the number of remaining non-zero columns in each factor matrix is used to estimate the tensor rank.  Extensive numerical studies using both synthetic data and real-world datasets have demonstrated the effectiveness of these methods \cite{PI1, PI2, CL2, CL3, CL5, C9, C10,  C12}. 

In addition to learning the tensor rank of basic CPD, the Gaussian-gamma prior has recently become an important part in the probabilistic modeling of several advanced tensor models. For example, it was utilized as the cornerstone in  \cite{new2} to infer the tensor tubal rank and multi-rank of tensor SVD,  and cleverly integrated with the spike-and-slab prior in  \cite{new3} to achieve simultaneous view-wise and feature-wise sparsity learning for multi-view tensor factorization, in which multiple tensors are  jointly  decomposed with shared latent factors.

%A breakthrough to the above problem has been achieved under the framework of Bayesian modeling and inference.  Notably,  since the seminal paper \cite{PI1}, the development of  structured large-scale tensor decompositions with automatic tensor rank learning \cite{PI2, CL2, CL3, CL4, CL5} has been flourishing. The game-changing idea is the adoption of sparsity-promotingGaussian-gamma prior and its variants for modeling the powers of columns in all the factor matrices, so that most columns in the factor matrices will be driven to zero during inference. Then, the number of remaining non-zero columns in each factor matrix gives the estimate of the tensor rank. Extensive numerical studies using both synthetic data and real-world datasets have demonstrated the effectiveness of these methods \cite{PI1, PI2, CL2, CL3, CL4, CL5}. 

\begin{figure}[!t]
\centering
\includegraphics[width= 3.5 in]{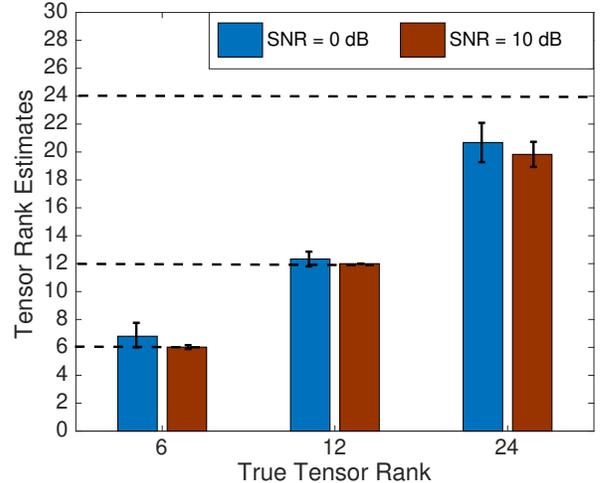}
\caption{Tensor rank learning results  from probabilistic tensor CPD with Gaussian-gamma prior \cite{PI1}. The vertical bars show the mean and the error bars indicate the standard derivation of tensor rank estimates. The black dashed lines show the true tensor rank.}
\label{fig_topology}
\end{figure}

While it may seem that tensor rank learning in CPD is solved, a closer inspection on the numerical results reveals that the performance of tensor rank learning deteriorates significantly when the noise power is large, and/or the true tensor rank is close to the dimension of the tensor data.  As an illustration, consider three dimensional (3D) signal tensors\footnote{\color{black}Each signal tensor is generated via $\mathcal X =  \llbracket  \boldsymbol U^{(1)},   \boldsymbol U^{(2)},  \boldsymbol U^{(3)} \rrbracket  \in \mathbb R^{30 \times 30 \times 30}$, where each element in the factor matrices $\{ \boldsymbol U^{(n)}\}_{n=1}^3$ is independently drawn from $\mathcal N(0,1)$.} with dimension $30 \times 30 \times 30$  and the tensor ranks {\color{black} belonging to} $\{6, 12, 24\}$. The observation tensor data are obtained by corrupting the signal tensors using additive white gaussian noises (AWGNs), with the noise power  characterized by  signal-to-noise ratio (SNR). With the tensor rank upper bound being  $30$,  the probabilistic tensor CPD algorithm \cite{PI1} was run on  the observation tensor data. The tensor rank learning results from $100$ Monte-Carlo trials are shown in Figure 2. It is clear that when the SNR is $0$ dB, the probabilistic tensor CPD algorithm with Gaussian-gamma model \cite{PI1} either over-estimates or under-estimates the tensor rank. At a high SNR (i.e., 10 dB), although the method \cite{PI1} estimates the correct rank value for low tensor ranks $\{6, 12\}$, it fails to recover the high tensor rank $24$. In practice, over-estimation of a tensor rank will either lead to overfitting of the noise components or  to generating   uninterpretable ``ghost'' components. On the other hand, if the tensor rank is under-estimated, it is prone to missing important signal components. Therefore, there is a need for further improving the accuracy of automatic tensor rank learning in CPD.

To achieve this goal via a principled approach, it is worthwhile to trace back the development of the Gaussian-gamma prior and the Bayesian framework from the early work of Mackay \cite{Mackay} and Tipping \cite{Tipping} on Bayesian neural networks and relevance vector machines (RVMs), in which important variables are automatically identified in the associated models. This hints us that further inspiration can be drawn from the early works \cite{Mackay, Tipping} and beyond \cite{AD1, AD2, AD3, AD4, AD5}. In particular, to adapt to different levels of sparsity, which is measured by the ratio of zero-valued parameters to the total number of parameters,   advanced sparsity-promoting priors including generalized-t distribution \cite{AD1}, normal-exponential gamma distribution \cite{AD2}, horseshoe distribution \cite{AD3} and generalized hyperbolic distribution \cite{AD4, AD5} could be employed. Since these advanced priors are much more flexible in their functional forms compared to the Gaussian-gamma prior (and some of them even include the Gaussian-gamma model as their special cases), improved performance of variable selection was witnessed in linear regression models. We conjecture that this group of advanced sparsity-promoting priors would improve CPD rank selection compared to the Gaussian-gamma model.

In this paper, we take {\it the first step} towards developing  {\it``Flexible Sparsity-Aware Modeling"} by introducing the generalized hyperbolic (GH) prior \cite{AD4} into the research of probabilistic tensor CPD. The reason for choosing this prior is that not only it includes the widely-used Gaussian-gamma prior and Laplacian prior as its special cases, but also its mathematical form allows for an efficient expectation computation.
Furthermore, the GH prior can be interpreted as a Gaussian scale mixture where the mixing distribution is the generalized inverse Gaussian (GIG) distribution \cite{GIG}. This interpretation allows for a hierarchical construction of the probabilistic model with the conjugacy property within the exponential distribution family, based on which efficient variational inference (VI) algorithms \cite{beal, VI1, VI2, Sergios2} can be devised with closed-form update expressions.

By making full use of these advantages, we design a novel probabilistic tensor CPD model and the corresponding inference algorithm. {\color{black} Since the GH prior provides a more flexible sparsity-aware modeling than the Gaussian-gamma prior, and all the latent variables are updated during the learning process, the GH  prior in essence has the potential to act as a better regularizer against the noise corruption, and to adapt to a wider range of sparsity levels.} Numerical studies using both synthetic data and real-world datasets have demonstrated the improved performance of the proposed method over the Gaussian-gamma CPD in terms of tensor rank learning and factor matrix recovery, especially in the challenging high-rank and/or low-SNR regimes.  

Note that the principle followed in this paper is a parametric way to seek flexible sparsity-aware modeling in the context of tensor CPD (specifically, investigating the advanced prior in the Gaussian scale mixture family). In parallel to this path, the pioneering work \cite{new1} proposed a non-parametric Bayesian CPD modeling based on a multiplicative gamma process (MGP) prior. Employing a Gibbs sampling method, the inference algorithm of \cite{new1}  can deal with both continuous and binary tensor data. Due to the decaying effects of the length scales learnt through MGP \cite{new1}, the inference algorithm is capable of learning low tensor rank, but it has the tendency to under-estimate the tensor rank when the ground-truth rank is  high (as validated in Section V), making it not very flexible in the high-rank regime.

{\color{black} The major contributions of this paper are summarized as follows:
\begin{itemize}
\item  {\bf Improved tensor rank learning capability:} This paper makes {\it the first} attempt to tackle the difficulty of tensor rank learning in high-rank and/or low-SNR regimes, under which previous works (especially those relying on Gaussian-gamma prior)  {\it do not} achieve  satisfactory performance. Therefore, this paper represents {\it the first step} to show that tensor rank learning performance can be {\it further improved} by employing a more advanced sparsity-promoting prior in the Gaussian scale mixture family.

\item {\bf Principled new prior design:} This paper exemplifies a principled design of a new parametric prior for tensor CPD, which can benefit  future research of other Bayesian tensor decomposition formats. In particular, this paper clarifies why the GH prior should be chosen, and how this prior, which was originally devised for modeling scalars or vectors, can be adopted to model the rank-1 components of a tensor, thereby providing enhanced tensor rank learning capability.  

\item {\bf Non-trivial inference algorithm derivation:}  Given a new prior, the derivation of inference algorithm under VI framework is non-trivial, due to the complicated tensor algebra involved and the newly introduced nonlinearities from the GH prior. The derivations  could be used as the reference or  basis results  for future related VI algorithm derivations. 

\end{itemize}}

The remainder of this paper is organized as follows. In Section II, the probabilistic tensor CPD using Gaussian-gamma prior is briefly reviewed. By leveraging the GH prior, we propose a new probabilistic model for tensor CPD in Section III. In Section IV, the framework of variational inference is utilized to derive an inference algorithm with closed-form update equations. In Section V and VI, numerical results are presented to demonstrate the superior performance of the proposed method. Finally, conclusions and future directions are presented in Section VII.

\textbf{Notation}: Boldface lowercase and uppercase letters will be used for vectors and matrices, respectively.
Tensors are written as calligraphic letters. $\mathbb E [~\cdot~] $ denotes the expectation of its argument. Superscript $T$ denotes transpose, and the operator $\textrm{Tr}\left( {\boldsymbol{A}} \right)$ denotes the trace of matrix $\boldsymbol{A}$. $\parallel \cdot \parallel_F$ represents the Frobenius norm of the argument.  $\mathcal {N} (\boldsymbol x | \boldsymbol u, \boldsymbol R)$ stands for the probability density function (pdf) of a  Gaussian vector $\boldsymbol x$ with mean $\boldsymbol u$ and covariance matrix $\boldsymbol R$. The $N \times N $ diagonal matrix with diagonal elements $a_1$ through $a_N$ is represented as $\mathrm{diag} \{a_1, a_2,...,a_N\}$, while $\boldsymbol I_{M}$ represents the $M \times M$ identity matrix. The $(i,j)^{th} $ element, the $i^{th}$ row, and the $j^{th}$ column of a matrix $\boldsymbol A$ are represented by   $\boldsymbol{A}_{i,j}$,  $\boldsymbol{A}_{i,:}$ and $\boldsymbol{A}_{:,j}$, respectively. For easy reference, a table of symbol notations is also given in  Appendix A.

%\textbf{Notation}: Boldface lowercase and uppercase letters will be used for vectors and matrices, respectively.
%Tensors are written as calligraphic letters. $\mathbb E [~\cdot~] $ denotes the expectation of its argument. Superscript $T$ denotes transpose, and the operator $\textrm{Tr}\left( {\boldsymbol{A}} \right)$ denotes the trace of matrix $\boldsymbol{A}$. $\parallel \cdot \parallel_F$ represents the Frobenius norm of the argument.  $\mathcal {N} (\boldsymbol x | \boldsymbol u, \boldsymbol R)$ stands for the probability density function (pdf) of a  Gaussian vector $\boldsymbol x$ with mean $\boldsymbol u$ and covariance matrix $\boldsymbol R$. The $N \times N $ diagonal matrix with diagonal elements $a_1$ through $a_N$ is represented as $\mathrm{diag} \{a_1, a_2,...,a_N\}$, while $\boldsymbol I_{M}$ represents the $M \times M$ identity matrix. The $(i,j)^{th} $ element, the $i^{th}$ row, and the $j^{th}$ column of a matrix $\boldsymbol A$ are represented by   $\boldsymbol{A}_{i,j}$,  $\boldsymbol{A}_{i,:}$ and $\boldsymbol{A}_{:,j}$, respectively. For easy reference, a table of symbol notations is also given in  Appendix A.

\section{Review of The Gaussian-gamma model for CPD}
In tensor CPD, as illustrated in Figure 1,  the $l^{th}$ columns in all the factor matrices ($\{ \boldsymbol U^{(n)}_{:,l} \}_{n=1}^N$) constitute the building block of the model. Given an upper bound value $L$ of the tensor rank, for each factor matrix, since there are $L - R$ columns being all zero,  sparsity-promoting priors should be imposed on the columns of each factor matrix to encode the information of over-parameterization\footnote{\color{black} The sparsity  level of over-parameterized CPD model can be measured by $\frac{L-R}{L} $.}. In the pioneering works \cite{PI1, C5}, assuming statistical independence among the columns in $\{ \boldsymbol U^{(n)}_{:,l}, \forall n, l\}$, a Gaussian-gamma prior was utilized to model them as 
\begin{align}
& p ( \{ \boldsymbol U^{(n)} \}_{n=1}^N | \{\gamma_l\}_{l=1}^L) =  \prod_{l=1}^L  p( \{ \boldsymbol U^{(n)}_{:,l} \}_{n=1}^N | \gamma_l )  \nonumber \\
& = \prod_{l=1}^L \prod_{n=1}^N  \mathcal N( \boldsymbol U^{(n)}_{:,l} | \boldsymbol 0_{J_n \times 1}, \gamma_l^{-1} \boldsymbol I_{J_n}), \label{eq2} \\
& p(\{\gamma_l\}_{l=1}^L | \{ c_l^0, d_l^0 \}_{l=1}^L) = \prod_{l=1}^L p(\gamma_l | c_l^0, d_l^0) \nonumber\\
& = \prod_{l=1}^L  \mathrm{gamma}(\gamma_l | c_l^0, d_l^0),   \label{eq3}
\end{align}
where $\gamma_l$ is the precision (i.e., the inverse of variance) of the $l^{th}$ columns $\{ \boldsymbol U^{(n)}_{:,l}\}_{n=1}^N$, and $\{c_l^0, d_l^0\}$ are pre-determined hyper-parameters.

\begin{figure}[!t]
\centering
\includegraphics[width= 3.5 in]{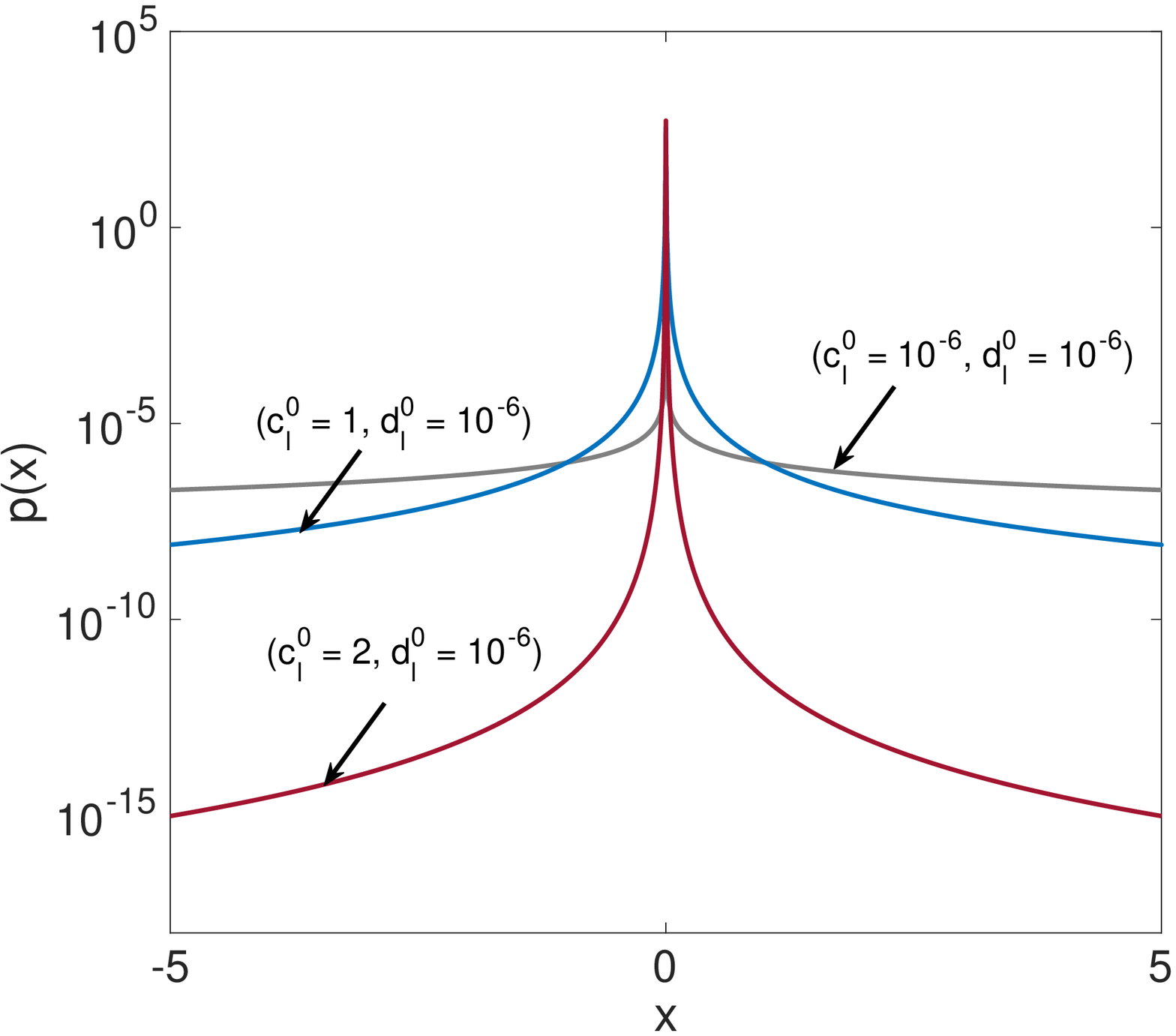}
\caption{Univariate marginal probability density function in  \eqref{eq4} with different values of hyper-parameters.}
\label{fig_topology}
\end{figure}

To see the sparsity-promoting property of the  above Gaussian-gamma prior, we marginalize the precisions  $\{\gamma_l\}_{l=1}^L$ to obtain the marginal probability density function (pdf) $p(\{ \boldsymbol U^{(n)} \}_{n=1}^N)$ as follows:
\begin{align}
& p(\{ \boldsymbol U^{(n)} \}_{n=1}^N) = \prod_{l=1}^L p(\{ \boldsymbol U^{(n)}_{:,l} \}_{n=1}^N) \nonumber \\
& =  \prod_{l=1}^L \int p( \{\boldsymbol U^{(n)}_{:,l}\}_{n=1}^N | \gamma_l ) p(\gamma_l | c_l^0, d_l^0) d \gamma_l \nonumber \\
&=  \prod_{l=1}^L   (\frac{1}{\pi})^{\frac{ \sum_{n=1}^N J_n}{2}} \frac{\Gamma(c_l^0 +  \sum_{n=1}^N \frac{J_n}{2} )}{ {2d_l^0}^{-c_l^0} \Gamma(c_l^0)} \nonumber \\
&\! \times \! \left(2d_l^0 + \bigparallel \mathrm{vec}\left(  \{\boldsymbol U^{(n)}_{:,l}\}_{n=1}^N \right) {\bigparallel}_2^2 \right)^{-c_l^0 -  \sum_{n=1}^N \frac{J_n}{2} }, \label{eq4}
\end{align}
where $\Gamma(\cdot)$ denotes the gamma function and $\mathrm{vec}(\cdot)$ denotes the vectorization\footnote{\color{black} The operation $\mathrm{vec}\left(  \{\boldsymbol U^{(n)}_{:,l}\}_{n=1}^N \right)$ simply stacks all these columns into a long vector, i.e., $\mathrm{vec}\left(  \{\boldsymbol U^{(n)}_{:,l}\}_{n=1}^N \right) = [\boldsymbol U^{(1)}_{:,l}; \boldsymbol U^{(2)}_{:,l}; \cdots; \boldsymbol U^{(N)}_{:,l} ] \in \mathbb R^{Z \times 1}$, with $Z = \sum_{n=1}^N J_n$.} of its argument. Equation \eqref{eq4} characterizes a multivariate student's t distribution with hyper-parameters $\{c_l^0, d_l^0\}_{l=1}^L$. To get insights from this marginal distribution, we illustrate its univariate case in Figure 3 with different  hyper-parameters. It is clear that each student's t pdf  is strongly peaked at zero and with heavy tails. The prior with such features is known as  sparsity-promoting prior,  since the peak at zeros will inform the learning process to look for values around ``zeros'' while the heavy tails still allow the learning process to obtain components with large values \cite{Sergios2}.

The probabilistic CPD model is completed by specifying the likelihood function of $\mathcal Y$:
%
%The least-squares (LS) cost function in (1) suggests the following generative model \cite{PI1}:
%\begin{align}
%\mathcal Y = \llbracket  \boldsymbol U^{(1)}, \boldsymbol U^{(2)}, \cdots,  \boldsymbol U^{(N)}   \rrbracket  + \mathcal W,
%\end{align} 
%where $\mathcal W$ is AWGN and the power of each element is  $\beta^{-1}$. This motivates the adoption of the Gaussian likelihood function as follow:
\begin{align}
&p \left(  \mathcal Y  \mid \{\boldsymbol U^{(n)}\}_{n=1}^N, \beta  \right) \nonumber \\
&\propto \exp \left( - \frac{\beta}{2} \parallel \mathcal Y - \llbracket  \boldsymbol U^{(1)},  \boldsymbol U^{(2)},..., \boldsymbol U^{(N)}  \rrbracket   \parallel_F^2 \right). \label{eq6}
\end{align}
Equation \eqref{eq6} assumes that the signal tensor $ \llbracket  \boldsymbol U^{(1)},  \boldsymbol U^{(2)},..., \boldsymbol U^{(N)}  \rrbracket$ is corrupted by AWGN tensor $\mathcal W$ with each element having power $\beta^{-1}$.  This is consistent with the least-squares (LS) problem in \eqref{eq1}  if the AWGN power $\beta^{-1}$ is known.  However, in Bayesian modeling, $\beta$ is modeled as another random variable.  Since we have no prior information about the noise power, a non-informative prior $p(\beta) = \mathrm{gamma}(\beta| \epsilon, \epsilon)$ with a very small $\epsilon$ (e.g., $10^{-6}$) is usually employed. 

By using the introduced prior distributions and likelihood function, a probabilistic model for tensor CPD was constructed, as illustrated in Figure 4. Based on this model, a VI based algorithm was derived in \cite{PI1} that can automatically drive most of the columns in each factor matrix to zero, by which the tensor rank is revealed. Inspired by the vanilla probabilistic CPD using the Gaussian-gamma prior, other structured and large-scale tensor CPDs with automatic tensor rank learning were further developed  \cite{PI2, CL2, CL3, CL5, C9, C10, C12}  in recent years.

\begin{figure}[!t]
\centering
\includegraphics[width= 3.5 in]{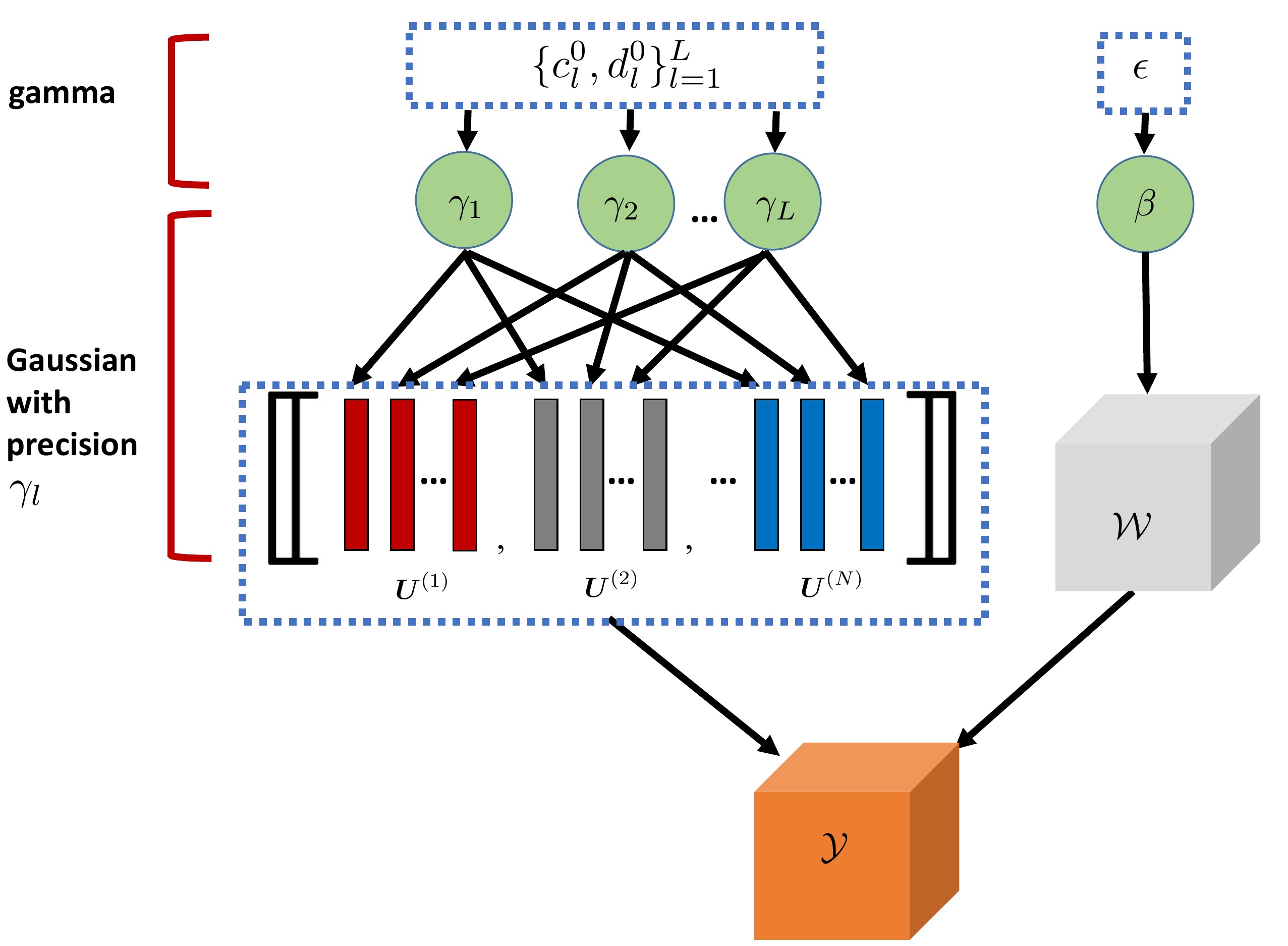}
\caption{Probabilistic CPD model with Gaussian-gamma prior.}
\label{fig_topology}
\end{figure}

{\color{black} The success of the previous works on automatic tensor rank learning}  \cite{PI1, PI2, CL2, CL3, CL5, C9, C10,  C12}  comes from the adoption of the sparsity-promoting Gaussian-gamma prior, while their performances are also limited by the rigid central and tail behaviors  in modeling different levels of the sparsity. {\color{black} More specifically, Gaussian-gamma prior is effective only if the sparsity pattern of data matches well with that of  Gaussian-gamma model. For example, when the tensor rank $R$ is low,   previous results have shown that a relatively large upper bound value $L$ (e.g., the maximal value of tensor dimensions \cite{PI1, PI2})  can give accurate tensor rank estimation.  
However, for a high tensor rank $R$, the upper bound value $L$ selected for the low-rank case would be too small to render a sparsity pattern of columns, and thus it leads to performance degradation. Even though we can increase the value of $L$ to  large numbers, as it will be shown in Figure 8 (b)-(c), tensor rank learning accuracy using Gaussian-gamma prior is still not satisfactory, showing its lack of flexibility to adapt to different sparsity levels.}  Therefore, to further enhance the tensor rank learning capability,  we explore the use of sparsity-promoting priors with more flexible central and tail behaviors.

\section{Novel Probabilistic Modeling: When  Tensor CPD Meets Generalized Hyperbolic Distribution}

In particular, we focus on the GH prior, since it not only includes the Gaussian-gamma prior as a special case, but also it can be treated as the generalization of other widely-used sparsity-promoting distributions including the Laplacian distribution, normal-inverse chi-squared distribution, normal-inverse gamma distribution,  variance-gamma distribution and Mckay's Bessel distribution \cite{AD4}. Therefore, it is expected that the functional flexibility of GH prior could lead to  more flexibility in modeling different sparsity levels and thus more accurate learning for tensor rank.  

%To see this more clearly, we firstly review some basic properties of GH distribution in the context of tensor CPD.

Recall that the model building block is the $l^{th}$ column group $ \{\boldsymbol  U^{(n)}_{:,l} \}_{n=1}^N$. With the  GH prior on each column group, we have a new prior distribution for factor matrices:
\begin{align}
& p ( \{ \boldsymbol  U^{(n)} \}_{n=1}^N ) = \prod_{l=1}^L \mathrm{GH}(\{ \boldsymbol  U^{(n)}_{:,l} \}_{n=1}^N | a_l^0, b_l^0, \lambda_l^0) \nonumber \\
%&=  \prod_{l=1}^L  2^{ \sum_{n=1}^N J_n } (\frac{1}{\pi})^{\frac{ \sum_{n=1}^N J_n}{2}} \frac{\Gamma(c_l^0 +  \sum_{n=1}^N \frac{J_n}{2} )}{ {2d_l^0}^{-c_l^0} \Gamma(c_l^0)} \nonumber \\
%&\! \times \! \left(2d_l^0 + \mathrm{vec}\left(  \{\boldsymbol  U^{(n)}_{:,l}\}_{n=1}^N \right)^T\mathrm{vec}\left(  \{\boldsymbol  U^{(n)}_{:,l}\}_{n=1}^N \right) \right)^{-c_l^0 -  \sum_{n=1}^N \frac{J_n}{2} }, \nonumber\\
& = \prod_{l=1}^L \frac{ (a^0_l)^{\frac{\sum_{n=1}^N J_n}{4} }} { (2\pi)^{\frac{\sum_{n=1}^N J_n}{2} }} \frac{(b^0_l)^{\frac{-\lambda_l^0}{2}}}{K_{\lambda_l^0} \left(\sqrt{a^0_lb^0_l} \right)}  \nonumber \\
& \times \frac{K_{\lambda_l^0 -  \frac{\sum_{n=1}^N J_n}{2}} \left( \sqrt{a_l^0 \left( b_l^0 +  \bigparallel \mathrm{vec}\left(  \{\boldsymbol  U^{(n)}_{:,l}\}_{n=1}^N \right){\bigparallel}_2^2  \right)  } \right)  }{\left( b_l^0 +  \bigparallel \mathrm{vec}\left(  \{\boldsymbol  U^{(n)}_{:,l}\}_{n=1}^N \right){\bigparallel}_2^2 \right)} \label{eq7},
\end{align}
where $K_{\cdot} (\cdot)$ is the modified Bessel function of the second kind, and $\mathrm{GH}(\{ \boldsymbol  U^{(n)}_{:,l} \}_{n=1}^N | a_l^0, b_l^0, \lambda_l^0)$ denotes the GH prior on the $l^{th}$ column group $\{ \boldsymbol  U^{(n)}_{:,l} \}_{n=1}^N$, in which the hyper-parameters $\{a_l^0, b_l^0, \lambda_l^0\}$ control the shape of the distribution. By setting  $\{a_l^0, b_l^0, \lambda_l^0\}$ to specific values, the GH prior \eqref{eq7} reduces to other prevalent sparsity-promoting priors. Details on how GH prior reduces to student-t and Laplacian distributions are given in Appendix B.

\begin{figure}[!t]
\centering
\includegraphics[width= 3.5 in]{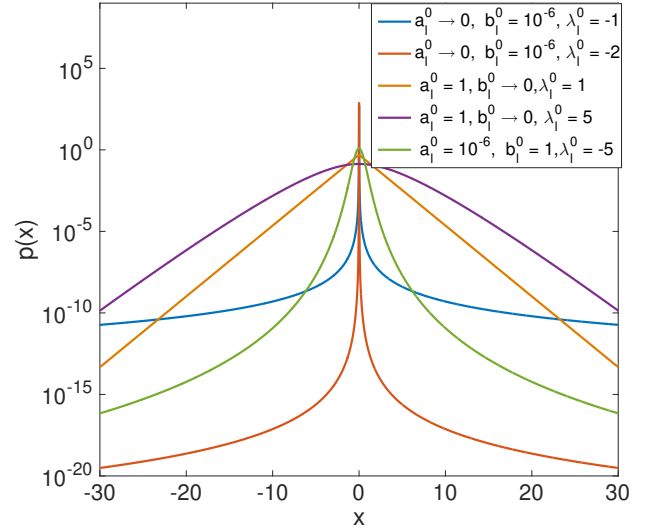}
\caption{Univariate marginal probability density function in  \eqref{eq7} with different values of hyper-parameters.}
\label{fig_topology}
\end{figure}

To visualize the GH distribution and its special cases,  the univariate GH pdfs with different hyper-parameters are illustrated in Figure 5. It can be observed that the blue line is with a similar shape to those of the student't distributions in Figure 3, while the orange one resembles the shapes of Laplacian distributions \cite{AD4,AD5}.  For other lines, they exhibit a wide range of the central and tail behaviors of the pdfs. This reveals the great functional flexibility of the GH prior in modeling different levels of sparsity.

On the other hand, the GH prior \eqref{eq7} can be expressed as a Gaussian scale mixture formulation \cite{AD4,AD5}:
\begin{align}
& p ( \{ \boldsymbol  U^{(n)} \}_{n=1}^N ) = \prod_{l=1}^L \mathrm{GH}(\{ \boldsymbol  U^{(n)}_{:,l} \}_{n=1}^N | a_l^0, b_l^0, \lambda_l^0) \nonumber \\
& = \prod_{l=1}^L  \int \mathcal N \left(  \mathrm{vec}\left(  \{\boldsymbol  U^{(n)}_{:,l}\}_{n=1}^N \right) | \boldsymbol 0_{\sum_{n=1}^N J_n \times 1},  z_l \boldsymbol I_{\sum_{n=1}^N J_n}  \right) \nonumber \\
&~~~~~~~~~ \times \mathrm{GIG}(z_l | a_l^0, b_l^0, \lambda_l^0) d{z_l} \label{eq11} ,
\end{align}
where $z_l$ denotes the variance of the Gaussian distribution, and $ \mathrm{GIG}(z_l | a_l^0, b_l^0, \lambda_l^0)$ represents the generalized inverse Gaussian (GIG) pdf: 
\begin{align}
 &\mathrm{GIG}(z_l | a_l^0, b_l^0, \lambda_l^0) \nonumber\\
 &=  \frac{\left(\frac{a_l^0}{b_l^0}\right)^{\frac{\lambda_l^0}{2}}}{2K_{\lambda_l^0} \left(\sqrt{a_l^0 b_l^0} 
\right) } z_l^{\lambda_l^0-1} \exp\left(-\frac{1}{2} \left( a_l^0 {\color{black} z_l} + b_l^0 z_l^{-1}\right) \right).
\end{align}
This Gaussian scale mixture formulation suggests that each GH distribution $\mathrm{GH}(\{ \boldsymbol  U^{(n)}_{:,l} \}_{n=1}^N | a_l^0, b_l^0, \lambda_l^0) $ can be regarded as an infinite mixture of Gaussians with the mixing distribution being a GIG distribution. Besides revealing its inherent structure,  the formulation \eqref{eq11} allows a hierarchical construction of each GH prior by introducing the latent variable $z_l$, as illustrated in Figure 6. This gives us the following important conjugacy property \cite{AD4}.

\begin{property}
For probability density functions (pdfs)
\begin{align}
& p\left(\{\boldsymbol  U^{(n)}_{:,l}\}_{n=1}^N | z_l \right)  \nonumber \\
&=  \mathcal N \left(  \mathrm{vec}\left(  \{\boldsymbol  U^{(n)}_{:,l}\}_{n=1}^N \right)  | \boldsymbol 0_{\sum_{n=1}^N J_n \times 1},  z_l \boldsymbol I_{\sum_{n=1}^N J_n}  \right),  \\
& p(z_l) = \mathrm{GIG}(z_l | a_l^0, b_l^0, \lambda_l^0), 
\end{align} 
pdf $ p(z_l)$ is conjugate\footnote{In Bayesian theory, a probability density function (pdf) $p(x)$ is said to be conjugate to a conditional pdf $p(y|x)$ if the resulting posterior pdf $p(x|y)$ is in the same distribution family as $p(x)$.} to $p\left(\{\boldsymbol  U^{(n)}_{:,l}\}_{n=1}^N | z_l \right)$.
\end{property}
As will be seen later, the conjugacy property greatly facilitates the derivation of the Bayesian inference algorithm. 

\begin{figure}[!t]
\centering
\includegraphics[width= 3.5 in]{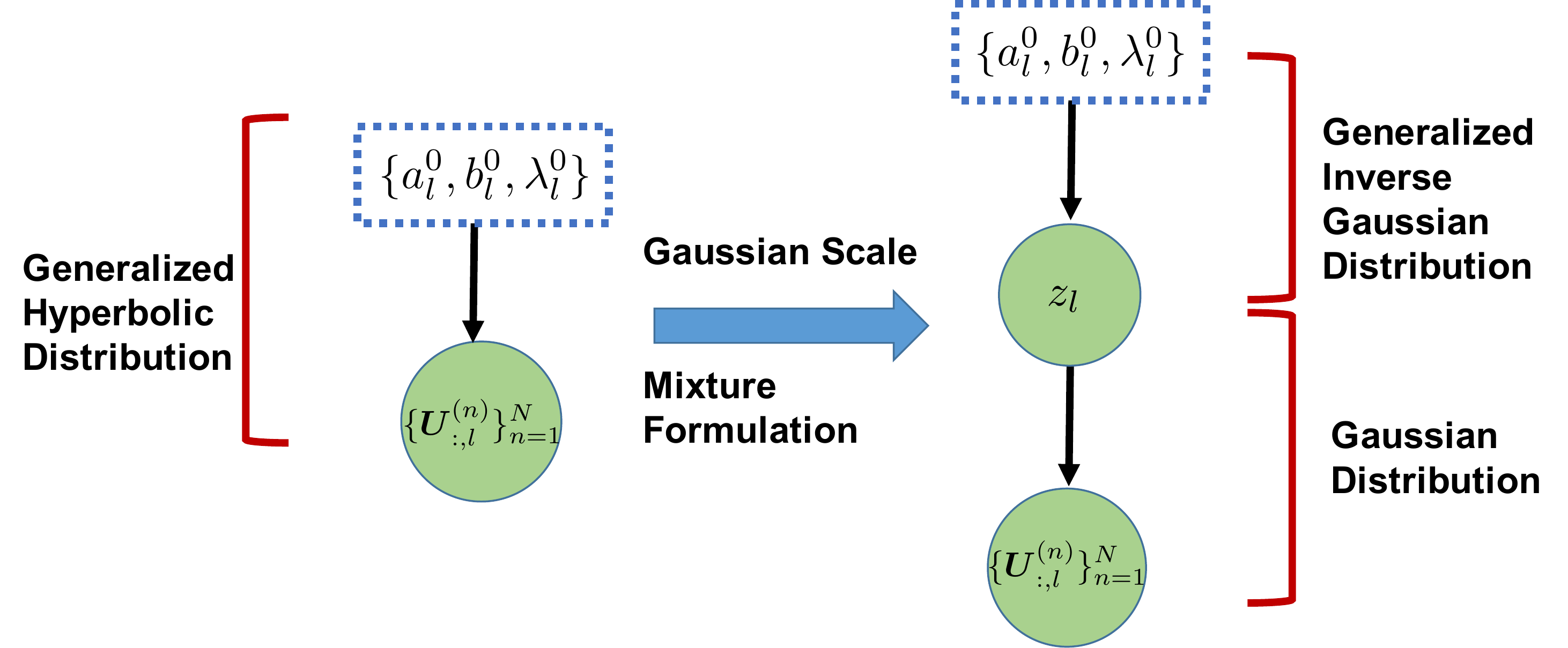}
\caption{Hierarchical construction of GH distribution.}
\label{fig_topology}
\end{figure}

\begin{figure}[!t]
\centering
\includegraphics[width= 3.5 in]{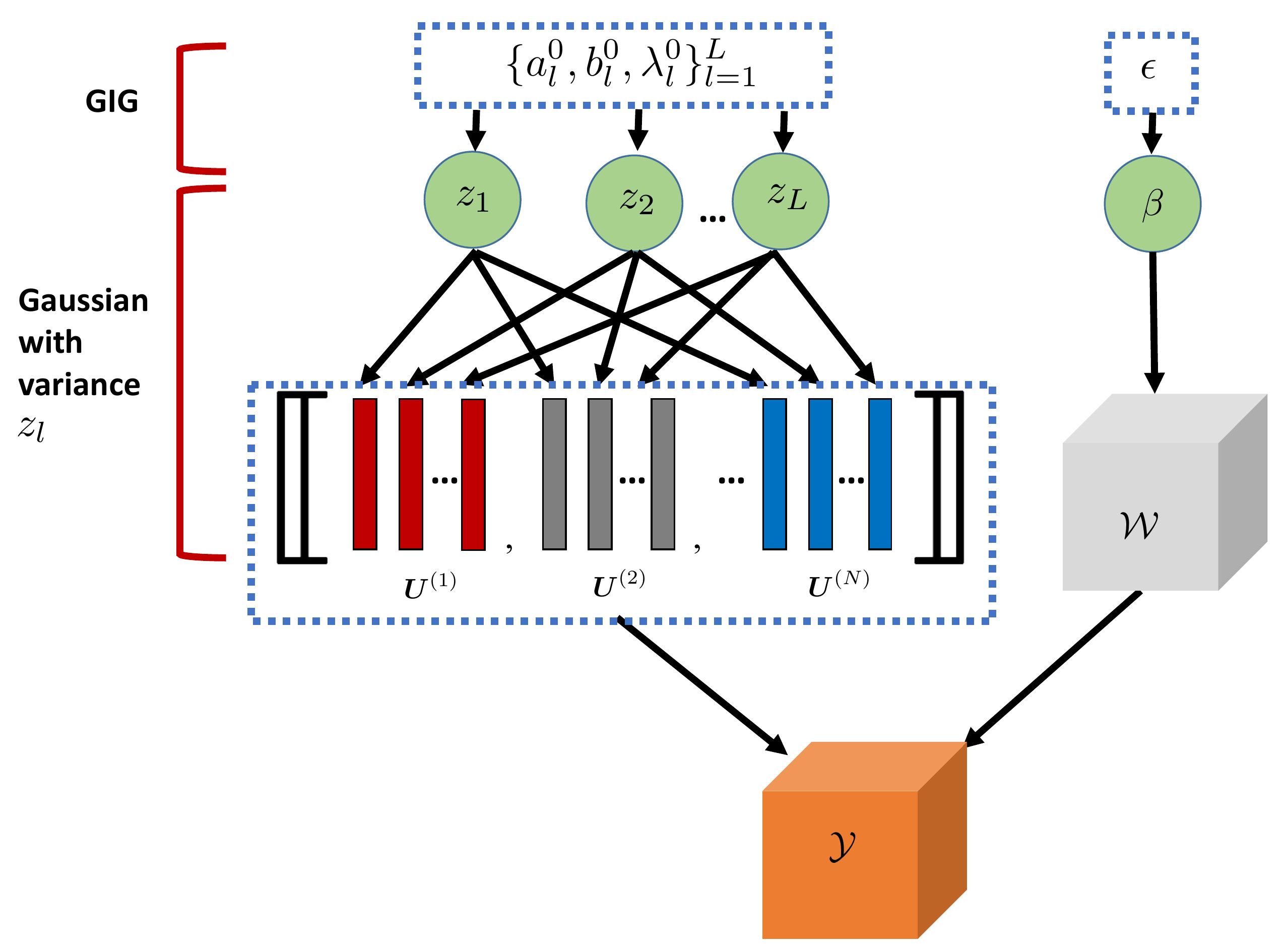}
\caption{The probabilistic tensor CPD model with GH prior.}
\label{fig_topology}
\end{figure}

Finally, together with the likelihood function in \eqref{eq6}, we propose a novel probabilistic model for tensor CPD using the hierarchical construction of the GH prior,  as shown in Figure 7. Denoting the model parameter set $\boldsymbol \Theta = \{  \{\boldsymbol  U^{(n)}\}_{n=1}^N, \{z_l\}_{l=1}^L, \beta\}$, the proposed probabilistic tensor CPD model can be fully described by the joint pdf $p(\mathcal Y, \boldsymbol \Theta )$ as
\begin{align}
&p(\mathcal Y, \boldsymbol \Theta ) = p \left(  \mathcal Y  \mid \{\boldsymbol  U^{(n)}\}_{n=1}^N, \beta  \right) p\left( \{\boldsymbol  U^{(n)}\}_{n=1}^N | \{ z_l\}_{l=1}^L \right)  \nonumber \\
& ~~~~~~~~~~~ \times p\left( \{ z_l\}_{l=1}^L  \right)p(\beta) \nonumber \\
& \propto \exp \Bigg\{  \frac{\prod_{n=1}^N J_n}{2}  \ln \beta  - \frac{\beta}{2} \parallel \mathcal Y - \llbracket  \boldsymbol U^{(1)},  \boldsymbol U^{(2)},..., \boldsymbol U^{(N)}  \rrbracket   \parallel_F^2   \nonumber \\
& + \sum_{n=1}^N \left[ \frac{J_n}{2} \sum_{l=1}^L \ln z_l^{-1}  - \frac{1}{2} \mathrm{Tr}\left( \boldsymbol  U^{(n)} \boldsymbol Z^{-1} \boldsymbol  U^{(n)T} \right)  \right] \nonumber \\
& + \sum_{l=1}^L \Big[  \frac{\lambda_l^0}{2} \ln \frac{a_l^0}{b_l^0} - \ln \left[2K_{\lambda_l^0} \left (\sqrt{a_l^0b_l^0} \right)\right]  + (\lambda_l^0-1) \ln z_l \nonumber \\
& - \frac{1}{2} \left( a_l^0 z_l + b_l^0 z_l^{-1}\right)\Big]  + (\epsilon -1) \ln \beta - \epsilon \beta \Bigg\}, \label{eq15}
\end{align}
where $\boldsymbol Z = \mathrm{diag}\{z_1, z_2, \cdots, z_L\}$.

\section{Inference Algorithm}

\subsection{General Philosophy of the Variational Inference}

Given the probabilistic model $p(\mathcal Y, \boldsymbol \Theta)$, the next task is to learn the model parameters in $\boldsymbol \Theta$ from the tensor data $\mathcal Y$, in which the posterior distribution $p(\boldsymbol \Theta | \mathcal Y)$ is to be sought. However, for such a complicated probabilistic model (11), the multiple integrations in computing the posterior distribution   $p(\boldsymbol \Theta | \mathcal Y)$ is not tractable. Fortunately, this challenge is not new, and similar obstacles have been faced in inferring other complicated Bayesian machine learning models such as Bayesian neural networks \cite{stein_vi1, Sergios3},  Bayesian structured matrix factorization \cite{matrix_decomp}, latent dirichlet allocation \cite{latent_da}, and Gaussian mixture model \cite{GMM}. It has been widely agreed that  variational inference (VI), due to its efficiency in computations and the theoretical guarantee of convergence, is the major driving force for inferring complicated probabilistic models \cite{VI2}. Rather than manipulating a huge number of samples from the probabilistic model, VI recasts the originally intractable multiple integration problem into the following functional optimization problem:
\begin{align}
& \min_ {Q(\boldsymbol \Theta)} \mathrm {KL} \big (Q\left( \boldsymbol  \Theta \right) \parallel  p \left(  \boldsymbol  \Theta \mid  \mathcal Y \right) \big) \nonumber \\
&~~~~~~~~\triangleq  - \mathbb E_{Q \left( \boldsymbol  \Theta \right) } \left \{ \ln  \frac{p\left(  \boldsymbol  \Theta \mid  \mathcal Y  \right)}{ Q \left( \boldsymbol  \Theta \right) }  \right\} \nonumber \\ 
&\mathrm{s.t.} ~~ Q(\boldsymbol \Theta) \in \mathcal F \label{eq16},
\end{align}
where $\mathrm  {KL} (\cdot || \cdot)$ denotes  the Kullback-Leibler (KL) divergence between two arguments, and $\mathcal F$ is a pre-selected family of pdfs. Its philosophy is to seek a tractable variational pdf $Q(\boldsymbol \Theta)$ in $\mathcal F$ that is the closest to the true posterior distribution $p(\boldsymbol \Theta | \mathcal Y)$ in terms of the KL divergence. Therefore, the art is to determine the family $\mathcal F$ to balance the tractability of the algorithm and the accuracy of the posterior distribution learning. In this paper, we adopt the mean-field family due to its prevalence in Bayesian tensor research \cite{PI1, PI2, CL2, CL3, CL5, C9, C10, C12}. Other advanced choices could be found in \cite{VI2}.

Using the mean-field family, which restricts $Q(\boldsymbol \Theta) = \prod_{k=1}^K Q(\boldsymbol \Theta_k)$ where $\boldsymbol \Theta$ is partitioned into mutually disjoint non-empty subsets $\boldsymbol \Theta_k$ (i.e., $\boldsymbol \Theta_k$ is a part of $\boldsymbol \Theta$ with $\cup_{k=1}^K \boldsymbol \Theta_k = \boldsymbol \Theta$ and $\cap_{k=1}^K \boldsymbol \Theta_k = \O $), the KL divergence minimization problem \eqref{eq16} becomes 
\begin{align}
& \min_ { \{ Q(\boldsymbol \Theta_k)\}_{k=1}^K}  - \mathbb E_{ \{ Q(\boldsymbol \Theta_k)\}_{k=1}^K } \left \{ \ln  \frac{p\left(  \boldsymbol  \Theta \mid  \mathcal Y  \right)}{ \prod_{k=1}^K Q(\boldsymbol \Theta_k) }  \right\} \label{eq17}. 
\end{align} 
The factorable structure in \eqref{eq17} inspires the idea of block minimization in optimization theory \cite{NP}. In particular, after fixing variational pdfs  $\{Q(\boldsymbol \Theta_j)\}_{j \neq k}$ other than $Q(\boldsymbol \Theta_k)$, the remaining problem is 
\begin{align}
\min_{Q(\boldsymbol \Theta_k)} \int Q(\boldsymbol \Theta_k)(-\mathbb E_{\prod_{j \neq k} Q(\boldsymbol \Theta_j)} \left[ \ln p(\boldsymbol \Theta, \mathcal Y)\right] + \ln Q(\boldsymbol \Theta_k) ) d{\boldsymbol \Theta_k},
\end{align}
and it has been shown  that the optimal solution is \cite{beal, Sergios2}:
\begin{align}
& Q^* \left(  \boldsymbol \Theta_k\right) = \frac{\exp\left ( \mathbb E_{ \prod_{j \neq k}Q \left(   \boldsymbol \Theta_j\right) } \left [ \ln  {p\left(  \boldsymbol  \Theta ,  \mathcal Y  \right)} \right]  \right)}{\int \exp\left ( \mathbb E_{ \prod_{j \neq k}Q \left(   \boldsymbol \Theta_j\right) } \left [ \ln  {p\left(  \boldsymbol  \Theta ,  \mathcal Y \right)} \right]  \right)  d\boldsymbol \Theta_k }. \label{eq19}
\end{align}
{\color{black} More discussions on the mean-field VI are provided in Appendix M. }

\begin{table}[!t]
\centering
\caption{Optimal variational density functions.}
\scalebox{0.7} {\begin{tabular}{@{}|l|l|@{}}
\toprule
Variational pdfs  &  Remarks \\ \midrule
$Q^*\left(  \boldsymbol  U^{(k)}\right) = \mathcal{MN} \left(\boldsymbol  U^{(k)} |\boldsymbol M^{(k)}, \boldsymbol I_{J_n}, \boldsymbol \Sigma^{(k)} \right), \forall k$  & Matrix normal distribution \\ & with mean $\boldsymbol M^{(k)}$ and covariance matrix  $\boldsymbol \Sigma^{(k)}$ \\& given in (16) and (17), respectively.  \\ \midrule
$Q^* \left( z_l\right)  =  \mathrm{GIG}( z_l | a_l, b_l, \lambda_l), \forall l$  &  Generalized inverse Gaussian distribution with \\& parameters $\{a_l, b_l, \lambda_l \}$ given in (18)-(20). \\ \midrule
$Q^* \left( \beta \right)  =  \mathrm{gamma}(\beta | e, f) $ & Gamma distribution with shape $e$ and rate $f$ \\& given in (21), (22). \\ \bottomrule
\end{tabular}}
\end{table}

\begin{table}[!t]
\centering
\caption{Computation results of expectations.}
\scalebox{0.8} {\begin{tabular}{@{}|l|l|@{}}
\toprule
Expectations &  Computation Results \\ \midrule
$\mathbb E \left[ \boldsymbol  U^{(k)}\right], \forall k$  &  $\boldsymbol M^{(k)},  \forall k$  \\ \midrule
$ \mathbb E \left[  z_l \right] ,\forall l$  &  $ \left( \frac{b_l}{a_l}\right)^{\frac{1}{2}} \frac{K_{\lambda_l +1} \left(\sqrt{a_l b_l}\right)}{K_{\lambda_l} \left(\sqrt{a_l b_l} \right) }$ \\ \midrule
$ \mathbb E \left[  z_l^{-1}\right] ,\forall l$  &  $ \left( \frac{b_l}{a_l}\right)^{-\frac{1}{2}} \frac{K_{\lambda_l -1} \left(\sqrt{a_l b_l}\right)}{K_{\lambda_l} \left(\sqrt{a_l b_l} \right) }$ \\ \midrule
$\mathbb E \left[\beta \right]  $ & $\frac{e}{f}$   \\ \midrule
$\mathbb E \left[  \left[\boldsymbol  U^{(n)}_{:,l} \right]^T \boldsymbol  U^{(n)}_{:,l}  \right]  $ & $ \left[\boldsymbol M^{(n)}_{:,l}\right]^T  \boldsymbol M^{(n)}_{:,l} + J_n \boldsymbol \Sigma^{(n)}_{l,l}$ \\ \midrule
$\mathbb E \Bigg[ \left(\mathop \odot  \limits_{n=1,n\neq k}^N \boldsymbol  U^{(n)}\right)^T$ & $ \mathop \circledast \limits_{n=1, n\neq k}^N  \left[  \left[\boldsymbol M^{(n)} \right]^T \boldsymbol M^{(n)}  + J_n 
\boldsymbol \Sigma^{(n)}  \right]$ 
\\ $~~~~ \times \left(\mathop \odot  \limits_{n=1,n\neq k}^N \boldsymbol  U^{(n)}\right) \Bigg] $ &  { } \\ \midrule
$ \mathbb E \left[ \bigparallel  \mathcal Y-   \llbracket \boldsymbol  U^{(1)}, \cdots, \boldsymbol  U^{(N)} \rrbracket  {\bigparallel}_F^2\right] $ & 
$ \parallel  \mathcal Y \parallel_F^2 + \mathrm {Tr} \Big ( \mathop \circledast \limits_{n=1}^{N} \Big[\boldsymbol  M^{(n)T} \boldsymbol  M^{(n)} +  J_n\boldsymbol   \Sigma^{(n)} \Big]\Big) $ \\ { }  &  $ -2 \mathrm {Tr} \Big( \mathcal Y(1) \Big( \mathop \odot \limits_{n= 2}^{N} \boldsymbol  M^{(n)} \Big )\boldsymbol  M^{(1)T} \Big) $
 \\ \bottomrule
\end{tabular}}
\end{table}

\subsection{Deriving Optimal Variational Pdfs}
The optimal variational pdfs $\{Q^* \left(  \boldsymbol \Theta_k\right)\}_{k=1}^K$ can be obtained by substituting \eqref{eq15} into \eqref{eq19}. Although straightforward as it may seem, the involvement of tensor algebras in  \eqref{eq15} and the multiple integrations in the denominator of  \eqref{eq19}   make the derivation a challenge. On the other hand, since the proposed probabilistic model  employs the GH prior,  and is different from previous works using Gaussian-gamma prior  \cite{PI1, PI2,C9, C10, C12, CL2, CL3, CL5}, each optimal variational pdf $Q^* \left(  \boldsymbol \Theta_k \right)$ needs to be derived from first principles. To keep the main body of this paper concise,  the lengthy derivations are moved to Appendix C, and we only present the optimal variational pdfs in Table I at the top of this page.

In particular, the optimal variational pdf $Q^*(\boldsymbol  U^{(k)})$ was derived to be a matrix normal distribution \cite{MND} $\mathcal{MN} \left(\boldsymbol  U^{(k)} |\boldsymbol M^{(k)}, \boldsymbol I_{J_n}, \boldsymbol \Sigma^{(k)} \right)$ with the covariance matrix
\begin{align}
\boldsymbol \Sigma^{(k)} = &  \Bigg[  \mathbb E\left[ \beta\right] \mathbb E \Bigg[ \left(\mathop \odot  \limits_{n=1,n\ne k}^N \boldsymbol  U^{(n)}\right)^T \nonumber\\
& ~~~~~~~~ \times \left( \mathop  \odot  \limits_{n=1,n\ne k}^N \boldsymbol  U^{(n)} \right) \Bigg] + \mathbb E \left[ \boldsymbol Z^{-1} \right] \Bigg]^{-1}, \label{eq20}
\end{align}
and mean matrix
\begin{align}
&\boldsymbol M^{(k)}  = \mathcal Y (k)  \mathbb E\left[ \beta \right] \left( \mathop \odot  \limits_{n=1,n\ne k}^N  \mathbb E\left[\boldsymbol  U^{(n)}\right] \right) \boldsymbol \Sigma^{(k)}. \label{eq21}
\end{align}
In \eqref{eq20} and \eqref{eq21}, $\mathcal Y{(k)}$ is a matrix obtained by unfolding the tensor $\mathcal Y$ along its $k^{th}$ dimension, and the multiple Khatri-Rao products $\mathop \odot  \limits_{n=1,n\ne k}^N   {\boldsymbol  A}^{(n)} =  {\boldsymbol  A}^{(N)}  \odot {\boldsymbol  A}^{(N-1)} \odot \cdots \odot {\boldsymbol  A}^{(k+1)} \odot  {\boldsymbol  A}^{(k-1)} \odot \cdots \odot  {\boldsymbol  A}^{(1)} $.  The expectations are taken with respect to the corresponding variational pdfs of the arguments.  For the optimal variational pdf $Q(z_l)$, by using the conjugacy result in \emph{Property 1}, it can be derived to be a GIG distribution   $\mathrm{GIG}( z_l | a_l, b_l, \lambda_l)$ with parameters
\begin{align}
&a_l = a_l^0, \label{al}\\
&b_l =  b_l^0 + \sum_{n=1}^N \mathbb E \left[  \left[\boldsymbol  U^{(n)}_{:,l} \right]^T \boldsymbol  U^{(n)}_{:,l}  \right], \label{bl}\\
&\lambda_l = \lambda_l^0 - \frac{1}{2}\sum_{n=1}^N J_n. \label{lambdal}
\end{align}
Finally, the optimal variational pdf  $Q(\beta)$ was derived to be a gamma distribution  $\mathrm{gamma}(\beta | e, f) $ with parameters
\begin{align}
&e =  \epsilon  + \frac{1}{2}\prod_{n=1}^N J_n,   \label{e}\\
&f =  \epsilon + \frac{1}{2}\mathbb E \left[ \bigparallel  \mathcal Y-   \llbracket \boldsymbol  U^{(1)}, \cdots, \boldsymbol  U^{(N)} \rrbracket  {\bigparallel}_F^2\right]. \label{f}
\end{align}

In (16)-(22), there are several expectations to be computed. They can be obtained either from the statistic literatures \cite{MND} or similar results in related works \cite{PI1, PI2,C9, C10, C12, CL2, CL3, CL5}. For easy reference, we listed the expectation results needed for (16)-(22) in Table II, where $\mathop \circledast  \limits_{n=1,n\ne k}^N   {\boldsymbol  A}^{(n)} =  {\boldsymbol  A}^{(N)} \circledast {\boldsymbol  A}^{(N-1)} \circledast \cdots \circledast {\boldsymbol  A}^{(k+1)} \circledast {\boldsymbol  A}^{(k-1)} \circledast \cdots \circledast {\boldsymbol  A}^{(1)} $ is the multiple Hadamard products.

\subsection{Setting the Hyper-parameters}
From Table I, it can be found that the variational pdf $Q(\boldsymbol  U^{(k)})$ and $\{Q(z_l)\}_{l=1}^L$ forms a more complicated Gaussian-GIG pdf pair than that in \emph{Property 1}. Therefore, the shape of the variational pdf $Q(\boldsymbol  U^{(k)})$, which determines both the factor matrix recovery and tensor rank learning, is affected by the variational pdf $\{Q(z_l)\}_{l=1}^L$. For each $Q(z_l)$, as seen in (18)-(20), its shape relies on the pre-selected hyper-parameters $\{a_l^0, b_l^0, \lambda_l^0\}$. In practice, we usually have no prior knowledge about the sparsity level before assessing the data, and a widely adopted approach is to make the prior non-informative. 

In previous works using Gaussian-gamma prior \cite{PI1, PI2,C9, C10,  C12, CL2, CL3, CL5}, hyper-parameters are set equal to very small values in order to approach  a non-informative prior. Although nearly zero hyper-parameters lead to an improper prior, the derived variational pdf is still proper since these parameters are updated using information from observations \cite{PI1, PI2,C9, C10, C12, CL2, CL3, CL5}. Therefore, in these works, the strategy of using non-informative prior is valid. On the other hand, for the employed GH prior, non-informative prior requires $\{a_l^0, b_l^0, \lambda_l^0\}$ all go to zero, which however would lead to an improper variational pdf $Q(z_l)$, since its parameter $a_l = a_l^0$ is fixed (as seen in (18)).  This makes the expectation computation $\mathbb E[z_l]$ in Table II problematic.
%and thus is not desirable. 

To tackle this issue, another  viable approach is to optimize these hyper-parameters $\{a_l^0, b_l^0, \lambda_l^0\}$ so that they can be adapted during the procedure of model learning. However, as seen in \eqref{eq15}, these three parameters are coupled together via the nonlinear modified Bessel function, and thus optimizing them jointly is prohibitively difficult. Therefore, in this paper,  we propose to only optimize the most critical one, i.e., $a_l^0$, since it directly determines the shape of $Q(z_l)$ and will not be updated in the learning procedure. For the other two parameters $\{b_l^0, \lambda_l^0\}$, as seen in (19) and (20), since they are updated with model learning results or tensor dimension,  according to the Bayesian theory \cite{Bayesian}, their effect on the posterior distribution would become negligible when the observation tensor is large enough. This justifies the optimization of  $a_l^0$ while not 
$\{b_l^0, \lambda_l^0\}$.

For optimizing $a_l^0$, following related works \cite{beal,Beal1}, we introduce a conjugate hyper-prior $p(a_l^0) = \mathrm{gamma}( a_l^0 | \kappa_{a_1}, \kappa_{a_2})$ to ensure the positiveness of $a_l^0$ during the optimization. To bypass the nonlinearity from the modified Bessel function, we set $b_l^0 \rightarrow 0$ so that $K_{\lambda_l^0} \left (\sqrt{a_l^0b_l^0}\right)$ becomes a constant.  In the framework of VI, after fixing other variables, it has been derived in Appendix C that the hyper-parameter $a_l^0$ is updated via
\begin{align}
a_l^0 = \frac{\kappa_{a_1}+ \frac{\lambda_l^0}{2} - 1}{\kappa_{a_2} + \frac{\mathbb E[z_l]}{2}}. \label{al0}
\end{align}
Notice that it requires  $\kappa_{a_1} > 1 -\lambda_l^0/2$ and $\kappa_{a_2} \geq 0$ to ensure the positiveness of $a_l^0$.

\subsection{Algorithm Summary and Insights}
From (16)-(23), it can be seen that the statistics of each variational pdf rely on other variational pdfs. Therefore, they need to be updated in an alternating fashion, giving rise to an iterative algorithm summarized in {\bf Algorithm 1}. To gain more insights from the proposed algorithm, discussions on its convergence property, computational complexity, and automatic tensor rank learning  are presented in the following.
\begin{algorithm}[!t]
    \caption{\bf Probabilistic Tensor CPD Using The GH Prior}
\noindent {\bf Initializations:}
Choose $L > R$ and initial values $\{\left[\boldsymbol M^{(n)}\right]^0, \left[\boldsymbol \Sigma^{(n)}\right]^0\}_{n=1}^N $, $\{m[z_l^{-1}]^{0}, a_l^0, b_l^0, \lambda_l^0\}_{l=1}^L$, $e^0, f^0$.  Choose $\kappa_{a_1} > -\lambda_l^0/2$ and $\kappa_{a_2} \geq 0$.

\noindent {\bf Iterations:} 

For the iteration $t+1$ ($t \geq 0$),

\noindent \underline{For $k = 1,\cdots, N$, update the parameters of  $Q(\boldsymbol  U^{(k)})^{t+1}$:}
\begin{align}
&\left[\boldsymbol \Sigma^{(k)}\right]^{t+1} \!\!= \!\! \Bigg[ \frac{c^t}{d^t} \mathop \circledast \limits_{n=1, n\neq k}^N \!\! \Bigg[  \left(\left[ \boldsymbol M^{(n)} \right]^{s} \right)^T \left[\boldsymbol M^{(n)}  \right]^{s} \nonumber\\
& + J_n \left[\boldsymbol \Sigma^{(n)}\right]^{s} \Bigg]  +\mathrm{diag}\left\{ m[z_1^{-1}]^{t}, m[z_2^{-1}]^{t}, ..., m[z_L^{-1}]^{t} \right \} \Bigg]^{-1},
\end{align}
\begin{align}
&\left[\boldsymbol M^{(k)}\right]^{t+1} \!\!=\!\! \mathcal Y(k) \frac{c^t}{d^t}  \left( \mathop \odot  \limits_{n=1,n\ne k}^N  \left[\boldsymbol M^{(n)}\right]^s \right) \left[\boldsymbol \Sigma^{(n)}\right]^{t+1},
\end{align}
where $s$ denotes the most recent update index, i.e., $s = t+1$ when $n < k$ , and $s = t$ otherwise.

\noindent \underline{Update the parameters of  $Q(z_l)^{t+1}$:}
\begin{align}
& a_l^{t+1}=  [a_l^0]^{t} , \\
& b_l^{t+1} =  b_l^0 +  \sum_{n=1}^N  \Bigg[ \left(\left[\boldsymbol M^{(n)}_{:,l}\right]^{t+1}\right)^T  \left[\boldsymbol M^{(n)}_{:,r}\right]^{t+1} \nonumber\\
& ~~~~~~~~~~~~~~~~~~~~~~~~~~~~~~~~~~+ J_n \boldsymbol \left[\boldsymbol \Sigma^{(n)}_{l,l}\right]^{t+1} \Bigg], \\
&\left[\lambda_l\right]^{t+1} =  \lambda_l^0  - \frac{1}{2}\sum_{n=1}^N J_n, \\
& m[z_l^{-1}]^{t+1}  =  \left( \frac{b_l^{t+1}}{a_l^{t+1}}\right)^{-\frac{1}{2}} \frac{K_{\left[\lambda_l\right]^{t+1} -1} \left(\sqrt{a_l^{t+1} b_l^{t+1}}\right)}{K_{\left[\lambda_l\right]^{t+1}} \left(\sqrt{a_l^{t+1} b_l^{t+1}} \right) },  \\
& m[z_l]^{t+1} =  \left( \frac{b_l^{t+1}}{a_l^{t+1}}\right)^{\frac{1}{2}} \frac{K_{\left[\lambda_l\right]^{t+1} +1} \left(\sqrt{a_l^{t+1}b_l^{t+1}}\right)}{K_{\left[\lambda_l\right]^{t+1}} \left(\sqrt{a_l^{t+1} b_l^{t+1}} \right) }.
\end{align}

\noindent \underline{Update the parameters of  $Q(\beta)^{t+1}$:}
\begin{align}
& e^{t+1} = \epsilon + \frac{\prod_{n=1}^N J_n}{2}, \\
& f^{t+1} = \epsilon +   \frac{\mathfrak f^{t+1}}{2}, 
\end{align}
where $\mathfrak f^{t+1}$ is computed using the result in the last row of Table II with  $\{\boldsymbol M^{(n)}, \boldsymbol \Sigma^{(n)} \}$ being replaced by  $\{\left[\boldsymbol M^{(n)}\right]^{t+1}, \left[\boldsymbol \Sigma^{(n)}\right]^{t+1}\}, \forall n$.

\noindent \underline{Update the hyper-parameter  $[a_l^0]^{t+1}$:}
\begin{align}
[a_l^0]^{t+1} = \frac{\kappa_{a_1}+ \frac{\lambda_l^0}{2} -1}{\kappa_{a_2} + \frac{ m[z_l]^{t+1} }{2}}.
\end{align}
\noindent {\bf{Until Convergence}}
\end{algorithm}
\subsubsection{Convergence Property} While the proposed algorithm is devised with the novel probabilistic tensor CPD model using GH prior, its derivation follows the  mean-field VI framework \cite{beal, VI1, VI2, Sergios2}. In particular, in each iteration,  after fixing other variational pdfs, the problem that optimizes a single variational pdf has been shown to be convex and has a unique solution \cite{beal, VI1, VI2, Sergios2}. By treating each update step in mean-field VI as a block coordinate descent (BCD) step over the functional space, the limit point generated by the VI algorithm is at least a stationary point of the KL divergence \cite{beal, VI1, VI2, Sergios2}.

\subsubsection{Automatic Tensor Rank Learning} During the iterations, the mean of parameter $z_l^{-1}$ (denoted by $m[z_l^{-1}]$) will be learnt {\color{black} using the updated parameters of other variational pdfs} as seen in (26)-(29). Due to the sparsity-promoting nature of the GH prior, some of $m[z_l^{-1}]$ will take very large values, e.g., in the order of $10^{6}$. Since  the inverse of $\{m[z_l^{-1}]\}_{l=1}^L$ contribute to the covariance matrix of each factor matrix (as seen in (24)) and then rescale the columns in each factor matrix (as seen in (25)), a very large $m[z_l^{-1}]$ will shrink the $l^{th}$ column of each factor matrix to all zero. Then, by enumerating how many non-zero columns in each factor matrix, the tensor rank can be automatically learnt. 

In practice, to accelerate the learning algorithm,  on-the-fly pruning is widely employed in Bayesian tensor research. In particular, in each iteration, if some of the columns in each factor matrix are found to be indistinguishable from all zeros, it indicates that these columns play no role in interpreting the data, and thus they can be safely pruned. This pruning procedure will not affect the convergence behavior of the algorithm, since each pruning is equivalent to restarting  the algorithm for a reduced probabilistic model  with the current variational pdfs acting as the initializations. {\color{black} More discussions on the pruning issue are provided in Appendix N. }

\subsubsection{Computational Complexity}  For the proposed algorithm, in each iteration, the computational complexity is dominated by updating the factor matrices, costing $O(N \prod_{n=1}^N J_n L^2 + L^3 \sum_{n=1}^N J_n)$. Therefore, the computational complexity of the proposed algorithm is $O(q (N \prod_{n=1}^N J_n L^2 + L^3 \sum_{n=1}^N J_n))$ where $q$ is the iteration number at convergence. The complexity  is comparable to that of the inference algorithm using Gaussian-gamma prior \cite{PI1}.

\section{Numerical Results on Synthetic Data}

In this and the next section, extensive numerical results are presented to assess the performance of the proposed algorithm using  synthetic data and real-world datasets, respectively. For the proposed algorithm, its initialization follows the suggestions given in Appendix D unless stated otherwise. All experiments were conducted in Matlab R2015b with an Intel Core i7 CPU at 2.2 GHz.

\subsection{Simulation Setup}
We consider three-dimensional tensors $\mathcal X =  \llbracket  \boldsymbol A^{(1)},   \boldsymbol A^{(2)},  \boldsymbol A^{(3)} \rrbracket  \in \mathbb R^{30 \times 30 \times 30}$ with different tensor ranks. Each element in the factor matrices $\{ \boldsymbol A^{(n)}\}_{n=1}^3$ is independently drawn from a zero-mean Gaussian distribution with unit power.  The observation model $\mathcal Y = \mathcal X + \mathcal W$, where each element  of the noise tensor $\mathcal W$ is independently drawn from a zero-mean Gaussian distribution with variance $\sigma^2_w$. This data generation process follows that of \cite{PI1}. The SNR is defined as $10 \log_{10} \left( \mathrm{var}\left(\mathcal X \right) / \sigma^2_w\right)$ \cite{PI1, PI2}, where $\mathrm{var}\left(\mathcal X \right) $ is the variance\footnote{\color{black}It means the empirical variance computed by treating all entries of the tensor as independent realizations of a same scalar random variable.} of the noise-free tensor $\mathcal X$.   All simulation results in this section are obtained by averaging 100 Monte-Carlo runs unless stated otherwise.

\subsection{PCPD-GH versus PCPD-GG} Since the principle of this paper follows a parametric way of Bayesian modeling, we first compare the proposed algorithm using GH prior (labeled as PCPD-GH) with the benchmarking algorithm \cite{PI1} (labeled as PCPD-GG). 

\begin{figure} [!t]
\setcounter{subfigure}{0}
\centering
\subfigure[] {
\includegraphics[width= 3.2in]{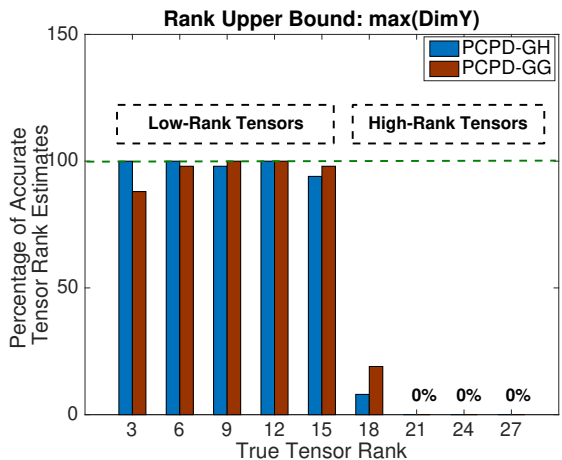}
}
\subfigure[] {
\includegraphics[width=3.2 in]{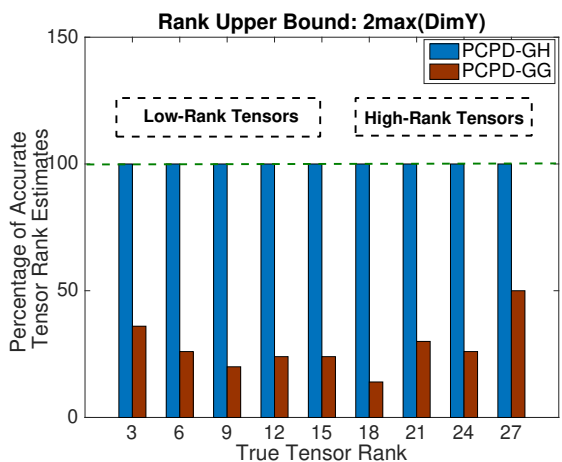}
}
\subfigure[] {
\includegraphics[width=3.2 in]{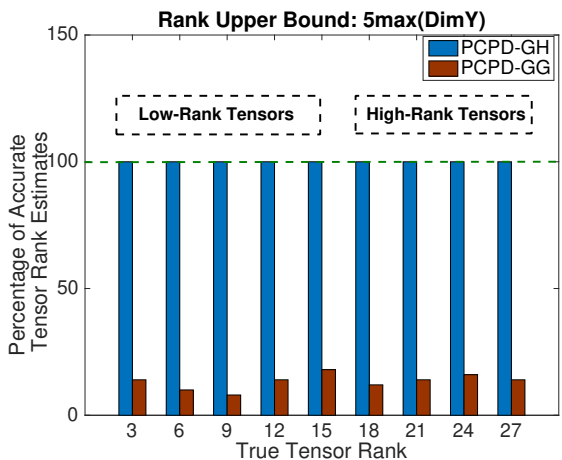}
}
\caption{Performance of tensor rank learning when the rank upper bound is (a) $ \max \{J_n\}_{n=1}^N$, (b) $2 \max \{J_n\}_{n=1}^N$ and (c) $5\max \{J_n\}_{n=1}^N$. }
\label{fig}
\end{figure}

\subsubsection{Tensor rank learning} The performance of tensor rank learning is firstly evaluated. We regard the tensors as low-rank tensors when their ranks are smaller than or equal to  half of the maximal tensor dimension, i.e., $R \leq \max \{J_n\}_{n=1}^N/2$. Similarly, high-rank tensors are those with $R > \max \{J_n\}_{n=1}^N /2$. In particular, in Figure 8, we assess the tensor rank learning performances of the two algorithms for low-rank tensors with $R = \{3, 6, 9, 12, 15 \}$ and high-rank tensors with $R = \{18, 21, 24, 27 \}$ under SNR = 10 dB.  In Figure 8 (a), the two algorithms are both with the tensor rank upper bound $\max\{ J_n \}_{n=1}^N$. It can be seen that the PCPD-GH algorithm and the PCPD-GG algorithm {\color{black} achieve}  comparable performances in learning low tensor ranks. More specifically, the PCPD-GH algorithm achieves higher learning accuracies when $R = \{ 3, 6\}$ while the PCPD-GG method performs better when $R = \{ 9, 15\}$.  However, when tackling high-rank tensors with $R > 15$,  as seen in Figure 8 (a),  both algorithms with tensor rank upper bound $\max\{ J_n \}_{n=1}^N$ fail to work properly. The reason is that the upper bound value $\max\{ J_n \}_{n=1}^N$ results in too small  sparsity level $\frac{L-R}{L}$ to leverage the power of the sparsity-promoting priors in tensor rank learning. Therefore, the upper bound value should be set larger in case that the tensor rank is high.  An immediate choice is $f \times \max\{J_n\}_{n=1}^N$ where $f = 1,2,3,\cdots$. In Figure 8 (b) and (c), we assess the performances of tensor rank learning for the two methods using the upper bound $2 \max \{J_n\}_{n=1}^N$ and $5 \max\{J_n\}_{n=1}^N$, respectively. It can be seen that the PCPD-GG algorithm is very sensitive to the rank upper bound value, in the sense that its performance deteriorates significantly for low-rank tensors after employing the larger upper bounds. While PCPD-GG has an improved performance for high-rank tensors after adopting a larger upper bound, the chance of getting the correct rank is still very low. In contrast, the performance of the proposed PCPD-GH algorithm is stable for all cases and it achieves nearly $100 \%$ accuracies of tensor rank learning in a wide range of scenarios, showing its flexibility in adapting to different levels of sparsity. {\color{black} In Appendix L, further numerical results on the tensor rank learning accuracies versus different sparsity levels are presented, which show the better performance of the proposed algorithm in a wide range of sparsity levels.}

\begin{figure} [!t]
\setcounter{subfigure}{0}
\centering
\subfigure[] {
\includegraphics[width=3.2 in]{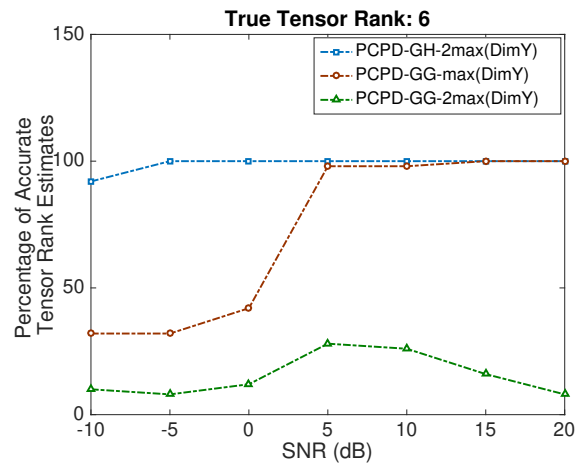}
}
\subfigure[] {
\includegraphics[width=3.2 in]{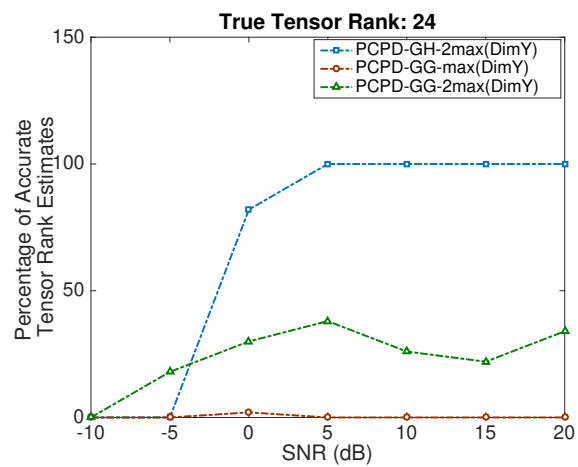}
}
\caption{Performance of tensor rank learning versus different SNRs: (a) low-rank tensors and (b) high-rank tensors. }
\label{fig}
\end{figure}

To assess the tensor rank learning performance under different SNRs, in Figure 9, the percentages of accurate tensor rank learning from the two methods are presented. We consider two scenarios: 1) low-rank tensor with $R = 6$ shown in Figure 9 (a) and 2) high-rank tensor with $R=24$ shown in Figure 9 (b). For the proposed PCPD-GH algorithm, due to its robustness to different rank upper bounds, $2 \max \{J_n\}_{n=1}^N$ is adopted as the upper bound value (labeled as PCPD-GH-2max(DimY)). For the PCPD-GG algorithm, both the upper bound value  $\max \{ \{ J_n \}_{n=1}^N\}$ and $2 \max \{J_n\}_{n=1}^N$ are considered (labeled as PCPD-GG-max(DimY) and  PCPD-GH-2max(DimY) respectively). From Figure 9, it is clear that the performance of the PCPD-GG method, for all cases, highly relies on the choice of the rank upper bound value. In particular, when adopting $2 \max \{J_n\}_{n=1}^N$, its performance in tensor rank learning is not good (i.e., below than $50\%$) for both the low-rank tensor and the high-rank tensor cases. In contrast, when adopting $\max \{ \{ J_n \}_{n=1}^N\}$, its performance becomes much better for the low-rank cases. In Figure 9 (a), when SNR is larger than 5 dB, the PCPD-GG with upper bound value  $\max \{ \{ J_n \}_{n=1}^N\}$ achieves nearly $100 \%$ accuracy, which is very close to the accuracies of the PCPD-GH method. However, when the SNR is smaller than 5 dB, although the PCPD-GH method still achieves nearly $100 \%$ accuracies in tensor rank learning, the accuracies of the PCPD-GG method fall below $50 \%$. For the high-rank case, as seen in Figure 9 (b), both the PCPD-GH and the PCPD-GG methods fail to recover the true tensor rank when SNR is smaller than 0 dB. However, when the SNR is larger than 0 dB, the accuracies of the PCPD-GH method are near $100\%$ while those of the PCPD-GG at most achieve about $50 \%$ accuracy. Consequently, it can be concluded from Figure 9 that the proposed PCPD-GH method achieves more stable and accurate tensor rank learning. 

In summary, Figures 8 and 9 show that the proposed method finds the correct tensor rank even if the initial tensor rank is {\it exceedingly over-estimated}, which is a {\it practically useful} feature since the rank is unknown in real-life cases.

\begin{figure} [!t]
\setcounter{subfigure}{0}
\centering
\subfigure[] {
\includegraphics[width=3.2 in, height = 2.2 in]{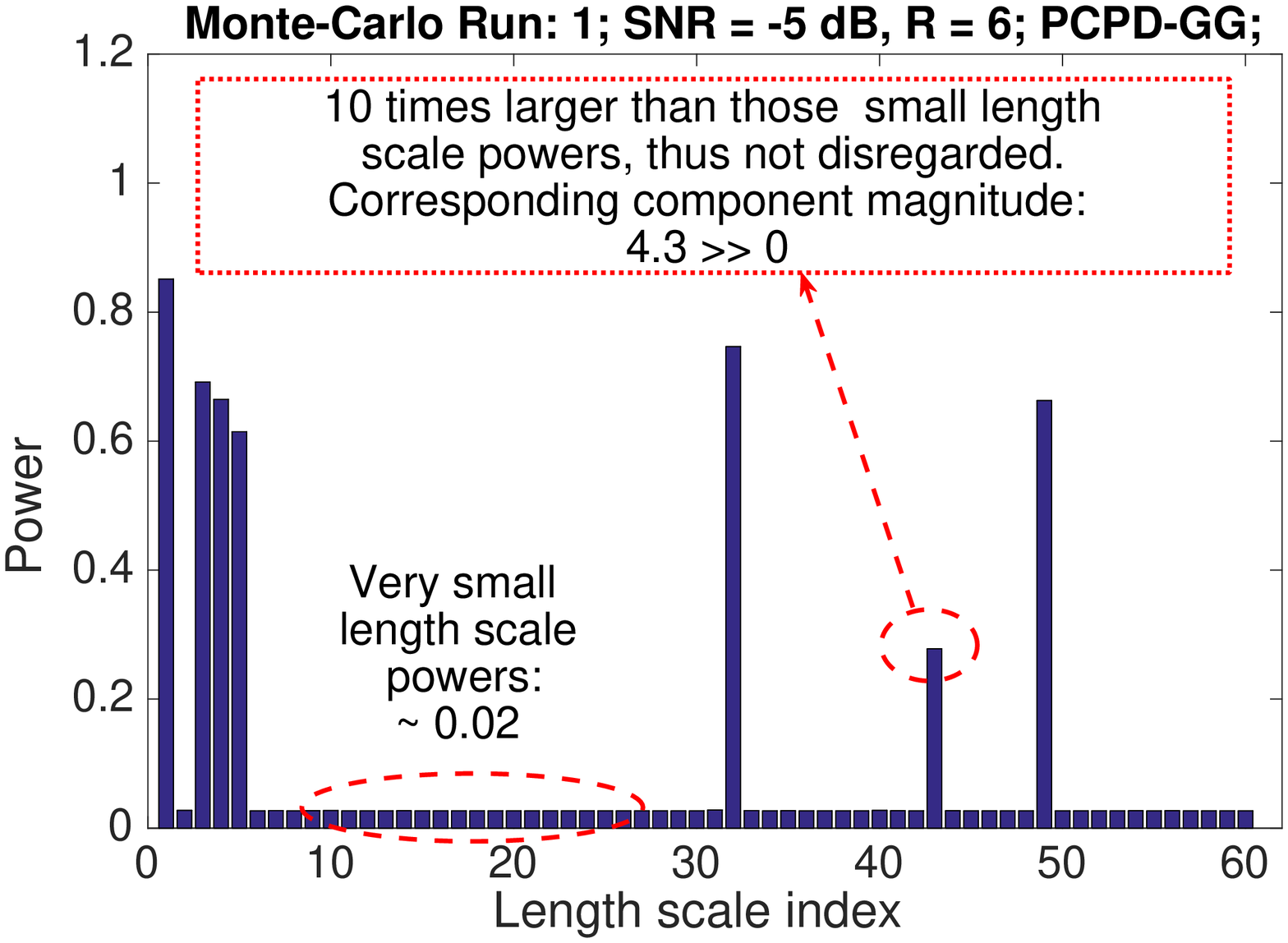}
}
\subfigure[] {
\includegraphics[width=3.2 in, height = 2.2 in]{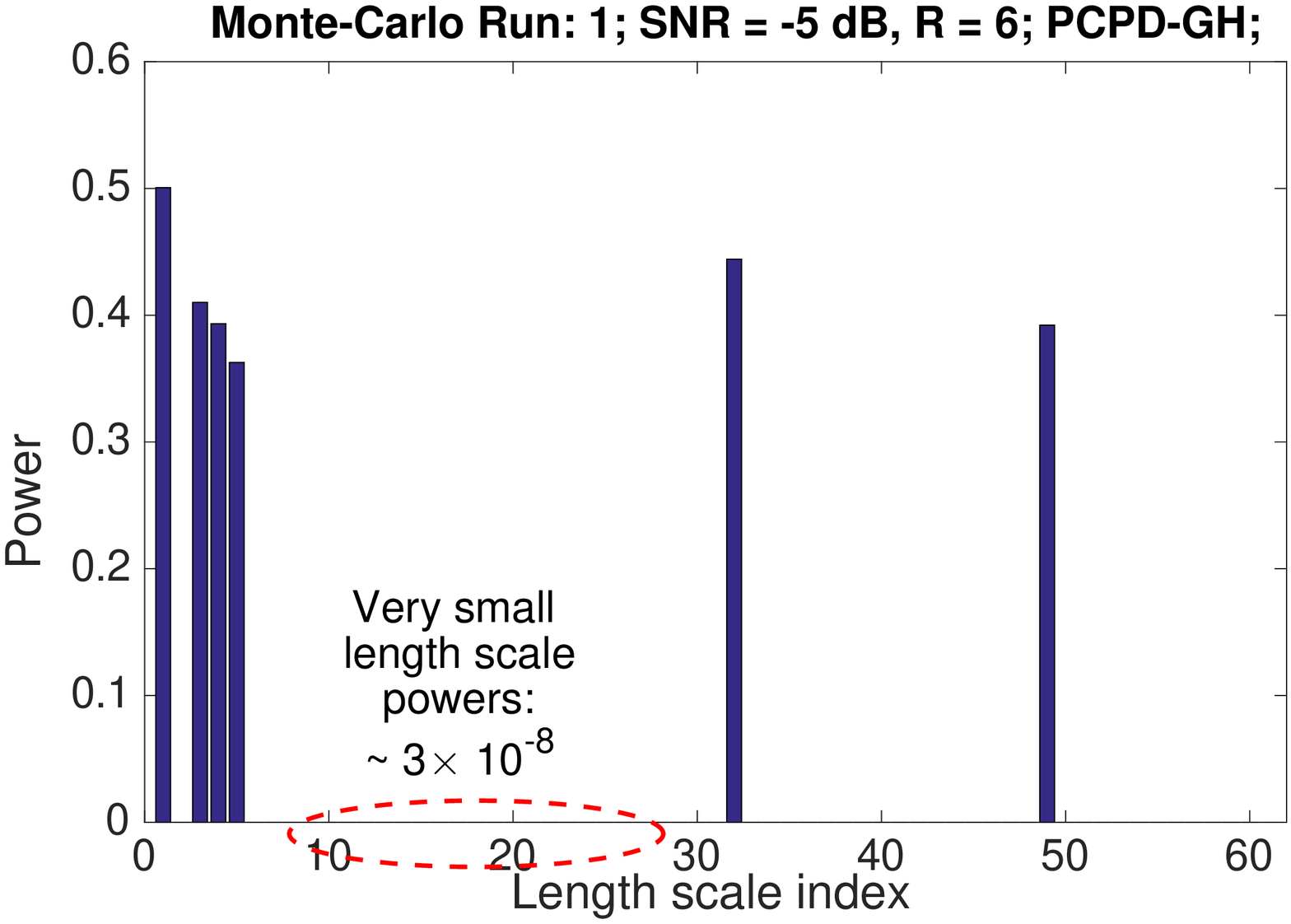}
}
\caption{(a) The powers of learnt length scales (i.e., $\{ \gamma_l^{-1}\}_l$) for PCPD-GG;  (b) The powers of learnt length scales (i.e., $\{ z_l\}_l$) for PCPD-GH.  It can be seen that  PCPD-GG recovers 7 components with non-negligible magnitudes, while  PCPD-GH recovers 6 components. The two algorithms are with the same upper bound value: 60.}
\end{figure}

%; ii) Case II: SNR = 5 dB, R = 21, corresponding to high SNR and high rank.

\subsubsection{Insights from learnt length scales} To clearly show the substantial  difference between the GG  and GH prior, we compare the two algorithms (PCPD-GG and PCPD-GH) in terms of their learnt length scales.  The length scale powers of GG and GH prior are denoted by $\left\{ \gamma_l^{-1} \right \}_l$ and $\left \{ z_l  \right\}_l$ respectively. To assess the patterns of learnt length scales, we turn off the pruning procedure and let the two algorithms directly output $\{\gamma_l^{-1} \}_l$ and $\{ z_l \}_l$ after convergence. Since in the different Monte-Carlo trial,  the learnt length scale powers are possibly of different sparsity patterns, averaging them over Monte-Carlo trials is not informative. Instead, we present the learnt values of $\{\gamma_l^{-1} \}_l$ and $\{ z_l \}_l$ in a single trial.

In particular, Figure 10 shows the result for a typical low SNR and low rank case (SNR= -5 dB, R = 6) with rank upper bound being 60.  From this figure,  it can be seen that the learnt length scales of the two algorithms substantially {\it differ} from each other,  in the sense that the number of learnt length scales (and the associated components) {\it with non-negligible magnitudes} are different. For example, in Figure 10,  PCPD-GG recovers 7 components with non-negligible magnitudes\footnote{ The magnitude of the $l$-th component is defined as $\left(\sum_{n=1}^3 \left[ \boldsymbol A^{(n)}_{:, l} \right]^T \boldsymbol A^{(n)}_{:, l} \right)^{\frac{1}{2}}$ \cite{PI1}.} (the smallest one has value $4.3 \gg 0$), while  PCPD-GH recovers 6 components.  Note that the ground-truth rank is 6, and PCPD-GG produces a ``ghost'' component {\it with magnitude  much larger than zero}.   Additional simulation runs, and results of other simulation settings (e.g., high SNR and high rank case: SNR = 5 dB, R = 21) are included in Appendix P, from which similar conclusions can be drawn.

%Similarly,  in Figure P3,  PCPD-GG recovers 22 components with non-negligible magnitudes  (the smallest one has value $3.9 \gg 0$), while PCPD-GH recovers 21 components. The ground-truth rank is 21, and thus PCPD-GG gives one surplus component estimate, whose magnitude is much larger than zero. 

%\begin{figure*} [!t]
%\setcounter{subfigure}{0}
%\centering
%\subfigure[] {
%\includegraphics[width=4 in]{mont_1_gg_6.eps}
%}
%\subfigure[] {
%\includegraphics[width=4 in]{mont_1_gh_6.eps}
%}
%
%\caption{(a) The powers of learnt length scales (i.e., $\{ \gamma_l^{-1}\}_l$) and the magnitudes of associated components for PCPD-GG;  (b) The powers of learnt length scales (i.e., $\{ z_l\}_l$) and the magnitudes of associated components for PCPD-GH.  It can be seen that  PCPD-GG recovers 7 components with non-negligible magnitudes, while  PCPD-GH recovers 6 components. The two algorithms are with the same upper bound value: 60.  SNR = -5 dB, R = 6; Monte-Carlo Run: 1. }
%\end{figure*}
%{\color{blue}

%\begin{figure*} [!t]
%\setcounter{subfigure}{0}
%\centering
%\subfigure[] {
%\includegraphics[width=6 in]{mont_1_gg_6.eps}
%}
%\subfigure[] {
%\includegraphics[width=6 in]{mont_1_gh_6.eps}
%}

\begin{figure} [!t]
\centering
\includegraphics[width=3.2 in]{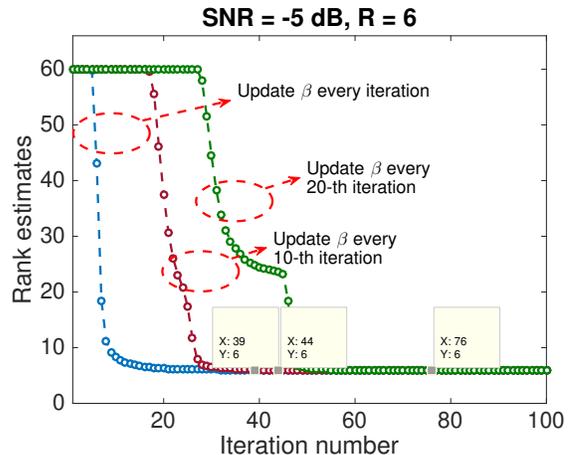}
\caption{Tensor rank estimates of PCPD-GH versus iteration number (averaged over 100 Monte-Caro runs) with different noise precision learning speeds.}
\end{figure}

\subsubsection{Insights on noise precision learning} The learning of the noise precision $\beta$ is crucial for reliable  inference, since  incorrect estimates will cause over-/under-regularization. To examine how the speed of noise learning affects the tensor rank (sparsity) learning when SNR is low (SNR = -5 dB), we turn on the pruning and present the rank learning results over iterations in three cases: i) Case I: update $\beta$ every iteration; ii) Case II: update $\beta$ every 10-th iteration; iii) Case III: update $\beta$ every 20-th iteration. In Figure 11, the rank estimates are averaged over 100 Monte-Carlo runs. It can be seen that updating the noise precision $\beta$ at earlier iterations will help will help the learning process to unveil the sparsity pattern more quickly. 

Then, we investigate under which scenario slowing the noise precision learning will be helpful. We consider a very low SNR case, that is, SNR = -10 dB, and then evaluate the percentages of accurate rank learning over 100 Monte-Carlo runs. The results are: i) Case I: 76\%; ii) Case II: 100 \%; iii) Case III: 100 \%. In other words, when the noise power is very large (e.g., SNR = -10 dB), slowing the noise precision learning will make the algorithm more robust to the noises.

 Finally, if we fix the noise precision $\beta$ and do not allow its update, PCPD-GH fails to identify the underlying sparsity pattern (tensor rank), see Figure Q1 in Appendix Q. Particularly, a small value of $\beta$ (e.g., 0.01)  leads to over-regularization, thus causing under-estimation  of non-zero components; a large value of $\beta$ (e.g., 100) causes under-regularization, thus inducing over-estimation of non-zero components. This shows the importance of modeling and updating of noise precision.

\subsubsection{Other performance metrics}  Additional results and discussions on the run time, tensor recovery root mean square error (RMSE),  algorithm performance under factor matrix correlation, convergence behavior of the proposed algorithm in terms of evidence lower bound (ELBO) [51], and hyper-parameter learning of PCPD-GG are included in Appendix E, F, and G, J, R, respectively. The {\it key messages} of these simulation results are given as follows: i) PCPD-GH generally costs more run time than PCPD-GG; ii) Incorrect estimation of tensor rank degrades the tensor signal recovery; iii) PCPD-GH performs well under factor matrix correlation;  iv) PCPD-GH monotonically increases the ELBO; v) The update of hyper-parameter of PCPD-GH does not help too much in improving rank estimation in the low SNR regime.

\subsection{Comparisons with Nonparametric PCPD-MGP} After comparing to the  parametric PCPD-GG,  further comparisons are performed with the non-parametric Bayesian tensor CPD using MGP prior (labeled as PCPD-MGP)\footnote{We appreciate Prof. Piyush Rai for sharing the code and data with us.}. We implement PCPD-MGP in the variational inference framework. The initializations and hyper-parameters follow those used in \cite{new1}.

\subsubsection{Tensor rank learning}    We first assess the performance of tensor rank learning over 100 Monte-Carlo
trials, to facilitate the comparisons among the three algorithms (PCPD-MGP, PCPD-GG, PCPD-GH). The simulation settings follow those of Figure 8 and Figure 9.

% \usepackage{multirow}
\begin{table}[!t]
\centering
\caption{Performance of tensor rank learning under different rank upper bound values. Algorithm: PCPD-MGP; SNR = 10 dB. The simulation settings are the same as those of Figure 8.}
\scalebox{1}{\begin{tabular}{|c|cccccc|}
\hline
\multirow{2}{*}{\begin{tabular}[c]{@{}c@{}}Rank \\ Upper Bound\end{tabular}} &
  \multicolumn{6}{c|}{True  Tensor Rank R} \\ \cline{2-7} 
 &
  \multicolumn{1}{c|}{3} &
  \multicolumn{1}{c|}{6} &
  \multicolumn{1}{c|}{9} &
  \multicolumn{1}{c|}{12} &
  \multicolumn{1}{c|}{15} &
  \multicolumn{1}{c|}{18-27} 
\\ \hline
max(DimY) &
  \multicolumn{1}{c|}{\bf 100\%} &
  \multicolumn{1}{c|}{\bf 100\%} &
  \multicolumn{1}{c|}{\bf 100\%} &
  \multicolumn{1}{c|}{98\%} &
  \multicolumn{1}{c|}{90\%} &
  \multicolumn{1}{c|}{0\%} 
 \\ \hline
2max(DimY) &
  \multicolumn{1}{c|}{\bf 100\%} &
  \multicolumn{1}{c|}{\bf 100\%} &
  \multicolumn{1}{c|}{\bf 100\%} &
  \multicolumn{1}{c|}{94\%} &
  \multicolumn{1}{c|}{64\%} &
  \multicolumn{1}{c|}{ 0\%} \\ \hline
5max(DimY) &
  \multicolumn{1}{c|}{\bf 100\%} &
  \multicolumn{1}{c|}{\bf 100\%} &
  \multicolumn{1}{c|}{\bf 100\%} &
  \multicolumn{1}{c|}{26\%} &
  \multicolumn{1}{c|}{24\%} &
  \multicolumn{1}{c|}{0\%} 
 \\ \hline
\end{tabular}}
\end{table}

\begin{table}[!t]
\centering
\caption{Performance of tensor rank learning under different SNRs. Algorithm: PCPD-MGP; Upper bound value: 2max(DimY).  The simulation settings are the same as those of Figure 9.}
\scalebox{1}{\begin{tabular}{|c|ccccccc|}
\hline
\multirow{2}{*}{\begin{tabular}[c]{@{}c@{}}True\\ Tensor Rank\end{tabular}} & \multicolumn{7}{c|}{SNR (dB)} \\ \cline{2-8} 
 &
  \multicolumn{1}{c|}{-10} &
  \multicolumn{1}{c|}{-5} &
  \multicolumn{1}{c|}{0} &
  \multicolumn{1}{c|}{5} &
  \multicolumn{1}{c|}{10} &
  \multicolumn{1}{c|}{15} &
  20 \\ \hline
R = 6 &
  \multicolumn{1}{c|}{2\%} &
  \multicolumn{1}{c|}{78\%} &
  \multicolumn{1}{c|}{90\%} &
  \multicolumn{1}{c|}{98\%} &
  \multicolumn{1}{c|}{\bf 100\%} &
  \multicolumn{1}{c|}{\bf 100\%} &
\bf  100\% \\ \hline
R = 24 &
  \multicolumn{1}{c|}{0\%} &
  \multicolumn{1}{c|}{0\%} &
  \multicolumn{1}{c|}{0\%} &
  \multicolumn{1}{c|}{0\%} &
  \multicolumn{1}{c|}{0\%} &
  \multicolumn{1}{c|}{0\%} &
  0\% \\ \hline
\end{tabular}}
\end{table}

In Table III, the percentages of accurate tensor rank estimates of PCPD-MGP under different rank upper bound values are presented.
 Comparing to Figure 8, we can draw the following conclusions. i) When the tensor rank is low (e.g., R = 3, 6, 9) and the SNR is high (e.g., SNR = 10 dB), PCPD-MGP correctly learns the tensor rank over 100 Monte-Carlo trials under different rank upper bound values. Its performance is insensitive to the selection of the rank upper bound values, due to the decaying effects of the learnt length scales \cite{new1}. Therefore, in the low-rank and high-SNR scenario, the rank learning performance of PCPD-MGP is comparable to that of PCPD-GH, and much better than PCPD-GG (see Figure 8). ii) When the tensor rank is high (e.g., R = 18, 21, 24, 27),  PCPD-MGP fails to learn the correct tensor rank under different rank upper bound values, since the decaying effects of learnt length scales tend to {\it under-estimate} the tensor rank. On the contrary, the PCPD-GH method always accurately estimates the high tensor ranks (see Figure 8).

In Table IV, we present the rank estimation performance of PCPD-MGP under different SNRs, with the same settings as those of Figure 9. It can be seen that when the tensor rank is low (e.g., R = 6),  PCPD-MGP shows good performance when SNR is larger than -5 dB. It outperforms the PCPD-GG method, but it is still not as good as the proposed PCPD-GH.  At SNR = -10 dB, PCPD-MGP fails to correctly estimate the tensor rank, while PCPD-GH shows good performance in a wider range of SNRs (from -10 dB to 20 dB). Furthermore, when the tensor rank becomes large (e.g., R = 24), PCPD-MGP fails to learn the underlying true tensor rank,   making it inferior to PCPD-GH (and even PCPD-GG) in the high-rank regime.

\subsubsection{Insights from learnt length scales} To reveal more insights, we present the learnt  length scales (without pruning) of PCPD-MPG. Following the notations in \cite{new1}, the learnt length scale powers of PCPD-MGP are denoted by $\left\{ \tau_l^{-1} \right\}_l$.

\begin{figure} [!t]
\setcounter{subfigure}{0}
\centering
\subfigure[] {
\includegraphics[width=3.2 in, height = 2.2 in]{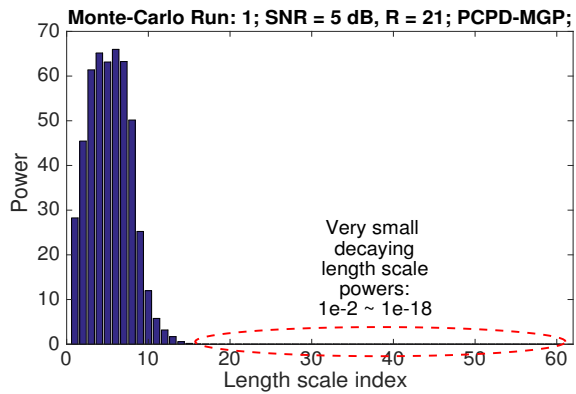}
}
\subfigure[] {
\includegraphics[width=3.2 in, height = 2.2 in]{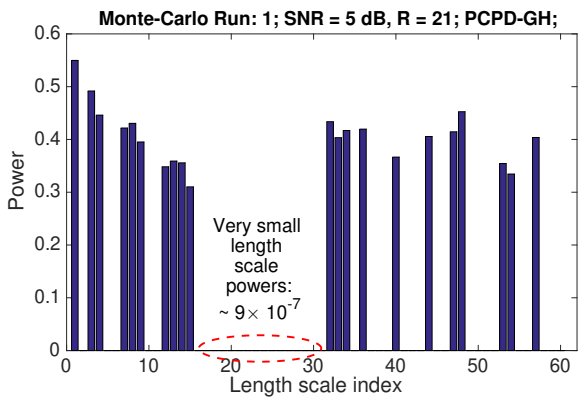}
}
\caption{(a) The powers of learnt length scales (i.e., $\{ \tau_l^{-1}\}_l$)  for PCPD-MGP;  (b) The powers  of learnt length scales (i.e., $\{ z_l\}_l$)   for PCPD-GH.  It can be seen that  PCPD-MGP recovers 15 components with non-negligible magnitudes, while PCPD-GH recovers 21 components. The two algorithms are with the same upper bound value: 60.  SNR = 5 dB, R = 21.}
\end{figure}

\begin{figure} [!t]
\setcounter{subfigure}{0}
\centering
\subfigure[] {
\includegraphics[width=3.2 in, height = 2.2 in]{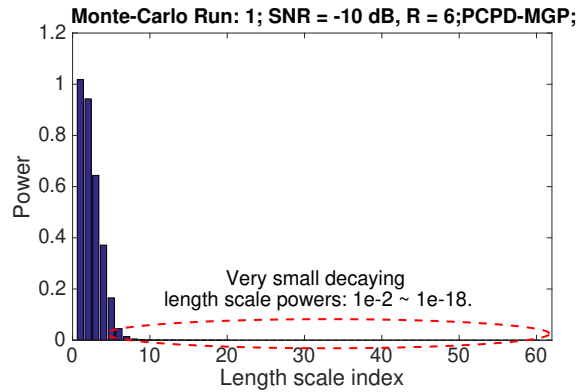}
}
\subfigure[] {
\includegraphics[width=3.2 in, height = 2.2 in]{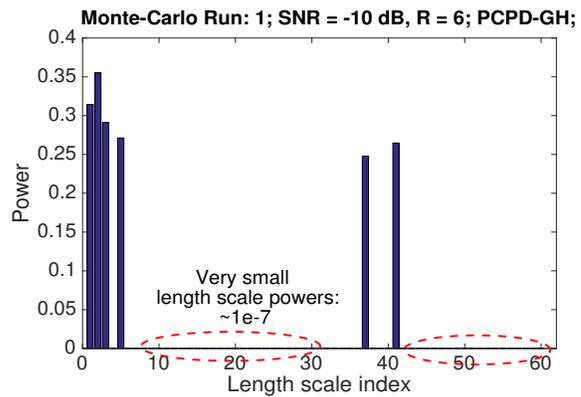}
}
\caption{(a) The powers of learnt length scales (i.e., $\{ \tau_l^{-1}\}_l$)  for PCPD-MGP;  (b) The powers  of learnt length scales (i.e., $\{ z_l\}_l$)   for PCPD-GH.  It can be seen that  PCPD-MGP recovers 5 components with non-negligible magnitudes, while PCPD-GH recovers 6 components. The two algorithms are with the same upper bound value: 60.  SNR = -10 dB, R = 6.}
\end{figure}

Two  typical cases are considered:  i) Case I: SNR = 5 dB, R = 21, corresponding to high rank and high SNR;  ii) Case II: SNR = -10 dB, R = 6, corresponding to very low SNR and low rank. For each case, we present the learnt length scales in a single trial in Figure 12 and Figure 13 respectively. From these figures, we have the following observations. i) Due to the {\it decaying effect} of the MGP prior, the learnt length scale  power $\tau_l^{-1}$  quickly decreases as $l$ becomes larger. This drives PCPD-MGP to {\it fail} to recover the sparsity-pattern of the high-rank CPD (e.g., R = 21), in which a large number of length scale powers should be much larger than zero, see Figure 12 (a). In contrast, without the decaying effect, the proposed PCPD-GH successfully identifies the 21 non-negligible components, as seen in Figure 12 (b). ii) When the SNR is very low (e.g., SNR = -10 dB) and the rank is low (e.g., R = 6), the sparsity pattern of  length scales learnt by PCPD-MGP is {\it not as accurate as} that obtained by PCPD-GH (see Figure 13). Additional simulation runs, and results of more simulation settings are included in Appendix S, from which similar conclusions can be drawn.

\section{(Semi-) Real-world Dataset Experiments}
 In this section, experimental results on several (semi-) real-world datasets are presented.  Since the scope of this paper is to investigate flexible sparsity-aware modeling for tensor CPD in the parametric Gaussian scale mixture family,  the comparisons are performed between PCPD-GG and PCPD-GH.

\subsection{Semi-real Experiment: Fluorescence Data Analytics}

Tensor CPD is an important tool in fluorescence data analytics, with the aim to reveal the underlying signal components. We consider the popular amino acids fluorescence data\footnote{http://www.models.life.ku.dk} $\mathcal X$ with size $5 \times 201 \times 61$ \cite{C1}, which consists of five laboratory-made samples. Each sample contains different amounts of tyrosine, tryptophan and phenylalanine dissolved in phosphate buffered water. Since there are three different types of amino acid, when adopting the CPD model, the optimal tensor rank should be 3. In particular, with the optimal tensor rank $3$, the clean spectra for the three types of amino acid, which are recovered by the  the alternative least-squares (ALS) algorithm \cite{C1}, are presented in Figure H1 (in Appendix H) as the benchmark. 

% \usepackage{multirow}
\begin{table}[!t]
\centering
\caption{Fit values and estimated tensor ranks of fluorescence data under different SNRs (with rank upper bound $\max\{J_n\}_{n=1}^N$).}
\resizebox{0.45\textwidth}{!}{\begin{tabular}{@{}|c|c|c|c|c|@{}}
\toprule
\multirow{2}{*}{\begin{tabular}[c]{@{}c@{}}SNR\\ (dB)\end{tabular}} & \multicolumn{2}{c|}{PCPD-GG}                                                 & \multicolumn{2}{c|}{PCPD-GH}                                                        \\ \cline{2-5} 
                                                                    & Fit Value & \begin{tabular}[c]{@{}c@{}}Estimated \\ Tensor Rank\end{tabular} & Fit Value        & \begin{tabular}[c]{@{}c@{}}Estimated \\ Tensor Rank\end{tabular} \\ \midrule
-10                                                                 & 71.8109   & 4                                                                & \textbf{72.6401} & \textbf{3}                                                       \\ \midrule
-5                                                                  & 83.9269   & 4                                                                & \textbf{84.3424} & \textbf{3}                                                       \\ \midrule
0                                                                   & 90.6007   & 4                                                                & \textbf{90.8433} & \textbf{3}                                                       \\ \midrule
5                                                                   & 94.2554   & 4                                                                & \textbf{94.3554} & \textbf{3}                                                       \\ \midrule
10                                                                  & 96.0907   & \textbf{3}                                                       & 96.0951          & \textbf{3}                                                       \\ \midrule
15                                                                  & 96.8412   & \textbf{3}                                                       & 96.8431           & \textbf{3}                                                       \\ \midrule
20                                                                  & 97.1197   & \textbf{3}                                                       & 97.1204          & \textbf{3}                                                       \\ \bottomrule
\end{tabular}}
\end{table}

In practice, it is impossible to know how many components are present in the data in advance, and this calls for automatic tensor rank learning. In this subsection, we assess both the rank learning performance and the noise mitigation performance for the two algorithms (i.e., PCPD-GH and PCPD-GG) under different levels of noise sources. In particular, the Fit value \cite{app5}, which is defined as $(1 - \frac{|| \hat{\mathcal X} - \mathcal X ||_F}{ ||\mathcal X||_F}) \times 100\%$, is adopted, where  $\hat{\mathcal X}$ represents the reconstructed fluorescence tensor data from the algorithm. In Table V and H1 (see Appendix H), the performances of the two algorithms are presented assuming different upper bound values of tensor rank. It can be observed that with different upper bound values, the proposed PCPD-GH algorithm always gives the correct tensor rank estimates, even when the SNR is smaller than 0 dB. On the other hand, the PCPD-GG method is quite sensitive to the choice of the upper bound value. Its performance with upper bound $2\max\{J_n\}_{n=1}^N$ becomes much worse than that with  $\max\{J_n\}_{n=1}^N$ in tensor rank learning. Even with the upper bound being equal to $\max\{J_n\}_{n=1}^N$, PCD-GG fails to recover the optimal tensor rank $3$ in the low SNR region (i.e., SNR $\leq$ 5 dB). With the over-estimated tensor rank, the reconstructed fluorescence tensor data $\hat{\mathcal X}$ will be overfitted to the noise sources, leading to lower Fit values. As a result, the Fit values of the proposed method are generally higher than those of the PCPD-GG method under different SNRs. 

In this application, since the tensor rank denotes the number of underlying components inside the data,  its incorrect estimation will not only lead to overfitting to the noise, but also will cause ``ghost" components that cannot be interpreted. The figures for recovered spectra and the ``ghost'' component under incorrect rank estimation are given in Appendix H.  In particular, in Figure H3 (of Appendix H), we present the  learnt length scale powers and the   associated component magnitudes from PCPD-GG and PCPD-GH. It can be seen that PCPD-GG recovers four components with non-negligible magnitudes. The smallest magnitude of the learnt four components is $10.6$, which is much larger than zero. In practical data analysis,  disregarding a learnt latent component with magnitude $10.6$ is not reasonable. 
Since the ``ghost'' component has a relatively large magnitude, it degrades the performance of the tensor signal recovery. From Table V, when the SNR is low (e.g.,  -10 dB), the Fit value of PCPD-GH is higher by 0.8, compared to PCPD-GG. Note that in the high SNR regime (no ``ghost'' component), their Fit value difference is about 0.001. Therefore, the ``ghost'' component largely degrades the tensor signal recovery.

\begin{table*}[!t]
\centering 
\caption{Tensor rank learning for CPD of Tongue dataset and ENRON Email dataset.}
\scalebox{0.85}{\begin{tabular}{@{}|c|c|c|c|c|c|c|c|c|c|@{}}
\toprule
\multirow{2}{*}{Dataset} & \multirow{2}{*}{\begin{tabular}[c]{@{}c@{}}Benchmarking \\ Tensor \\ Rank\end{tabular}} & \multicolumn{4}{c|}{PCPD-GG}                                                                                                                                                                                                                     & \multicolumn{4}{c|}{PCPD-GH}                                                                                                                                                                                                                     \\ \cmidrule(l){3-10} 
                         &                                                                                         & \multicolumn{2}{c|}{\begin{tabular}[c]{@{}c@{}}max(DimY):\\  13\end{tabular}}                                          & \multicolumn{2}{c|}{\begin{tabular}[c]{@{}c@{}}2max(DimY): \\ 26\end{tabular}}                                          & \multicolumn{2}{c|}{\begin{tabular}[c]{@{}c@{}}max(DimY): \\ 13\end{tabular}}                                          & \multicolumn{2}{c|}{\begin{tabular}[c]{@{}c@{}}2max(DimY): \\ 26\end{tabular}}                                          \\ \midrule
Tongue                   & 2                                                                                       & \multicolumn{2}{c|}{2}                                                                                                 & \multicolumn{2}{c|}{3}                                                                                                  & 2                                                        & 2                                                           & 2                                                           & 2                                                         \\ \midrule
\multirow{2}{*}{ENRON}   & 5                                                                                       & \begin{tabular}[c]{@{}c@{}}max(DimY):\\ 184\end{tabular} & \begin{tabular}[c]{@{}c@{}}max(DimY)+20:\\ 204\end{tabular} & \begin{tabular}[c]{@{}c@{}}max(DimY)+40:\\ 224\end{tabular} & \begin{tabular}[c]{@{}c@{}}2max(DimY):\\ 368\end{tabular} & \begin{tabular}[c]{@{}c@{}}max(DimY):\\ 184\end{tabular} & \begin{tabular}[c]{@{}c@{}}max(DimY)+20:\\ 204\end{tabular} & \begin{tabular}[c]{@{}c@{}}max(DimY)+40:\\ 224\end{tabular} & \begin{tabular}[c]{@{}c@{}}2max(DimY):\\ 368\end{tabular} \\ \cmidrule(l){2-10} 
                         & 5                                                                                       & 5                                                        & 5                                                           & 6                                                           & 11                                                        & 5                                                        & 5                                                           & 5                                                           & 9                                                         \\ \bottomrule
\end{tabular}}
\end{table*}

\subsection{Real-world Dataset Experiments}

In this subsection, four real-world datasets, namely the ENRON Email dataset\footnote{The original source of the data is from \cite{Enron}, and we greatly appreciate Prof. Vagelis Papalexakis for sharing the data with us.}, the Tongue dataset \cite{Tounge},  the Salinas-A  hyperspectral image (HSI) dataset\footnote{http://www.ehu.eus/ccwintco/index.php/Hyperspectral\_Remote\_Sensing\_ \\ Scenes\#Salinas-A\_scene.}, and the Indian Pines HSI dataset\footnote{http://www.ehu.eus/ccwintco/index.php/Hyperspectral\_Remote\_Sensing \\ \_Scenes\#Indian\_Pines.} are used for evaluation. The  ENRON Email dataset   (with size $184 \times 184 \times 44$) contains e-mail communication records between 184 people within 44 months, and the Tongue dataset  (with size $5 \times 10 \times 13$) was obtained from X-rays  of five different speakers during their pronunciation of the vowels. The Salinas-A dataset (with size $83 \times 86 \times 204$) was collected by the AVIRIS\footnote{An acronym for the Airborne Visible InfraRed Imaging Spectrometer.} sensor over Salinas Valley, California; and the Indian Pines dataset   (with size $145 \times 145 \times 200$) was collected by AVIRIS sensor over the Indian Pines test site in North-western Indiana.

For the ENRON Email dataset and the Tongue dataset, it was reported in previous research, see, e.g., \cite{papa1}, \cite{Tounge}, that interpretable decomposition results can be obtained in the context of each data analysis task, when the tensor rank is $5$ and $2$, respectively. In particular, rank-5 tensor CPD can give interpretable clustering results for the ENRON Email dataset \cite{papa1}; and  rank-2 tensor CPD is good for extracting the two underlying components from  the Tongue dataset \cite{Tounge}. Using these as benchmarking tensor ranks, the performance of the  PCPD-GH method and the PCPD-GG method was reported in Table VI, from which it can be observed that the proposed PCPD-GH method is more robust to different  tensor rank upper bounds, and outperforms the PCPD-GG method, in the sense that the estimated  tensor ranks of PCPD-GH match benchmarking results in more scenarios than PCPD-GG.

\begin{table}[!t]
\centering
\caption{SNR outputs and estimated tensor ranks of HSI data under different rank upper bounds.}
\resizebox{0.48\textwidth}{!}{
\begin{tabular}{@{}|c|c|c|c|c|c|@{}}
\toprule
\multicolumn{2}{|c|}{Algorithm}                                                       & \multicolumn{2}{c|}{PCPD-GG}                                                                                          & \multicolumn{2}{c|}{PCPD-GH}                                                                                         \\ \midrule
Dataset                       & \begin{tabular}[c]{@{}c@{}} Rank\\ Upper\\ Bound\end{tabular} & \begin{tabular}[c]{@{}c@{}}SNR \\ Output \\ (dB)\end{tabular} & \begin{tabular}[c]{@{}c@{}}Estimated \\ Tensor \\ Rank\end{tabular} & \begin{tabular}[c]{@{}c@{}}SNR\\ Output \\ (dB)\end{tabular} & \begin{tabular}[c]{@{}c@{}}Estimated \\ Tensor \\ Rank\end{tabular} \\ \midrule
\multirow{2}{*}{Salinas-A}    & $\max\{J_n\}_{n=1}^N$                                                      & 43.7374                                            & 137                                                              & \textbf{44.0519}                                  & 143                                                              \\ \cmidrule(l){2-6} 
                              & $2\max\{J_n\}_{n=1}^N$                                                   & 46.7221                                            & 257                                                              & \textbf{46.7846}                                  & 260                                                              \\ \midrule
\multirow{2}{*}{Indian  Pines} & \multicolumn{1}{l|}{$\max\{J_n\}_{n=1}^N$ }                              & \multicolumn{1}{l|}{30.4207}                       & 169                                                              & \multicolumn{1}{l|}{\textbf{30.5541}}             & 178                                                              \\ \cmidrule(l){2-6} 
                              & \multicolumn{1}{l|}{$2\max\{J_n\}_{n=1}^N$}                              & \multicolumn{1}{l|}{31.9047}                       & 317                                                              & \multicolumn{1}{l|}{\textbf{32.0612}}                      & 335                                                              \\ \bottomrule
\end{tabular}}
\end{table}

On the other hand, since hyperspectral image (HSI) data are naturally three dimensional (two spatial dimensions and one spectral dimension), tensor CPD is a good tool to analyze such data.  However, due to the radiometric noise, photon effects and calibration errors, it is crucial to mitigate these corruptions before putting the HSI data into use. Since each HSI is rich in details, previous works using searching-based methods \cite{denoising1, denoising2} revealed that the tensor rank in HSI data is usually larger than half of the maximal tensor dimension. This corresponds to the {\it  \color{black} high tensor rank} scenario considered in this paper.  In these two real-world datasets, different bands of HSIs were corrupted by different levels of noises. Some of the HSIs are quite clean while some of them are quite noisy. For such types of real-world data, since no ground-truth is available, a no-reference quality assessment score is usually adopted \cite{denoising1, denoising2}. Following \cite{denoising1}, the SNR output, which is defined as $10 \log_{10} || \hat{\mathcal X} ||_F^2 / || \mathcal X - \hat{\mathcal X}||_F^2$ is utilized as the denoising performance measure, where $\hat{\mathcal X}$ is the restored tensor data and ${\mathcal X}$ is the original HSI data. In Table VII, the SNR outputs of the two methods using different rank upper bound values are presented, from which it can be seen that the proposed PCPD-GH method gives higher SNR outputs than PCPD-GG.  Samples of denoised HSIs are depicted in Appendix I.

%
%Samples of denoised HSIs are shown in Figure J1 (in Appendix J). On the left side of Figure J1, the relatively clean Salinas-A HSI in band 190 is presented to serve as a reference, from which it can be observed that the landscape exhibits ``stripe" pattern. For the noisy HSI in band 1, the denoising results from the two methods using the rank upper bound $\max\{J_n\}_{n=1}^N$ are presented. It is clear that the proposed method recovers better ``stripe" pattern than the PCPD-GG method. Similarly, the results from Indian Pines dataset are presented in the right side of Figure J1. For noisy HSI in band 1,  with the relatively clean image in band 10 serving as the reference, it can be observed that the proposed PCPD-GH method recovers more details than the PCPD-GG method, when both using rank upper bound  $2\max\{J_n\}_{n=1}^N$.}

\emph{Remark}: In HSI denoising, the state-of-the-art performance\footnote{https://paperswithcode.com/task/hyperspectral-image-classification.} is usually achieved via the integration of a tensor method and deep learning. In this work, we never claim that the proposed method gives the best performance in this specific task, but provides a more advanced solution for the CPD, which can be utilized as a building component for future HSI machine learning model design.

\section{Conclusions and Future Directions}
In this paper, we investigated the automatic tensor rank learning problem for  canonical polyadic decomposition (CPD) models under the framework of {parametric Bayesian modeling} and variational inference. By noticing that the performance of tensor rank learning is related to the flexibility of the prior distribution, we introduced the generalized hyperbolic (GH) prior to the probabilistic modeling of the CPD problem, based on which an inference algorithm is further developed.  Extensive numerical results on synthetic data showed that the proposed method exhibits excellent performance in learning both  low and  high tensor ranks, even when the noise power is large. This advantage is further evidenced by real-world data analysis.

This paper exemplified  how tensor rank learning performance can be enhanced via employing more advanced prior distributions {in the Gaussian scale mixture family}. Besides GH prior, there are many other advanced priors (including generalized-t distribution \cite{AD1}, normal-exponential gamma distribution \cite{AD2}) worth investigating. Besides exploiting the variants of the GH prior, informative structures such as non-negativeness \cite{CL5} and orthogonality \cite{CL2} can also be incorporated into the newly proposed probabilistic CPD model. Finally,  recent deep models utilize tensor decomposition formats to represent its multi-dimensional parameters \cite{B1,B2,B3}, and thus there is a surging need to automatically learn the ranks of these decompositions for  overfitting avoidance. Early attempts based on Gaussian-gamma prior were reported in \cite{B3}. Given that the GH model performs better than the Gaussian-gamma prior model, it is viable to extend these related works \cite{B1,B2,B3} by using the  prior model  proposed in this paper.  In the nonparametric front, replacing the gamma distribution of MGP \cite{new1} by a more advanced scale distribution in Gaussian scale mixture family might further improve its flexibility in sparsity modeling and learning. More discussions on future theoretical research are provided in Appendix O.

% {{} Future research directions are discussed in Appendix J.} 

\section{Appendix}
See  supplementary document. 
%\section{The Derivations of The Optimal Variational Pdfs in Table I}
%
%\section{The Derivations of The Update of Hyper-parameter $a_l^0$}

 \begin{IEEEbiography}
 [{\includegraphics[width=1in,height=1.25in,clip,keepaspectratio]{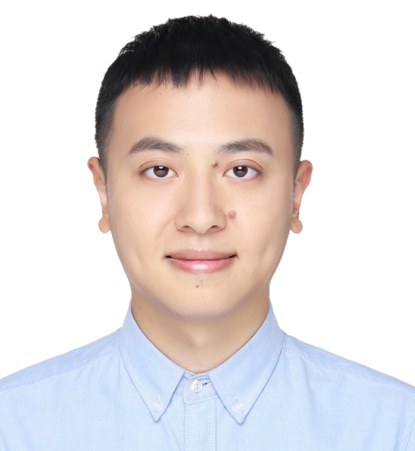}}]  {Lei Cheng} is currently Assistant Professor (ZJU Young Professor) with the College of Information Science and Electronic Engineering, Zhejiang University, Hangzhou, China. He received the B.Eng. degree from Zhejiang University in 2013, and the Ph.D. degree from The University of Hong Kong in 2018. He was a Research Scientist in Shenzhen Research Institute of Big Data, The Chinese University of Hong Kong, Shenzhen, from 2018 to 2021. His research interests are in Bayesian machine learning for tensor data analytics, and interpretable machine learning for information systems.
 \end{IEEEbiography}

 \begin{IEEEbiography}
 [{\includegraphics[width=1in,height=1.25in,clip,keepaspectratio]{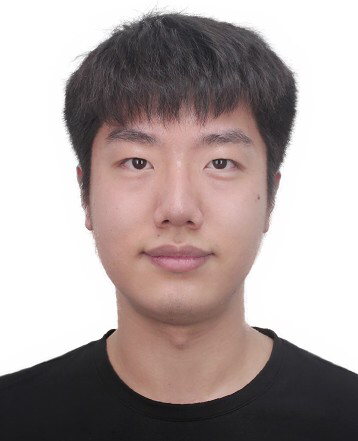}}]  {Zhongtao Chen} received the B.Eng. degree from The Chinese University of Hong Kong, Shenzhen, China, in 2021. He is currently working toward the Ph.D. degree at The University of Hong Kong, Pokfulam, Hong Kong. His research interests include signal processing and machine learning using Bayesian methods.
 \end{IEEEbiography}

  \begin{IEEEbiography}
 [{\includegraphics[width=1in,height=1.25in,clip,keepaspectratio]{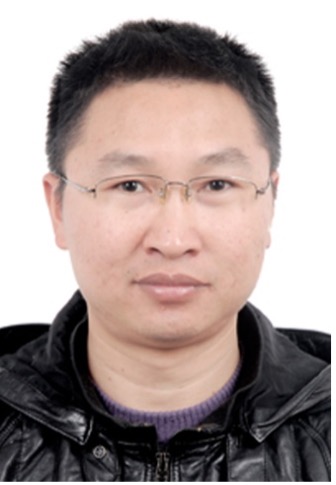}}]  {Qingjiang Shi} received his Ph.D. degree in electronic engineering from Shanghai Jiao Tong University, Shanghai, China, in 2011. From September 2009 to September 2010, he visited Prof. Z.-Q. (Tom) Luo's research group at the University of Minnesota, Twin Cities. In 2011, he worked as a Research Scientist at Bell Labs China. From 2012, He was with the School of Information and Science Technology at Zhejiang Sci-Tech University. From Feb. 2016 to Mar. 2017, he worked as a research fellow at Iowa State University, USA. From Mar. 2018, he is currently a full professor with the School of Software Engineering at Tongji University. He is also with the Shenzhen Research Institute of Big Data. His interests lie in algorithm design and analysis with applications in machine learning, signal processing and wireless networks. So far he has published more than 70 IEEE journals and filed about 30 national patents.

Dr. Shi was an Associate Editor for the IEEE TRANSACTIONS ON SIGNAL PROCESSING. He was the recipient of the Huawei Outstanding Technical Achievement Award in 2021, the Huawei Technical Cooperation Achievement Transformation Award (2nd Prize) in 2022, the Golden Medal at the 46th International Exhibition of Inventions of Geneva in 2018, the First Prize of Science and Technology Award from China Institute of Communications in 2017, the National Excellent Doctoral Dissertation Nomination Award in 2013, the Shanghai Excellent Doctorial Dissertation Award in 2012, and the Best Paper Award from the IEEE PIMRC'09 conference.

 \end{IEEEbiography}

%  \begin{IEEEbiography}
% [{\includegraphics[width=1in,height=1.25in,clip,keepaspectratio]{qj}}]  {Qingjiang Shi} received his Ph.D. degree in electronic engineering from Shanghai Jiao Tong University, Shanghai, China, in 2011. From September 2009 to September 2010, he visited Prof. Z.-Q. (Tom) Luo's research group at the University of Minnesota, Twin Cities. In 2011, he worked as a Research Scientist at Bell Labs China. From 2012, He was with the School of Information and Science Technology at Zhejiang Sci-Tech University. From Feb. 2016 to Mar. 2017, he worked as a research fellow at Iowa State University, USA. From Mar. 2018, he is currently a full professor with the School of Software Engineering at Tongji University. He is also with the Shenzhen Research Institute of Big Data. His interests lie in algorithm design and analysis with applications in machine learning, signal processing and wireless networks. So far he has published more than 50 IEEE journals and filed about 30 national patents.
%
%Dr. Shi was an Associate Editor for the IEEE TRANSACTIONS ON SIGNAL PROCESSING. He was awarded Golden Medal at the 46th International Exhibition of Inventions of Geneva in 2018, and also was the recipient of the First Prize of Science and Technology Award from China Institute of Communications in 2017, the National Excellent Doctoral Dissertation Nomination Award in 2013, the Shanghai Excellent Doctorial Dissertation Award in 2012, and the Best Paper Award from the IEEE PIMRC'09 conference.
%
% \end{IEEEbiography}
% 

 \begin{IEEEbiography}
 [{\includegraphics[width=1in,height=1.25in,clip,keepaspectratio]{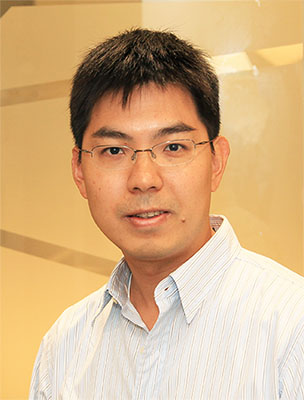}}]  {Yik-Chung Wu} (Senior Member, IEEE)
received the B.Eng. (EEE) and M.Phil. degrees from The University of Hong Kong (HKU) in 1998 and 2001, respectively, and the Ph.D. degree from Texas A\&M University, College Station, in 2005. From 2005 to 2006, he was with Thomson Corporate Research, Princeton, NJ, USA, as a Member of Technical Staff. Since 2006, he has been with HKU, where he is currently as an Associate Professor. He was a Visiting Scholar at Princeton University in Summers of 2015 and 2017. His research interests include signal processing, machine learning, and communication systems. He served as an Editor for IEEE COMMUNICATIONS LETTERS and IEEE TRANSACTIONS ON COMMUNICATIONS. He is currently an Editor for IEEE TRANSACTIONS ON SIGNAL PROCESSING and Journal of Communications and Networks.
 \end{IEEEbiography}

 \begin{IEEEbiography}
 [{\includegraphics[width=1in,height=1.25in,clip,keepaspectratio]{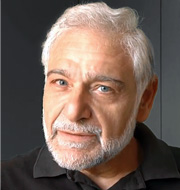}}]  {Sergios Theodoridis} (Life Fellow, IEEE)
 is currently Professor Emeritus with the Department of Informatics and Telecommunications, National and Kapodistrian University of Athens and he is Distinguished Professor with Aalborg University, Denmark. His research areas lie in the cross section of Signal Processing and Machine Learning. He is the author of the book “Machine Learning: A Bayesian and Optimization Perspective” Academic Press, 2nd Ed, 2020, the co-author of the best-selling book “Pattern Recognition”, Academic Press, 4th ed. 2009, and co-author of the book “Introduction to Pattern Recognition: A MATLAB Approach”, Academic Press, 2010.  He is the co-author of seven papers that have received {\it Best Paper Awards} including the 2014 IEEE Signal Processing Magazine Best Paper award and the 2009 IEEE Computational Intelligence Society Transactions on Neural Networks Outstanding Paper Award. He is the recipient of the 2021 IEEE SP Society {\it Norbert Wiener Award}, the 2017 EURASIP {\it Athanasios Papoulis Award}, the 2014 IEEE Signal Processing Society {\it Carl Friedrich Gauss Education Award} and the 2014 EURASIP {\it Meritorious Service Award}.   He has served as Vice President IEEE Signal Processing Society, EiC IEEE Transactions on SP, and as President of the European Association for Signal Processing (EURASIP).  He currently serves as the chair for the IEEE SPS awards committee.

He is Fellow of IET, a Corresponding Fellow of the Royal Society of Edinburgh (RSE), a Fellow of EURASIP and a Life Fellow of IEEE.
 \end{IEEEbiography}

\appendices

\clearpage
\newpage
\noindent {\bf Supplementary Document for ``Towards Flexible Sparsity-Aware Modeling: Automatic Tensor Rank Learning Using  Generalized Hyperbolic Prior''.}

\section{Table of Notation Symbols}
The notation symbols are summarized in Table A1.
\begin{table}[!h]
  \centering
  \caption*{ Table A1: Notation symbols.}
  \begin{tabular}{@{}|l|l|@{}}
  \toprule
  Notation         & Description    \\ \midrule
  $a, \boldsymbol{a}, \boldsymbol{A}, \mathcal{A}$   & Scalar, vector, matrix, tensor. \\ \midrule
  $\mathbb E [~\cdot~] $ & Expectation of its argument. \\ \midrule
  $\boldsymbol{A}^{T}$ & Transpose of matrix $\boldsymbol{A}$. \\ \midrule
  $\textrm{Tr}\left( {\boldsymbol{A}} \right)$ & Trace of matrix $\boldsymbol{A}$. \\ \midrule
  $\parallel \cdot \parallel_F$ & Frobenius norm of the argument. \\ \midrule
  $\mathcal {N} (\boldsymbol x | \boldsymbol u, \boldsymbol R)$ & Gaussian distribution with mean $\boldsymbol u$ \\ & and covariance matrix $\boldsymbol R$ for vector $\boldsymbol x$.  \\ \midrule
  $\mathrm{diag} \{a_1, a_2,...,a_N\}$ & $N \times N $ diagonal matrix with \\ & diagonal elements $a_1$ through $a_N$. \\ \midrule
  $\boldsymbol I_{M}$ & $M \times M$ identity matrix. \\ \midrule
  $\boldsymbol{A}_{:,j}$ & $j^{th}$ column of matrix $\boldsymbol A$. \\ \midrule
  $\boldsymbol{A}_{i,j}$ & $(i,j)^{th} $ element of matrix $\boldsymbol A$. \\ \midrule
  $\boldsymbol{A}_{i,:}$ & $i^{th}$ row of matrix $\boldsymbol A$. \\ \midrule
  $\circ$ & Vector outer product. \\ \midrule
  $\llbracket  \cdots \rrbracket$ & Kruskal operator. \\ \midrule
  $\mathrm{gamma}(x \mid a,b)$ & Gamma distribution with shape $a$ \\ & and rate $b$ for scalar $x$. \\ \midrule
  $\Gamma(\cdot)$ & Gamma function.\\ \midrule
  $\mathrm{vec}(\cdot)$ & Vectorization of its argument. \\ \midrule
  $K_{\cdot} (\cdot)$ & Modified Bessel function of the second kind. \\ \midrule 
  $\mathrm{GH}(\boldsymbol{x} \mid a,b,\lambda)$ &  Generalized hyperbolic distribution \\ & parametrized by $\{a,b,\lambda\}$ for vector $\boldsymbol x$. \\ \midrule
  $\mathrm{GIG}(x \mid a,b,\lambda)$ & Generalized inverse Gaussian distribution  \\ & parametrized by $\{a,b,\lambda\}$ for scalar $x$. \\ \midrule
  $\mathrm  {KL} (\cdot || \cdot)$ & KL divergence between two arguments. \\ \midrule 
  $\mathcal Y{(k)}$ & Matrix obtained by unfolding the tensor $\mathcal Y$ \\ &along its $k^{th}$ dimension. \\ \midrule
  $\odot$ & Khatri-Rao product. \\ \midrule
  $\circledast$ & Hadamard product. \\ \bottomrule
  $\mathop \circledast  \limits_{n=1,n\ne k}^N   {\boldsymbol  A}^{(n)}$ &  ${\boldsymbol  A}^{(N)}  \circledast \cdots \circledast {\boldsymbol  A}^{(k+1)} \circledast {\boldsymbol  A}^{(k-1)} \circledast \cdots \circledast {\boldsymbol  A}^{(1)} $.  \\ \bottomrule
$\mathop \odot  \limits_{n=1,n\ne k}^N   {\boldsymbol  A}^{(n)}$ & $ {\boldsymbol  A}^{(N)}  \odot \cdots \odot {\boldsymbol  A}^{(k+1)} \odot  {\boldsymbol  A}^{(k-1)} \odot \cdots \odot  {\boldsymbol  A}^{(1)} $. \\ \bottomrule
  \end{tabular}
\end{table}

\section{Special cases of GH prior}
\subsubsection{Student't Distribution} When $a_l^0 \rightarrow 0$ and $\lambda_l^0 < 0$, it can be shown that the GH prior \eqref{eq7}  reduces to \cite{AD4, AD5}
\begin{align}
& p ( \{ \boldsymbol  U^{(n)} \}_{n=1}^N ) = \prod_{l=1}^L \mathrm{GH}(\{ \boldsymbol  U^{(n)}_{:,l} \}_{n=1}^N | a_l^0 \rightarrow 0, b_l^0, \lambda_l^0 < 0) \nonumber \\
&=  \prod_{l=1}^L  (\frac{1}{\pi})^{\frac{ \sum_{n=1}^N J_n}{2}} \frac{\Gamma(\lambda_l^0 +  \sum_{n=1}^N \frac{J_n}{2} )}{ {b_l^0}^{\lambda_l^0} \Gamma(-\lambda_l^0)} \nonumber \\
&~~~~\! \times \! \left(b_l^0 + \bigparallel \mathrm{vec}\left(  \{\boldsymbol  U^{(n)}_{:,l}\}_{n=1}^N \right) {\bigparallel}_2^2 \right)^{\lambda_l^0 -  \sum_{n=1}^N \frac{J_n}{2} }\label{eq8}.
\end{align}
By comparing the functional form of \eqref{eq8} to that of  \eqref{eq4}, it is clear that pdf \eqref{eq8} is a student's t distribution with hyper-parameters $\{ b_l^0, \lambda_l^0\}$.

\subsubsection{Laplacian Distribution} When $b_l^0 \rightarrow 0$ and $\lambda_l^0 > 0$, it can be shown that the GH prior \eqref{eq7}  reduces to \cite{AD4, AD5}
\begin{align}
& p ( \{ \boldsymbol  U^{(n)} \}_{n=1}^N ) = \prod_{l=1}^L \mathrm{GH}(\{ \boldsymbol  U^{(n)}_{:,l} \}_{n=1}^N | a_l^0, b_l^0  \rightarrow 0, \lambda_l^0 > 0) \nonumber \\
& =  \prod_{l=1}^L  \frac{ (a_l^0)^{ \frac{ \sum_{n=1}^N J_n}{4} + \frac{\lambda_l^0}{2}  } } {  \left( \pi^{\frac{ \sum_{n=1}^N J_n}{2}}  \right) \left(2^{  \frac{ \sum_{n=1}^N J_n}{2} + \lambda_l^0 -1}  \right)}  \nonumber \\
& ~~~~\times \frac{  \bigparallel \mathrm{vec}\left(  \{\boldsymbol  U^{(n)}_{:,l}\}_{n=1}^N \right) {\bigparallel}_2^{\lambda_l^0 - \frac{\sum_{n=1}^N J_n}{2}}  } { 2^{\lambda_l^0}\Gamma \left( \lambda_l^0\right)} \nonumber \\
& ~~~~\times K_{\lambda_l^0 -  \frac{\sum_{n=1}^N J_n}{2}} \left( \sqrt{a_l^0} \bigparallel \mathrm{vec}\left(  \{\boldsymbol  U^{(n)}_{:,l}\}_{n=1}^N \right) {\bigparallel}_2  \right) \label{eq9}.
\end{align}
Pdf \eqref{eq9} characterizes a generalized Laplacian distribution. By setting $\lambda_l^0 = \frac{\sum_{n=1}^N J_n}{2}  + 1$,  pdf \eqref{eq9} can be reduced to a standard Laplacian pdf: 
\begin{align}
& p ( \{ \boldsymbol  U^{(n)} \}_{n=1}^N ) \nonumber \\
& = \prod_{l=1}^L \mathrm{GH}(\{ \boldsymbol  U^{(n)}_{:,l} \}_{n=1}^N | a_l^0, b_l^0  \rightarrow 0, \lambda_l^0 =  \frac{\sum_{n=1}^N J_n}{2}  + 1) \nonumber \\
& \propto  \prod_{l=1}^L  (a_l^0)^{ \frac{ \sum_{n=1}^N J_n}{2}} \exp\left( - \sqrt{a_l^0} \bigparallel \mathrm{vec}\left(  \{\boldsymbol  U^{(n)}_{:,l}\}_{n=1}^N \right) {\bigparallel}_2 \right). \label{eq10}
\end{align}

\section{Derivations for optimal variational density functions}

\subsection{The log of the joint distribution}
The joint distribution of the observation $\mathcal Y$ and model parameters $\boldsymbol \Theta = \{  \{\boldsymbol  U^{(n)}\}_{n=1}^N, \{z_l\}_{l=1}^L, \beta\}$ is
\begin{align}
  &p(\mathcal Y, \boldsymbol \Theta ) = p \left(  \mathcal Y  \mid \{\boldsymbol  U^{(n)}\}_{n=1}^N, \beta  \right) p\left( \{\boldsymbol  U^{(n)}\}_{n=1}^N | \{ z_l\}_{l=1}^L \right)  \nonumber \\
  & ~~~~~~~~~~~ \times p\left( \{ z_l\}_{l=1}^L  \right)p(\beta).
\end{align}
By using \eqref{eq15}, the log of joint distribution can be computed to be:
\begin{align}
    &\ln p(\mathcal Y, \boldsymbol \Theta ) = \nonumber \\ 
    & \frac{\prod_{n=1}^N J_n}{2}  \ln \beta  - \frac{\beta}{2} \parallel \mathcal Y - \llbracket  \boldsymbol U^{(1)},  \boldsymbol U^{(2)},..., \boldsymbol U^{(N)}  \rrbracket   \parallel_F^2   \nonumber \\
    & + \sum_{n=1}^N \left[ \frac{J_n}{2} \sum_{l=1}^L \ln z_l^{-1}  - \frac{1}{2} \mathrm{Tr}\left( \boldsymbol  U^{(n)} \boldsymbol Z^{-1} \boldsymbol  U^{(n)T} \right)  \right] \nonumber \\
    & + \sum_{l=1}^L \Big[  \frac{\lambda_l^0}{2} \ln \frac{a_l^0}{b_l^0} - \ln \left[2K_{\lambda_l^0} \left (\sqrt{a_l^0b_l^0} \right)\right]  + (\lambda_l^0-1) \ln z_l \nonumber \\
    & - \frac{1}{2} \left( a_l^0 z_l + b_l^0 z_l^{-1}\right)\Big]  + (\epsilon -1) \ln \beta - \epsilon \beta + \mathrm{const}, \label{logjoint}
\end{align}
where $\boldsymbol Z = \mathrm{diag}\{z_1, z_2, \cdots, z_L\}$.

Then, the optimal variational density funcitons can be derived by using
\begin{align}
    Q^* \left(  \boldsymbol \Theta_k\right) = \frac{\exp\left ( \mathbb E_{ \prod_{j \neq k}Q \left(   \boldsymbol \Theta_j\right) } \left [ \ln  {p(\mathcal Y, \boldsymbol \Theta )} \right]  \right)}{\int \exp\left ( \mathbb E_{ \prod_{j \neq k}Q \left(   \boldsymbol \Theta_j\right) } \left [ \ln  {p(\mathcal Y, \boldsymbol \Theta )} \right]  \right)  d\boldsymbol \Theta_k }.
\end{align}
We present the detailed derivation for each optimal variational density function in the following subsections.

\subsection{Optimal variational density function of $\{\boldsymbol  U^{(k)}\}_{k=1}^N$}

\begin{align}
&\ln Q^* \left( \boldsymbol  U^{(k)} \right) \nonumber \\
&= \mathbb E_{Q \left( \boldsymbol \Theta \setminus \boldsymbol  U^{(k)} \right)} \left[ \ln p(\mathcal Y, \boldsymbol \Theta ) \right] + \mathrm{const} \nonumber \\
&= \mathbb E \Big[ - \frac{\beta}{2} \parallel \mathcal Y - \llbracket  \boldsymbol U^{(1)},  \boldsymbol U^{(2)},..., \boldsymbol U^{(N)}  \rrbracket   \parallel_F^2 \nonumber \\
& - \frac{1}{2} \mathrm{Tr}\left( \boldsymbol  U^{(k)} \boldsymbol Z^{-1} \boldsymbol  U^{(k)T} \right) \Big]  + \mathrm{const} \nonumber \\
&= \mathbb E \left[ - \frac{\beta}{2} \mathrm{Tr} \left( \left( \mathcal Y (k) - \boldsymbol  U^{(k)} \left(\mathop \odot  \limits_{n=1,n\ne k}^N \boldsymbol  U^{(n)}\right)^{T} \right) \right. \right. \nonumber\\ 
& ~~~ \times \left. \left. \left( \mathcal Y (k) - \boldsymbol  U^{(k)} \left(\mathop \odot  \limits_{n=1,n\ne k}^N \boldsymbol  U^{(n)}\right)^{T} \right)^{T} \right) \right. \nonumber \\
& ~~~ \left. - \frac{1}{2} \mathrm{Tr}\left( \boldsymbol  U^{(k)} \boldsymbol Z^{-1} \boldsymbol  U^{(k)T} \right) \right] + \mathrm{const} \nonumber \\
&= \mathbb E \left[ - \frac{1}{2} \mathrm{Tr} \left( \boldsymbol  U^{(k)} \left( \beta \left(\mathop \odot  \limits_{n=1,n\ne k}^N \boldsymbol  U^{(n)}\right)^T \left( \mathop  \odot  \limits_{n=1,n\ne k}^N \boldsymbol  U^{(n)} \right) \right. \right. \right. \nonumber \\
& ~~ \left. \left. \left. + \boldsymbol Z^{-1} \right) \boldsymbol  U^{(k)T} \right) + \mathrm{Tr} \left( \beta \mathcal Y (k) \left( \mathop  \odot  \limits_{n=1,n\ne k}^N \boldsymbol  U^{(n)} \right) \boldsymbol  U^{(k)T} \right) \right] \nonumber \\
& ~~~ + \mathrm{const} \nonumber \\
&= - \frac{1}{2} \mathrm{Tr} \left( \boldsymbol  U^{(k)} \left( \mathbb E \left[ \beta \right] \mathbb E \left[ \left(\mathop \odot  \limits_{n=1,n\ne k}^N \boldsymbol  U^{(n)}\right)^T \right. \right. \right. \nonumber \\ 
& ~~ \times \left. \left. \left. \left( \mathop  \odot  \limits_{n=1,n\ne k}^N \boldsymbol  U^{(n)} \right) \right] + \mathbb E \left[ \boldsymbol Z^{-1} \right] \right) \boldsymbol  U^{(k)T} \right) \nonumber \\
& ~~~ + \mathrm{Tr} \left( \mathbb E \left[ \beta \right] \mathcal Y (k)   \left( \mathop  \odot  \limits_{n=1,n\ne k}^N \mathbb E \left[ \boldsymbol  U^{(n)} \right] \right) \boldsymbol  U^{(k)T} \right) + \mathrm{const}.
\end{align}

By comparing the functional form of $Q^* \left( \boldsymbol  U^{(k)} \right)$ to that of matrix normal distribution \cite{MND}, it can be concluded that $Q^* \left( \boldsymbol  U^{(k)} \right)$ follows the matrix normal distrbution $\mathcal{MN} \left(\boldsymbol  U^{(k)} |\boldsymbol M^{(k)}, \boldsymbol I_{J_n}, \boldsymbol \Sigma^{(k)} \right)$ with covariance $\boldsymbol \Sigma^{(k)}$ and mean $\boldsymbol M^{(k)}$ expressed in \eqref{eq20} and \eqref{eq21} respectively.

\subsection{Optimal variational density function of $\{z_l\}_{l=1}^L$}

\begin{align}
& \ln Q^* \left( \{z_l\}_{l=1}^L \right) \nonumber \\
& = \mathbb E_{Q \left( \boldsymbol \Theta \setminus \{z_l\}_{l=1}^L \right)} \left[ \ln p(\mathcal Y, \boldsymbol \Theta ) \right] + \mathrm{const} \nonumber \\
&= \mathbb E \left[ \sum_{n=1}^N \left[ \frac{J_n}{2} \sum_{l=1}^L \ln z_l^{-1}  - \frac{1}{2} \mathrm{Tr}\left( \boldsymbol  U^{(n)} \boldsymbol Z^{-1} \boldsymbol  U^{(n)T} \right)  \right] \right. \nonumber \\
& ~~ \left. + \sum_{l=1}^L \left[ (\lambda_l^0-1) \ln z_l - \frac{1}{2} \left( a_l^0 z_l + b_l^0 z_l^{-1}\right) \right] \right] + \mathrm{const} \nonumber\\
&= \mathbb E \left[ \sum_{n=1}^N \left[ \frac{J_n}{2} \sum_{l=1}^L \ln z_l^{-1}  - \frac{1}{2} \sum_{l=1}^L \left[\boldsymbol  U^{(n)}_{:,l} \right]^{T} \boldsymbol  U^{(n)}_{:,l} z_l^{-1} \right] \right. \nonumber \\
& ~~\left. + \sum_{l=1}^L \left[ (\lambda_l^0-1) \ln z_l - \frac{1}{2} \left( a_l^0 z_l + b_l^0 z_l^{-1}\right) \right] \right] + \mathrm{const} \nonumber \\
&= \sum_{l=1}^L \mathbb E \left[ \frac{1}{2} \sum_{n=1}^N J_n \ln z_l^{-1} - \frac{1}{2} \sum_{n=1}^N \left[\boldsymbol  U^{(n)}_{:,l} \right]^{T} \boldsymbol  U^{(n)}_{:,l} z_l^{-1} \right. \nonumber \\
& ~~ \left.+ (\lambda_l^0-1) \ln z_l - \frac{1}{2} \left( a_l^0 z_l + b_l^0 z_l^{-1}\right) \right] + \mathrm{const} \nonumber \\
&= \sum_{l=1}^L -\frac{1}{2} a_l^0 z_l - \frac{1}{2} \left( b_l^0 + \sum_{n=1}^N \mathbb E \left[  \left[\boldsymbol  U^{(n)}_{:,l} \right]^T \boldsymbol  U^{(n)}_{:,l}  \right] \right) z_l^{-1} \nonumber \\
& ~~ + \left( \lambda_l^0 - \frac{1}{2} \sum_{n=1}^N J_n - 1 \right) \ln z_l + \mathrm{const}. \label{qzl}
\end{align}

From \eqref{qzl}, it can be seen that $Q^* \left( \{z_l\}_{l=1}^L \right) = \prod_{l=1}^L Q^* (z_l) $, where
\begin{align}
& Q^* (z_l) \propto \exp \Biggl\{ -\frac{1}{2} a_l^0 z_l - \frac{1}{2} \left( b_l^0 + \sum_{n=1}^N \mathbb E \left[  \left[\boldsymbol  U^{(n)}_{:,l} \right]^T \boldsymbol  U^{(n)}_{:,l}  \right] \right) z_l^{-1} \nonumber \\
& ~~ + \left( \lambda_l^0 - \frac{1}{2} \sum_{n=1}^N J_n - 1 \right) \ln z_l \Biggr\}.
\end{align}
After comparing the functional form of $Q^* (z_l)$ to that of GIG distribution \cite{GIG}, it can be concluded that $Q^* (z_l) \sim \mathrm{GIG}(z_l \mid a_l, b_l, \lambda_l)$, with $a_l, b_l, \lambda_l$ expressed in \eqref{al}-\eqref{lambdal} respectively.

\subsection{Optimal variational density function of $\beta$}

\begin{align}
&\ln Q^* \left( \beta \right) \nonumber \\
&= \mathbb E_{Q \left( \boldsymbol \Theta \setminus \beta \right)} \left[\ln p(\mathcal Y, \boldsymbol \Theta ) \right] + \mathrm{const} \nonumber \\
&= \mathbb E \left[ \frac{\prod_{n=1}^N J_n}{2} \ln \beta -\frac{\beta}{2} \parallel \mathcal Y - \llbracket  \boldsymbol U^{(1)},  \boldsymbol U^{(2)},..., \boldsymbol U^{(N)}  \rrbracket   \parallel_F^2 \right. \nonumber \\
& ~~ \left. + \left( \epsilon - 1 \right) \ln \beta - \epsilon \beta \right] + \mathrm{const} \nonumber \\
&= - \left( \epsilon + \frac{1}{2} \mathbb E \left[ \parallel \mathcal Y - \llbracket  \boldsymbol U^{(1)},  \boldsymbol U^{(2)},..., \boldsymbol U^{(N)}  \rrbracket   \parallel_F^2 \right] \right) \beta \nonumber \\
& ~~ + \left( \epsilon + \frac{1}{2} \prod_{n=1}^N J_n - 1 \right) \ln \beta + \mathrm{const}. \label{qbeta}
\end{align}
After comparing the functional form of \eqref{qbeta} to that of gamma distribution, it can be concluded that $Q^* \left( \beta \right) = \mathrm{gamma} (\beta \mid e,f)$ where $e$ and $f$ are expressed in \eqref{e} and \eqref{f} respectively.

\subsection{Optimization of hyper-parameter $\{a^0_l\}_{l=1}^L$}

After introducing the conjugate hyper-prior $p(a_l^0) = \mathrm{gamma} (a_l^0 \mid \kappa_{a_1}, \kappa_{a_2})$ for $a_l^0$ and let $b_l^0 \rightarrow 0$, in the log of the joint distribution $\ln p(\mathcal Y, \boldsymbol \Theta )$ expressed in \eqref{logjoint}, the term that is relevant to $a_l^0$ is
\begin{align}
  \sum_{l=1}^L \left( \frac{\lambda_l^0}{2} \ln a_l^0 - \frac{1}{2} a_l^0 z_l \right) + \left( \kappa_{a_1} - 1 \right) \ln a^0_l - \kappa_{a_2} a^0_l.
\end{align}
Therefore, the optimal values of $\{a^0_l\}_{l=1}^L$ can be derived by solving 
\begin{align}
&\max_{\{a^0_l\}_{l=1}^L } \mathbb E_{Q \left( \boldsymbol \Theta \setminus \{a^0_l\}_{l=1}^L \right)} \left[\ln p(\mathcal Y, \boldsymbol \Theta ) \right]  \nonumber \\
&= \mathbb E \left[ \sum_{l=1}^L \left( \frac{\lambda_l^0}{2} \ln \frac{a_l^0}{b_l^0} - \frac{1}{2} a_l^0 z_l \right) + \left( \kappa_{a_1} - 1 \right) \ln a^0_l - \kappa_{a_2} a^0_l \right] \nonumber \\
&= \sum_{l=1}^L \left[ \left( \kappa_{a_1} + \frac{\lambda_l^0}{2} - 1 \right) \ln a^0_l - \left( \kappa_{a_2} + \frac{\mathbb E \left[ z_l \right] }{2} \right) a^0_l \right].
\end{align}
By taking the derivative with respect to $a^0_l$,  it is easy to show that  each $a^0_l$ is updated as \eqref{al0}.

\section{Initialization for Algorithm 1}
Due to the BCD nature of the mean-field VI, it is important to choose good initial values for the proposed learning algorithm in order to avoid poor local minima. In particular, being consistent to the released codes of \cite{PI1,PI2}, the upper bound of tensor rank could be set to be the maximum of the tensor dimensions, i.e., $L = \max\{ J_n\}_{n=1}^N$. On the other hand, the initial mean factor matrix $[\boldsymbol M^{(n)}]^0$ is set as the  singular value decomposition (SVD) approximation. In particular,  when $L \leq J_n$, the initial mean factor matrix $[\boldsymbol M^{(n)}]^0$ is set as the  singular value decomposition (SVD) approximation $\boldsymbol U_{:,1:L} \left(\boldsymbol S_{1:L,1:L}\right)^{\frac{1}{2}} $, where $[\boldsymbol U, \boldsymbol S, \boldsymbol V] = \mathrm{SVD}[\mathcal Y^{(n)}]$. When $L > J_n$, the initial mean factor matrix $[\boldsymbol M^{(n)}]^0 = [\boldsymbol U_{:,1:L} \left(\boldsymbol S_{1:L,1:L}\right)^{\frac{1}{2}}, \boldsymbol Q ]$, where  each element  in $\boldsymbol Q \in \mathbb R^{J_n \times (L-J_n)}$ is drawn from $\mathcal N(0,1)$. $e^0$ and $f^0$ are set to be a very small number, e.g., $10^{-6}$ to indicate the non-informativeness of the noise power. The initial mean  $m[z_l^{-1}]^0 = 1$. For the initial hyper-parameters $\{a_l^0, b_l^0, \lambda_l^0\}$ of the GH prior, as discussed in Section IV.C, $b_l^0$ can be set to zero and $a_l^0$ will be updated by the algorithm. For $\lambda_l^0$, it was found that the smaller value will lead to a higher peak at the zero point of the GH distribution \cite{AD4}. To adapt for different tensor sizes, it is set as $-\min\{ J_n\}_{n=1}^N$. Finally, since $\kappa_{a_1} > 1 -\lambda_l^0/2$ and $\kappa_{a_2} \geq 0$, they can be set as $\kappa_{a_1} =  2 -\lambda_l^0/2$ and   $\kappa_{a_2} = 10^{-6}$.

% \usepackage{multirow}
\begin{table*}[!t]
\centering
\caption*{Table EI: The average run time (sec) of different algorithms versus SNRs (dB) presented in Fig. 9.}
\begin{tabular}{|c|c|ccccccc|}
\hline
\multirow{2}{*}{\begin{tabular}[c]{@{}c@{}}True Tensor \\ Rank\end{tabular}} &
  \multirow{2}{*}{Algorithm} &
  \multicolumn{7}{c|}{SNR (dB)} \\ \cline{3-9} 
 &
   &
  \multicolumn{1}{c|}{-10} &
  \multicolumn{1}{c|}{-5} &
  \multicolumn{1}{c|}{0} &
  \multicolumn{1}{c|}{5} &
  \multicolumn{1}{c|}{10} &
  \multicolumn{1}{c|}{15} &
  20 \\ \hline
\multirow{4}{*}{R = 6} &
  PCPD-GG-max(DimY) &
  \multicolumn{1}{c|}{0.21} &
  \multicolumn{1}{c|}{0.33} &
  \multicolumn{1}{c|}{0.65} &
  \multicolumn{1}{c|}{0.07} &
  \multicolumn{1}{c|}{0.04} &
  \multicolumn{1}{c|}{0.04} &
  0.05 \\ \cline{2-9} 
 &
  PCPD-GG-2max(DimY) &
  \multicolumn{1}{c|}{0.24} &
  \multicolumn{1}{c|}{0.35} &
  \multicolumn{1}{c|}{0.72} &
  \multicolumn{1}{c|}{0.30} &
  \multicolumn{1}{c|}{0.32} &
  \multicolumn{1}{c|}{0.53} &
  0.58 \\ \cline{2-9} 
 &
  PCPD-GH-max(DimY) &
  \multicolumn{1}{c|}{0.27} &
  \multicolumn{1}{c|}{0.57} &
  \multicolumn{1}{c|}{1.40} &
  \multicolumn{1}{c|}{2.22} &
  \multicolumn{1}{c|}{0.69} &
  \multicolumn{1}{c|}{1.02} &
  0.09 \\ \cline{2-9} 
 &
  PCPD-GH-2max(DimY) &
  \multicolumn{1}{c|}{0.32} &
  \multicolumn{1}{c|}{0.63} &
  \multicolumn{1}{c|}{1.61} &
  \multicolumn{1}{c|}{3.12} &
  \multicolumn{1}{c|}{1.84} &
  \multicolumn{1}{c|}{1.83} &
  1.81 \\ \hline
\multirow{4}{*}{R = 24} &
  PCPD-GG-max(DimY) &
  \multicolumn{1}{c|}{0.17} &
  \multicolumn{1}{c|}{0.24} &
  \multicolumn{1}{c|}{0.54} &
  \multicolumn{1}{c|}{0.97} &
  \multicolumn{1}{c|}{1.69} &
  \multicolumn{1}{c|}{2.43} &
  2.69 \\ \cline{2-9} 
 &
  PCPD-GG-2max(DimY) &
  \multicolumn{1}{c|}{0.21} &
  \multicolumn{1}{c|}{0.29} &
  \multicolumn{1}{c|}{0.60} &
  \multicolumn{1}{c|}{1.17} &
  \multicolumn{1}{c|}{3.13} &
  \multicolumn{1}{c|}{3.59} &
  1.02 \\ \cline{2-9} 
 &
  PCPD-GH-max(DimY) &
  \multicolumn{1}{c|}{0.18} &
  \multicolumn{1}{c|}{0.37} &
  \multicolumn{1}{c|}{1.30} &
  \multicolumn{1}{c|}{1.94} &
  \multicolumn{1}{c|}{3.31} &
  \multicolumn{1}{c|}{3.47} &
  3.54 \\ \cline{2-9} 
 &
  PCPD-GH-2max(DimY) &
  \multicolumn{1}{c|}{0.31} &
  \multicolumn{1}{c|}{0.48} &
  \multicolumn{1}{c|}{1.18} &
  \multicolumn{1}{c|}{3.11} &
  \multicolumn{1}{c|}{6.27} &
  \multicolumn{1}{c|}{0.75} &
  1.11 \\ \hline
\end{tabular}
\end{table*}

\section{The run time of PCPD-GH and PCPD-GG}

in Table E1, we present the run time of PCPD-GH and PCPD-GG under two upper bound values (max(DimY) and 2max(DimY), following the simulation settings of Figure 9.  In general,  PCPD-GH-2max(DimY), which exhibits the best performance,  costs the most time in nearly all the cases, since it starts with a high upper bound value and needs to update more parameters. Under some simulation settings (e.g., R = 24, SNR = 20 dB), the algorithms with max(DimY) fail to estimate the tensor rank and have difficulty in convergence, thus resulting in more run time than the counterparts with 2max(DimY). To draw further insights, in Figure E1, we present the run time of the two algorithm, with 1x/2x  max(DimY), over different tensor ranks under SNR = 10 dB. From Figure E1, we can observe that  if the algorithm accurately learns the tensor rank, its run time increases with the ground-truth tensor rank.

\begin{figure} [!t]
\setcounter{subfigure}{0}
\centering
\subfigure[] {
\includegraphics[width= 3 in]{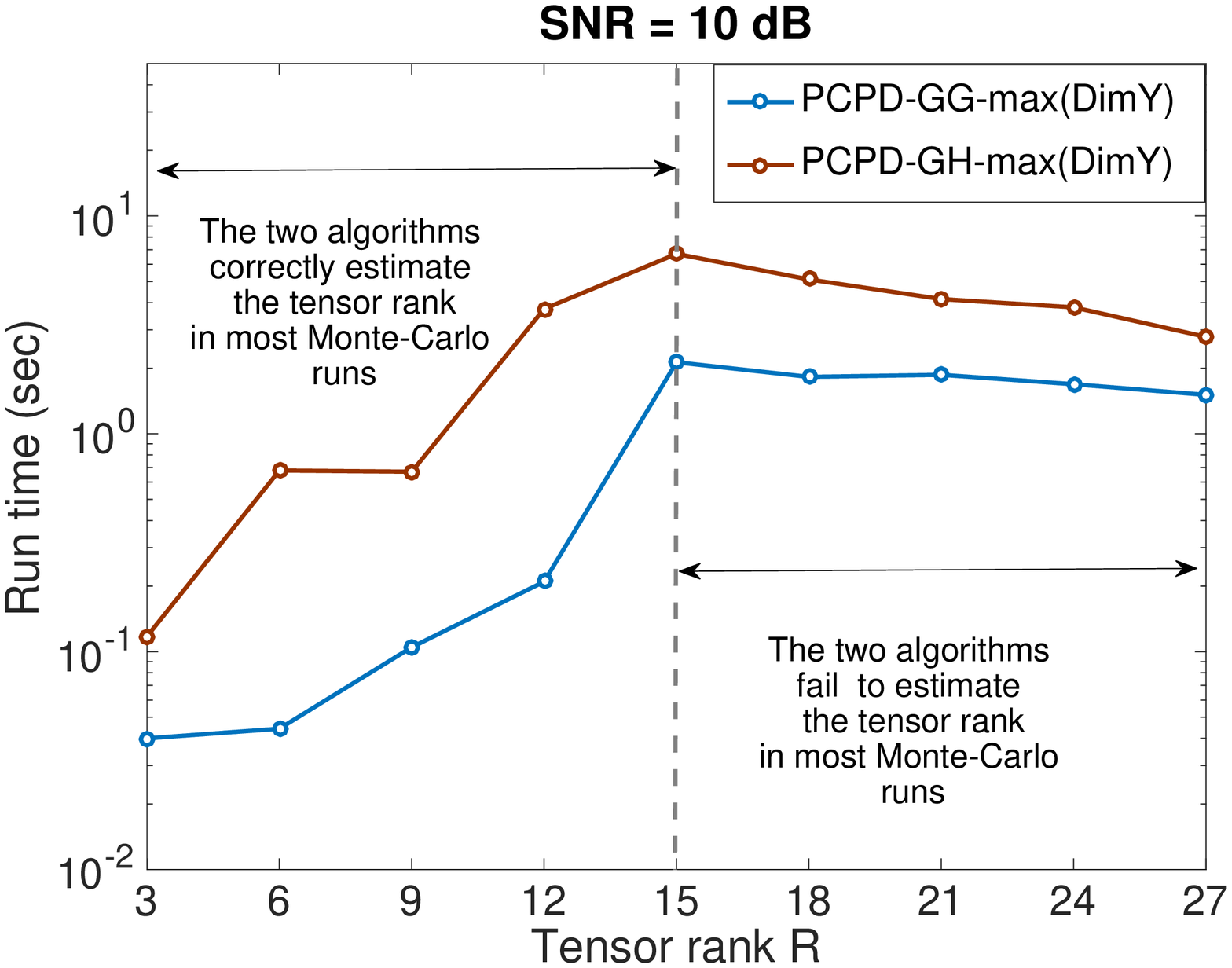}
}
\subfigure[] {
\includegraphics[width= 3 in]{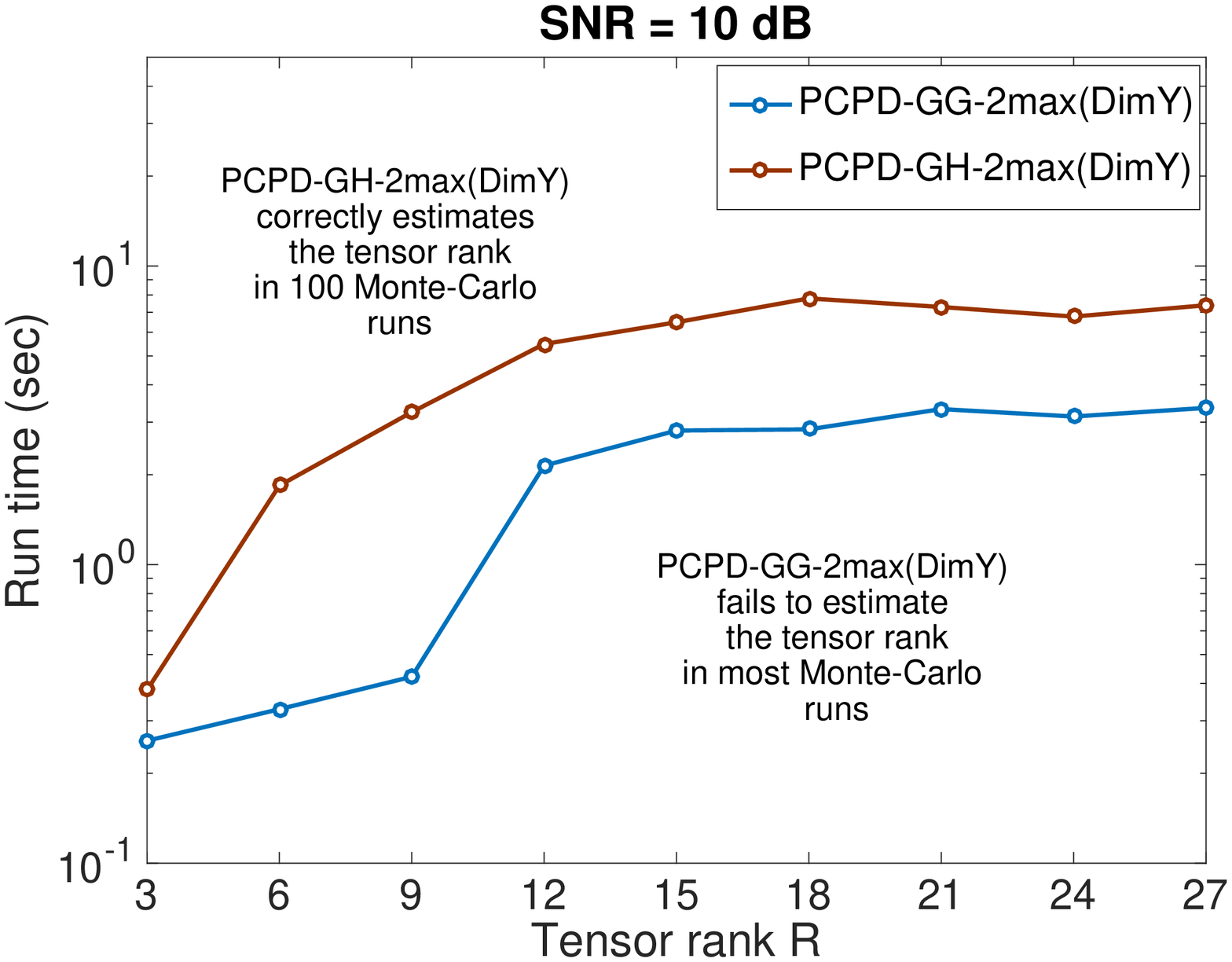}
}
\caption*{Figure E1: The run time (sec) versus different tensor ranks under SNR = 10 dB. (a) The rank upper bound is max(DimY). (b) The rank upper bound is 2max(DimY).}
\end{figure}

\vspace{0.3cm}

\section{RMSE for tensor recovery under different SNRs and tensor ranks}

In Table F1 (at the top of the next page), we present the root mean square errors (RMSEs) for tensor recovery under different SNRs and tensor ranks. The RMSE is defined as $\left(\frac{1}{\prod_{n=1}^3 I_n} || \mathcal X-  \llbracket  \boldsymbol M^{(1)},   \boldsymbol M^{(2)},  \boldsymbol M^{(3)} \rrbracket   ||_F^2\right)^{\frac{1}{2}}$.  We consider the low-rank tensor scenario (i.e., $R=6$) and the high-rank tensor scenario (i.e., $R=24$) for different SNRs. In general, a correct learnt tensor rank is essential in avoiding overfitting of noises or underfitting of signals. Therefore, in Table F1, it can be observed the PCPD-GH method gives the best RMSE in most cases due to its superior capabilities in tensor rank learning.

\begin{table*}[!h]
\centering
\caption*{Table F1: RMSE for tensor recovery under different SNRs and tensor ranks}
\begin{tabular}{@{}|c|c|c|c|c|c|c|c|c|@{}}
\toprule
SNR (dB)              & \multicolumn{2}{c|}{-10}          & \multicolumn{2}{c|}{-5}           & \multicolumn{2}{c|}{0}            & \multicolumn{2}{c|}{5}            \\ \midrule
True Tensor Rank      & R=6             & R=24            & R=6             & R=24            & R=6             & R=24            & R=6             & R=24            \\ \midrule
PCPD-GH-2max(DimY)    & \textbf{1.1895} & 4.7272          & \textbf{0.6462} & 3.2074          & \textbf{0.3631} & \textbf{1.3932} & \textbf{0.2042} & \textbf{0.7801} \\ \midrule
PCPD-GG-max(DimY) & 1.2083          &  4.5208          & 0.6765          &  3.1333           & 0.3744          &  1.9039          & 0.2043          & 1.6349           \\ \midrule
PCPD-GG-2max(DimY)    & 1.2320          & \textbf{4.5032} & 0.6939          & \textbf{2.6082} & 0.3883          & 1.3996          & 0.2147          & 0.7858          \\ \midrule
SNR (dB)              & \multicolumn{2}{c|}{10}           & \multicolumn{2}{c|}{15}           & \multicolumn{2}{c|}{20}           & \multicolumn{2}{c|}{-}            \\ \midrule
True Tensor Rank      & R=6             & R=24            & R=6             & R=24            & R=6             & R=24            & -               & -               \\ \midrule
PCPD-GH-2max(DimY)    & \textbf{0.1149} & \textbf{0.4381} & \textbf{0.0646} & \textbf{0.2463} & \textbf{0.0363} & \textbf{0.1385} & -               & -               \\ \midrule
PCPD-GG-max(DimY) & \textbf{0.1149} &  1.6513         & \textbf{0.0646} & 1.6372           & \textbf{0.0363} &  1.5945          & -               & -               \\ \midrule
PCPD-GG-2max(DimY)    & 0.1219          & 0.4438          & 0.0688          & 0.2498          & 0.0395          & 0.1402          & -               & -               \\ \bottomrule
\end{tabular}
\end{table*}

\section{Additional Validations on Synthetic Data}

\subsection{L-curve for Tensor Rank Estimation}

\begin{figure} [!h]
\setcounter{subfigure}{0}
\centering
\subfigure[] {
\includegraphics[width=3 in]{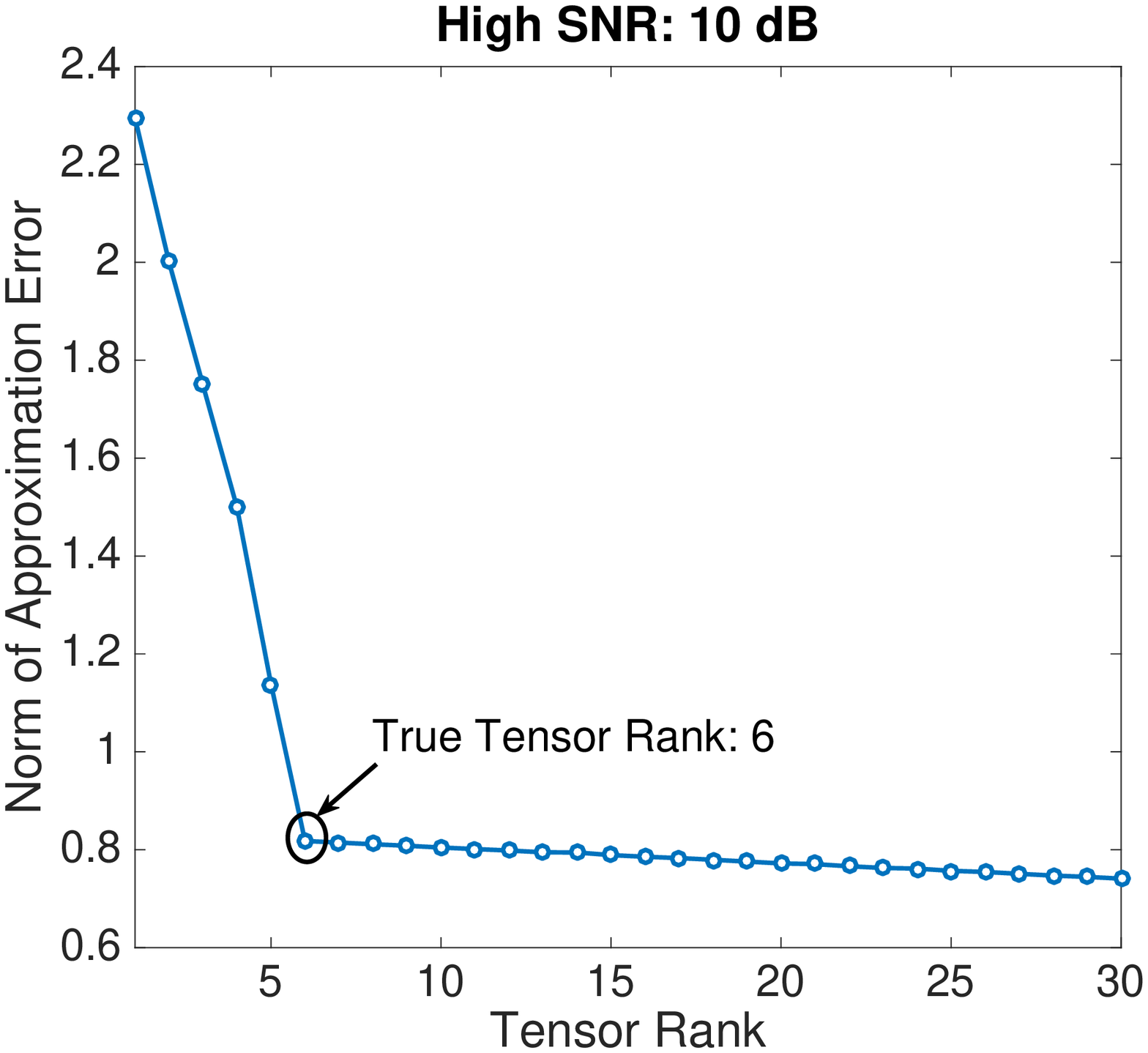}
}
\subfigure[] {
\includegraphics[width=3 in]{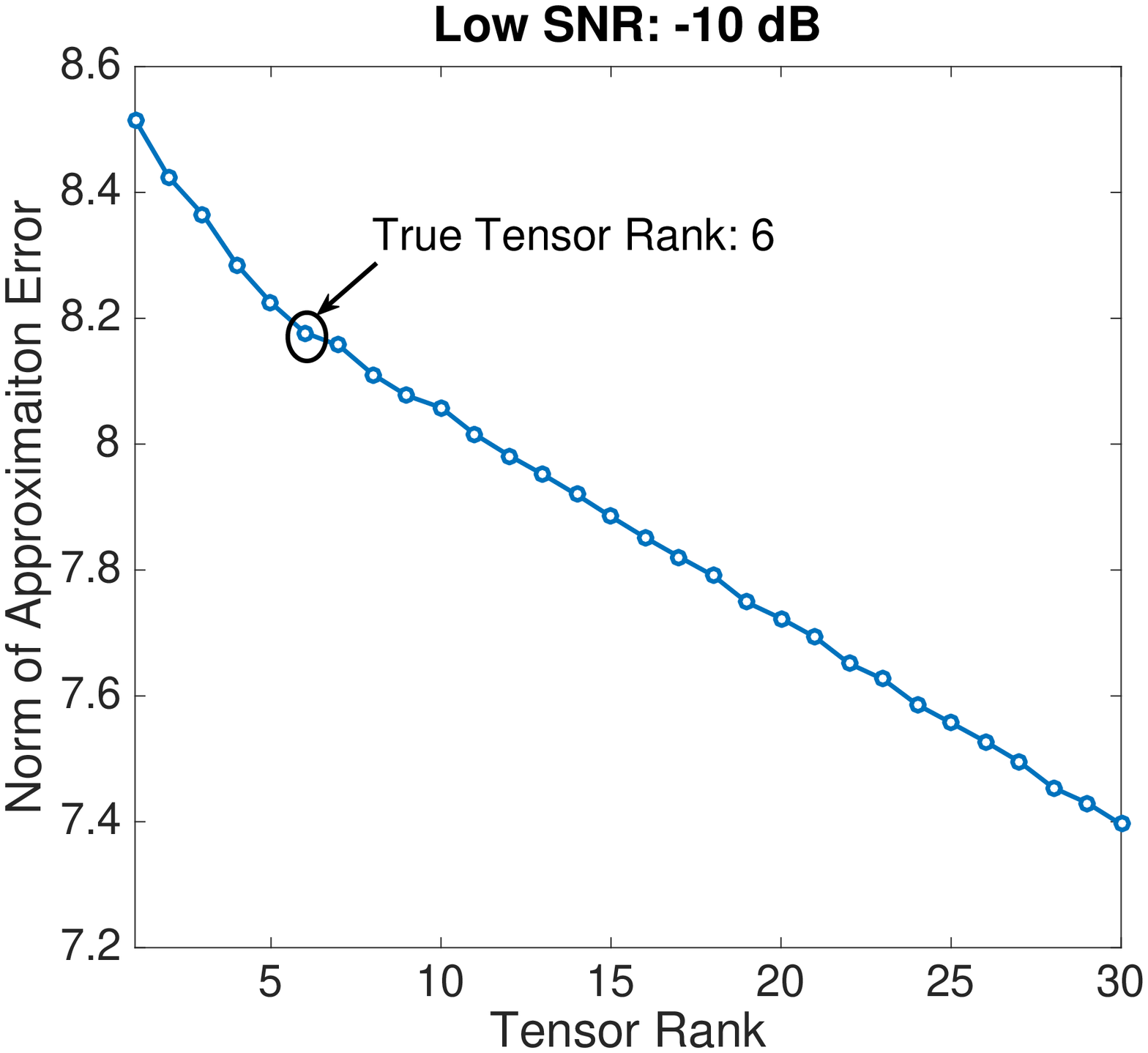}
}
\caption*{ Figure G1. L-curve for tensor rank estimation under (a) SNR = 10 dB and (b) SNR = -10 dB. Given fixed tensor rank,  CPD is computed by  function $\mathrm{cpd}$ in Tensorlab 3.0 toolbox \cite{tensorlab}.}
\label{fig}
\end{figure}

Following synthetic tensor data generation in the main body of this paper, we investigate the difficulty of using 
L-curve (the error between the observed tensor and the CPD approximation under different  tensor ranks) \cite{Sergios2} to find the tensor rank. It is not  a difficult problem in high SNR regime. As illustrated in Figure G1 (a), finding the corner of the L-curve can estimate tensor rank well when SNR = 10 dB.  However, for the very noisy cases, e.g., SNR = -10 dB, it is difficult to identify the the corner of the L-curve, as illustrated in Figure G1 (b), making tensor rank estimation a challenging task.

\subsection{Tensor Rank Estimation When Each Factor Matrix Has Correlated Elements}
The numerical experiments are conducted by considering tensor  $\mathcal X =  \llbracket  \boldsymbol A^{(1)},   \boldsymbol A^{(2)},  \boldsymbol A^{(3)} \rrbracket  \in \mathbb R^{30 \times 30 \times 30}$ with different tensor rank $R$. For each factor matrix $\boldsymbol A^{(k)}$, each of its row is generated from a zero mean Gaussian distribution with covariance matrix  $\boldsymbol P^{(k)} = \boldsymbol F^{(k)} \left[\boldsymbol  F^{(k)}\right]^T$, where each element of $\boldsymbol F^{(k)}$ is drawn from $\mathcal N(0,1)$. The illustration of covariance matrix realization is presented in Figure G2 (a), which is significantly different from the identity covariance matrix (illustrated in Figure G2 (b)).  Using these factor matrices with correlations, the tensor rank estimation performances under different upper bound values and SNRs are presented in Figure G3 and Figure G4, respectively. It can be observed  the conclusions drawn from  Figure G3 and Figure G4 are similar to those  in the main body.

\begin{figure} [!h]
\setcounter{subfigure}{0}
\centering
\subfigure[] {
\includegraphics[width=3 in]{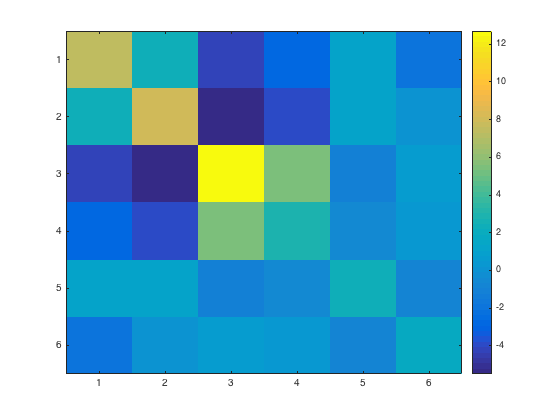}
}
\subfigure[] {
\includegraphics[width=3 in]{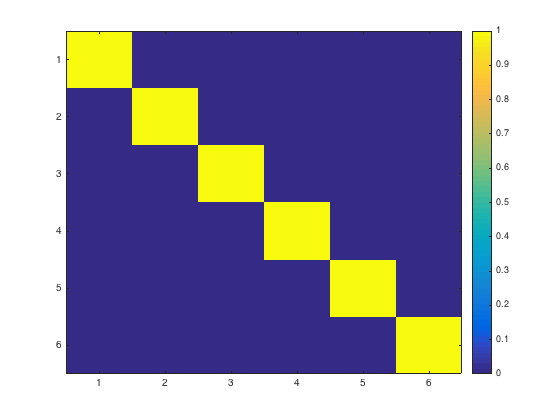}
}
\caption*{ Figure G2. Illustrations of covariance matrices: (a) $\boldsymbol P^{(k)} = \boldsymbol F^{(k)} \left[\boldsymbol  F^{(k)}\right]^T$, where each element of $\boldsymbol F^{(k)}$ is drawn from $\mathcal N(0,1)$; (b) $\boldsymbol P^{(k)} = \boldsymbol I$.}
\label{fig}
\end{figure}

\begin{figure} [!t]
\setcounter{subfigure}{0}
\centering
\subfigure[] {
\includegraphics[width=3 in]{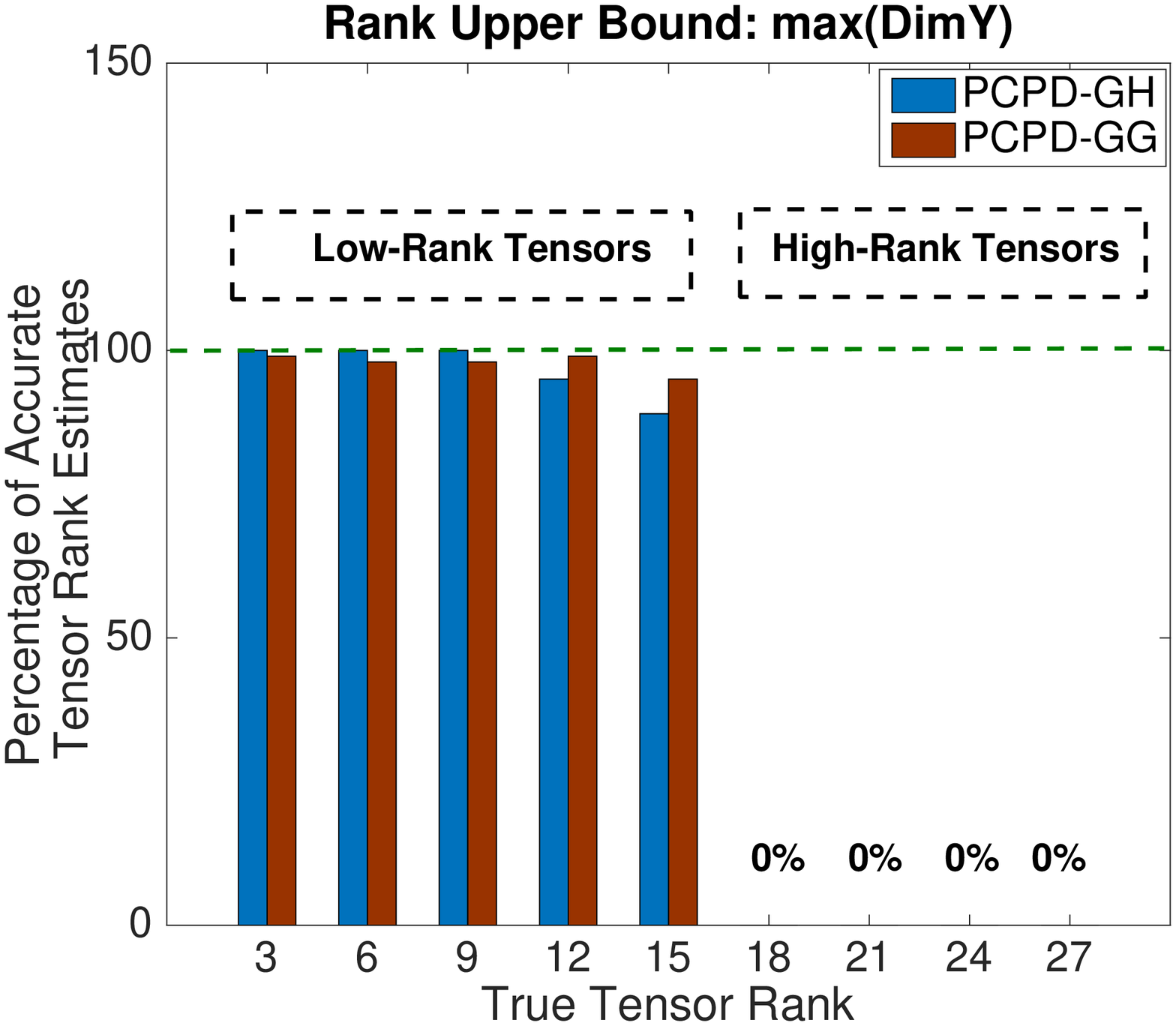}
}
\subfigure[] {
\includegraphics[width=3 in]{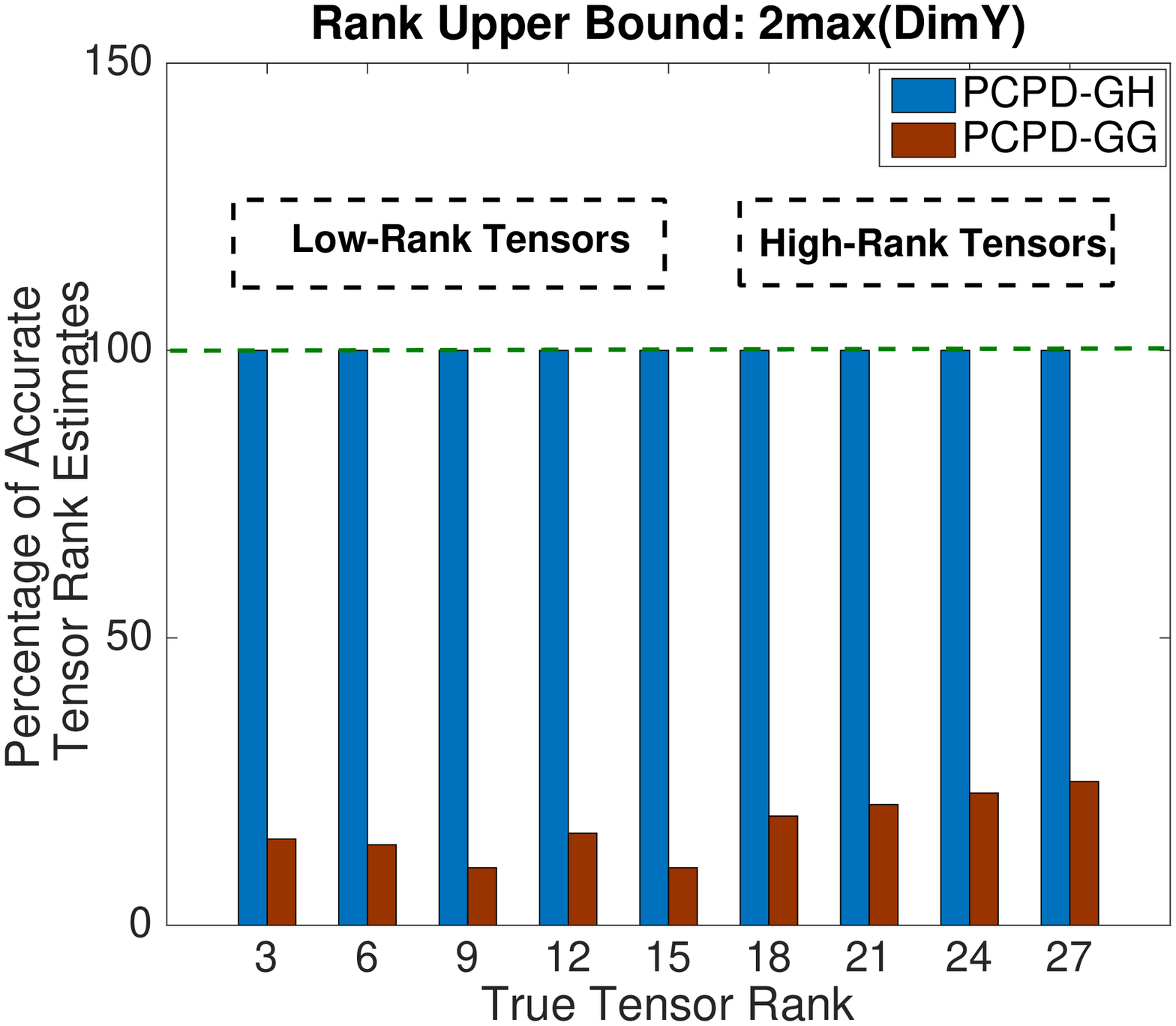}
}
\subfigure[] {
\includegraphics[width=3 in]{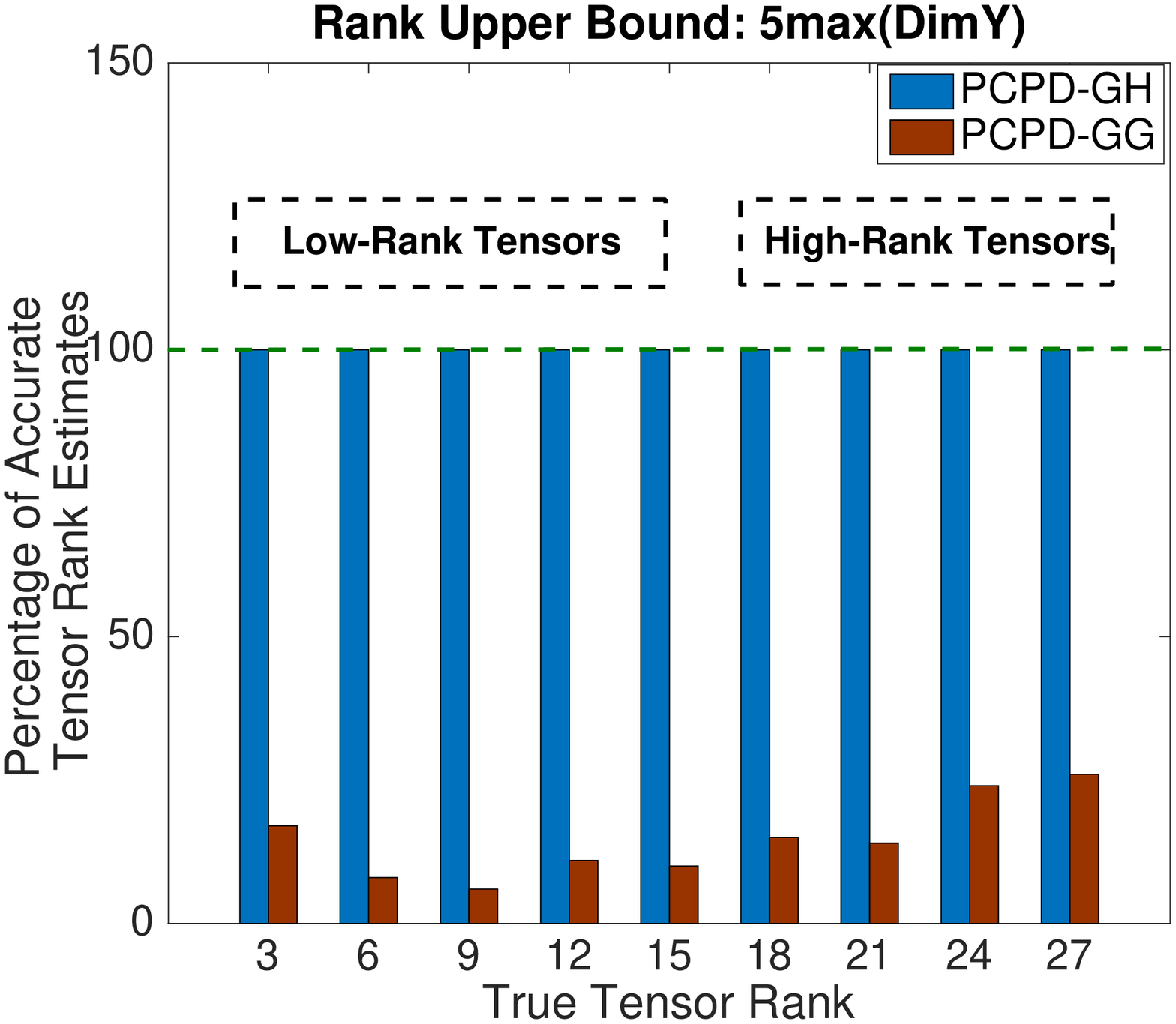}
}
\caption*{ Figure G3. Performance of tensor rank learning for factor matrices with correlations when the rank upper bound is: (a) $ \max \{J_n\}_{n=1}^N$, (b) $2 \max \{J_n\}_{n=1}^N$ and (c) $5\max \{J_n\}_{n=1}^N$.}
\label{fig}
\end{figure}

\begin{figure} [!t]
\setcounter{subfigure}{0}
\centering
\subfigure[] {
\includegraphics[width=3 in]{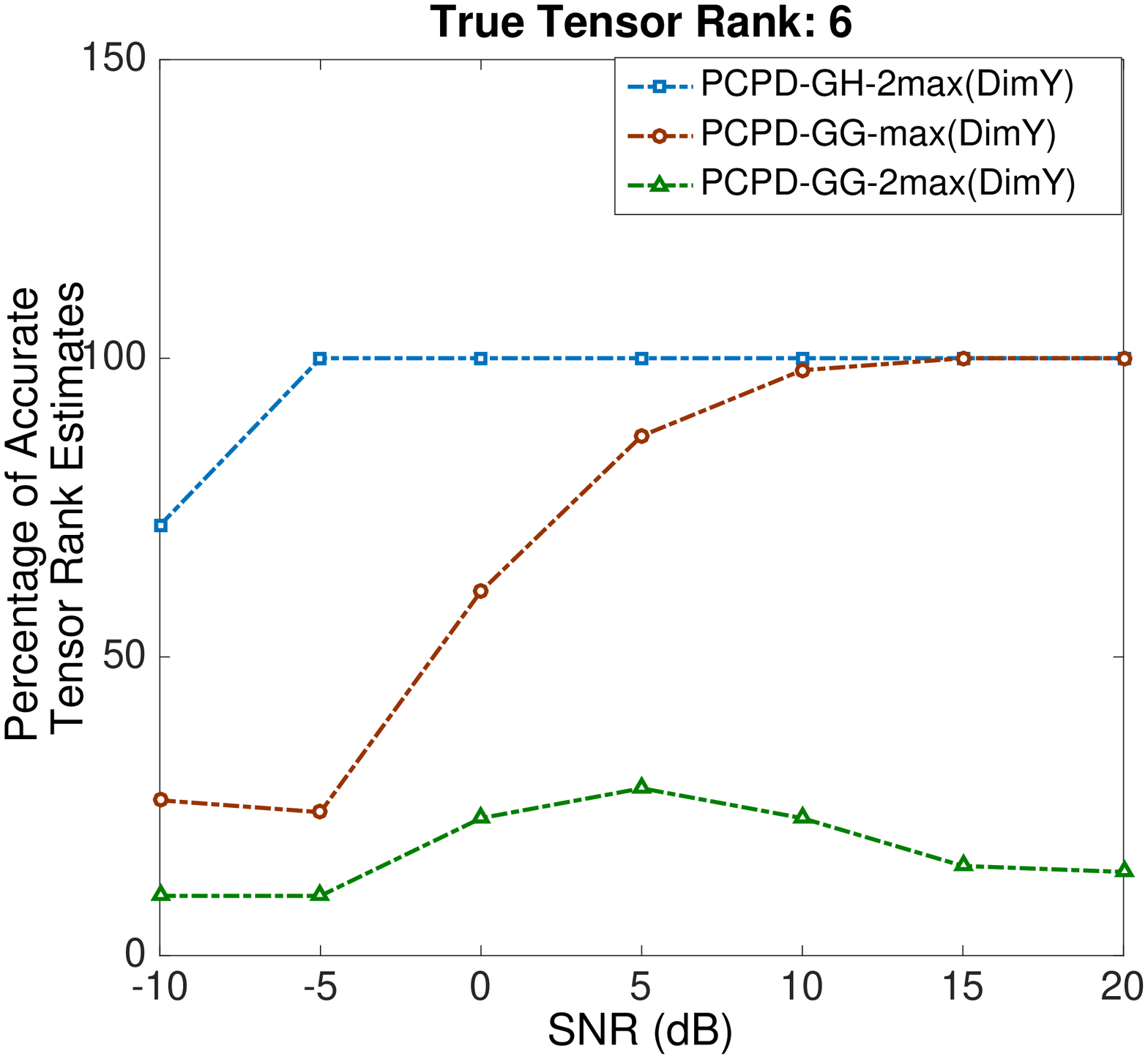}
}
\subfigure[] {
\includegraphics[width=3 in]{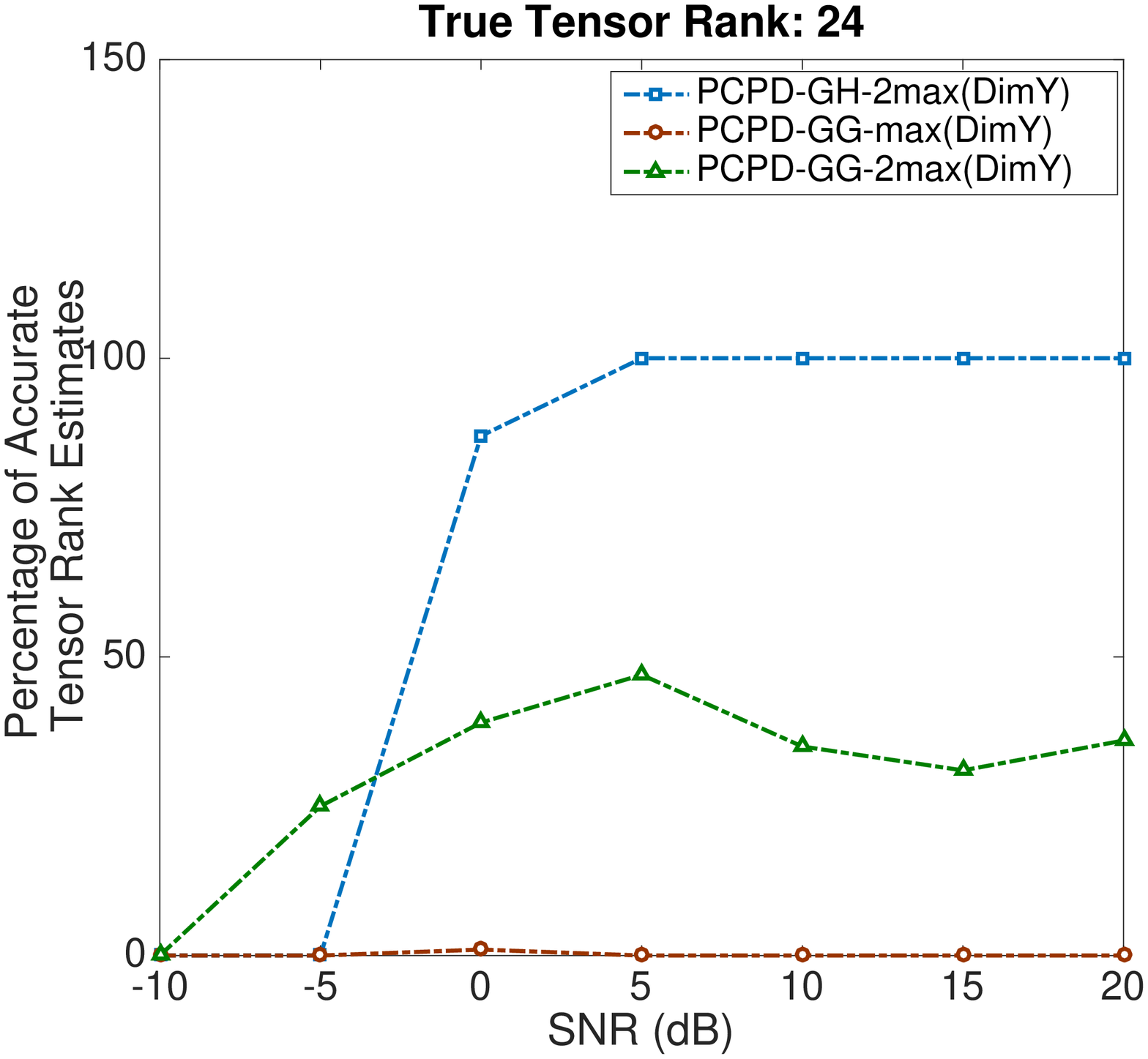}
}
\caption*{ Figure G4. Performance of tensor rank learning for factor matrices with correlations versus different SNRs: (a) low-rank tensors and (b) high-rank tensors.}
\label{fig}
\end{figure}

\begin{table}[!t]
\centering
\caption*{ Table H1: Fit values and estimated tensor ranks of fluorescence data under different SNRs (with rank upper bound $2\max\{J_n\}_{n=1}^N$)}
\begin{tabular}{@{}|c|c|c|c|c|@{}}
\toprule
\multirow{2}{*}{\begin{tabular}[c]{@{}c@{}}SNR\\ (dB)\end{tabular}} & \multicolumn{2}{c|}{PCPD-GG}                                                 & \multicolumn{2}{c|}{PCPD-GH}                                                        \\ \cline{2-5} 
                                                                    & Fit Value & \begin{tabular}[c]{@{}c@{}}Estimated \\ Tensor Rank\end{tabular} & Fit Value        & \begin{tabular}[c]{@{}c@{}}Estimated \\ Tensor Rank\end{tabular} \\ \midrule
-10                                                                 & 71.8197   & 4                                                                & \textbf{72.6401} & \textbf{3}                                                       \\ \midrule
-5                                                                  & 83.5101   & 4                                                                & \textbf{84.3424} & \textbf{3}                                                       \\ \midrule
0                                                                   & 90.3030   & 5                                                                & \textbf{90.8433} & \textbf{3}                                                       \\ \midrule
5                                                                   & 94.0928 & 5                                                                & \textbf{94.3555} & \textbf{3}                                                       \\ \midrule
10                                                                  & 96.0369  & 4                                                       & {\bf 96.0955}          & \textbf{3}                                                       \\ \midrule
15                                                                  & 96.8412   & \textbf{3}                                                       & 96.8432           & \textbf{3}                                                       \\ \midrule
20                                                                  & 97.1197   & \textbf{3}                                                       & 97.1204          & \textbf{3}                                                       \\ \bottomrule
\end{tabular}
\end{table}

\begin{figure}[!t]
\centering
\includegraphics[width= 3.5 in]{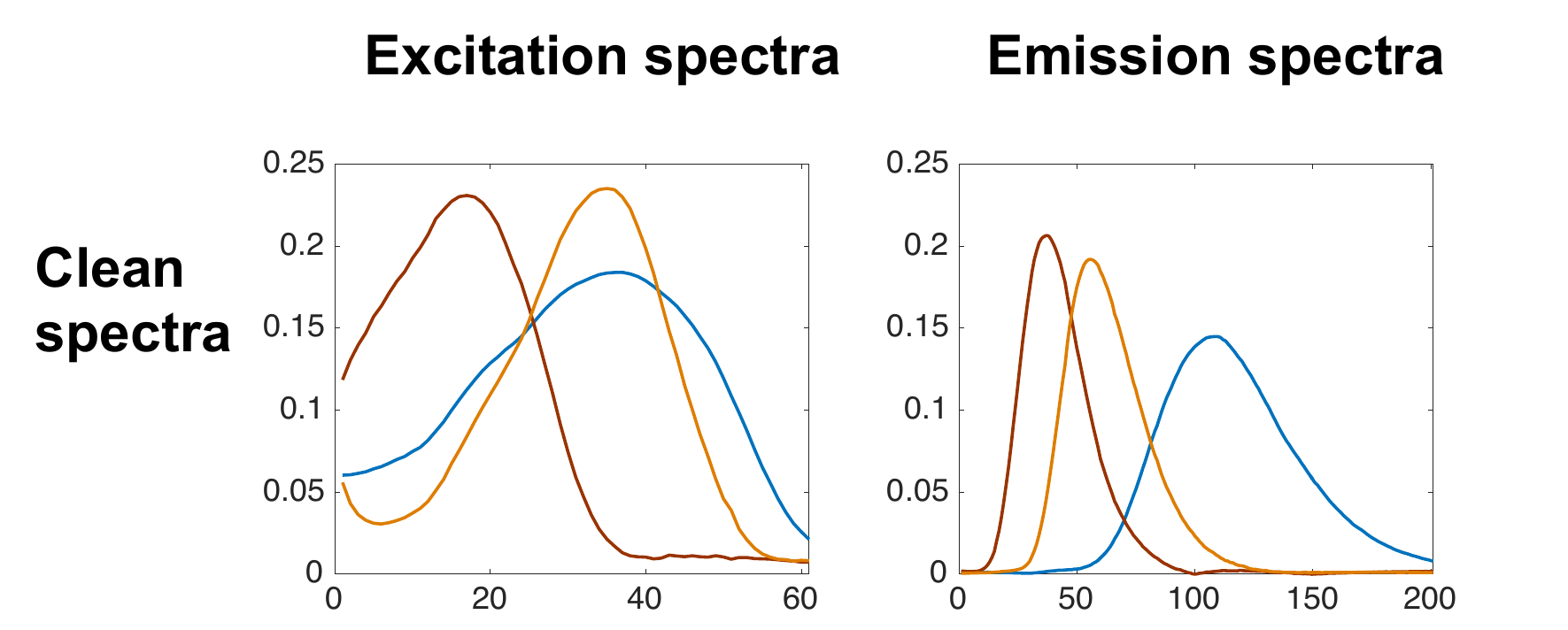}
\caption*{Figure H1: The clean spectra recovered from the noise-free fluorescence  tensor data assuming the knowledge of tensor rank.}
\label{fig_topology}
\end{figure}

\begin{figure*}[!t]
\centering
\includegraphics[width= 7 in]{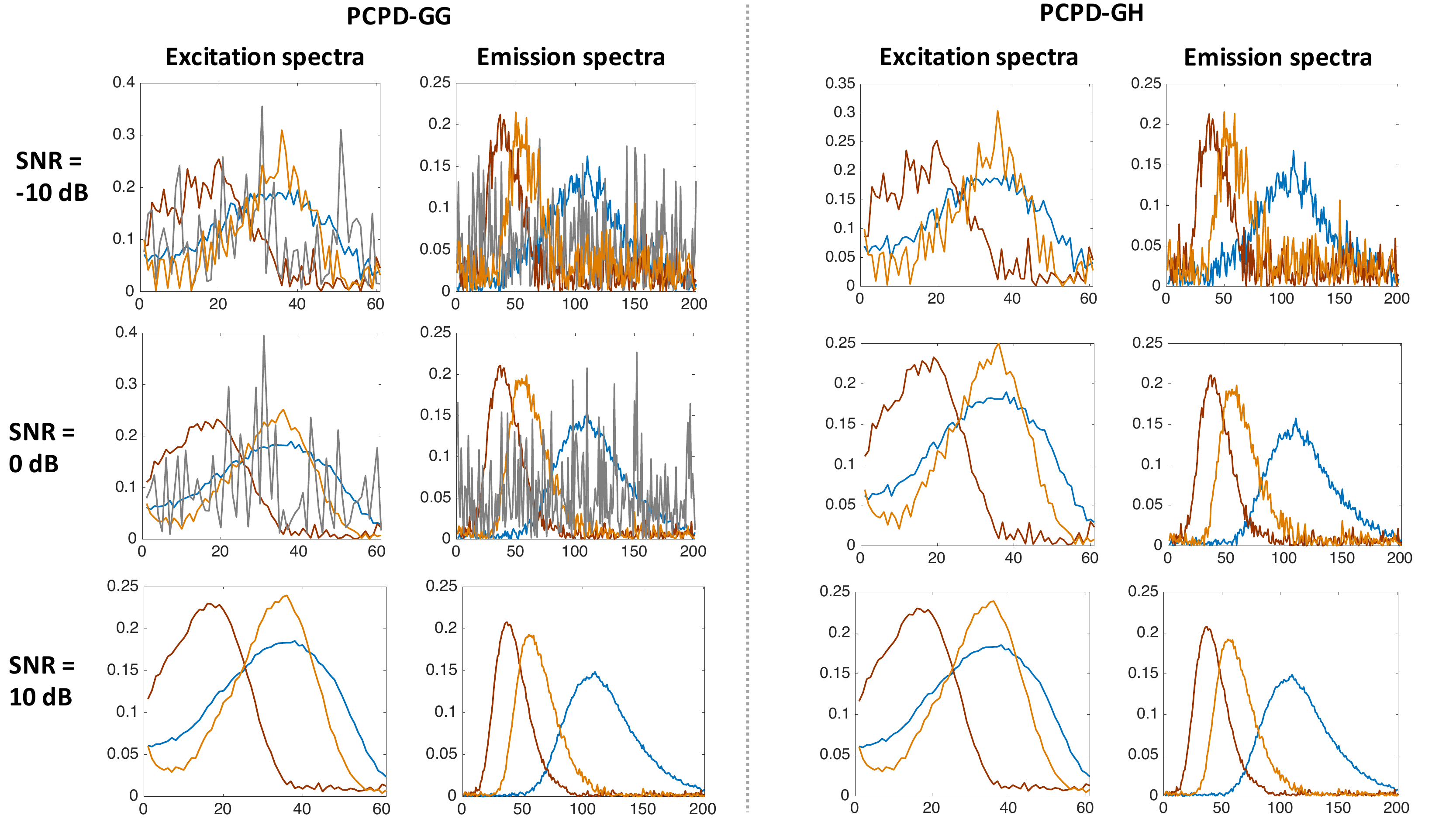}
\caption*{Figure H2: The recovered spectra of fluorescence data under different SNRs.}
\label{fig_topology}
\end{figure*}

\begin{figure*} [!t]
\setcounter{subfigure}{0}
\centering
\subfigure[] {
\includegraphics[width=7  in]{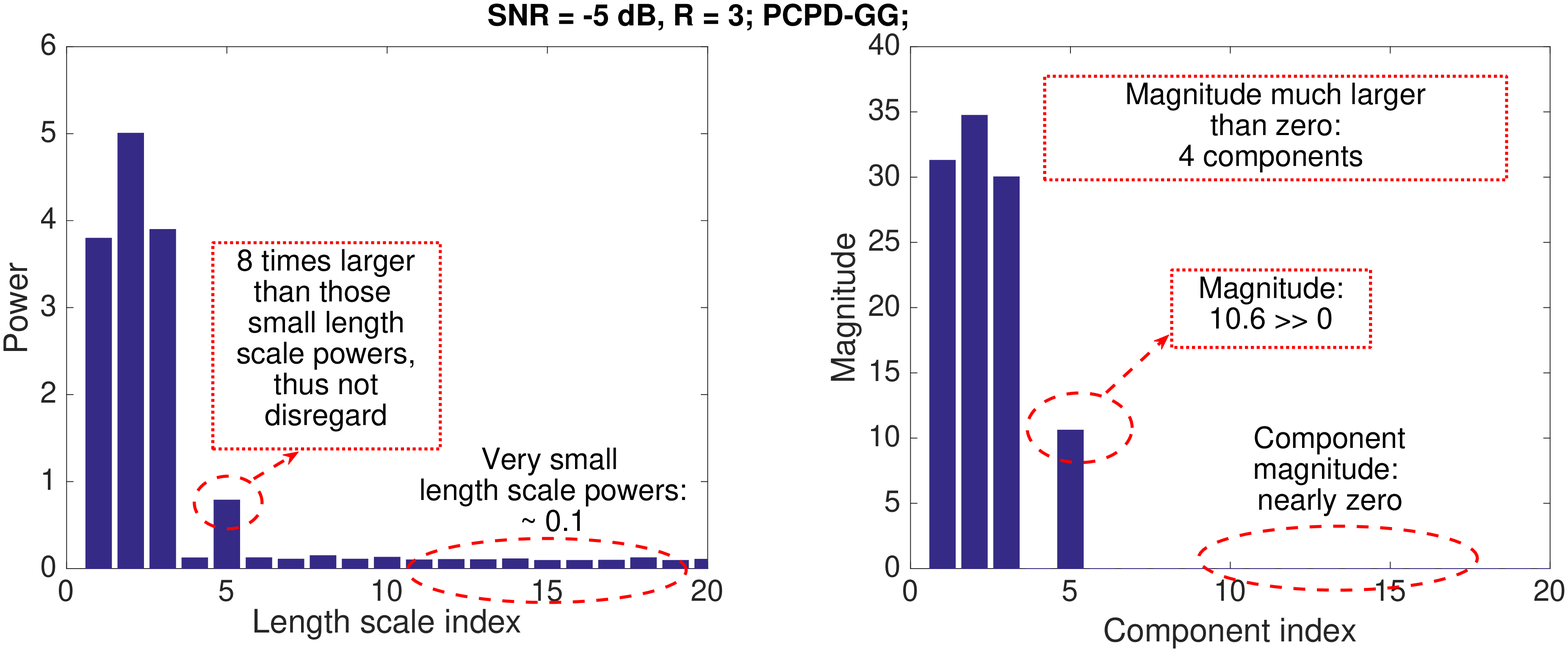}
}
\subfigure[] {
\includegraphics[width=7 in]{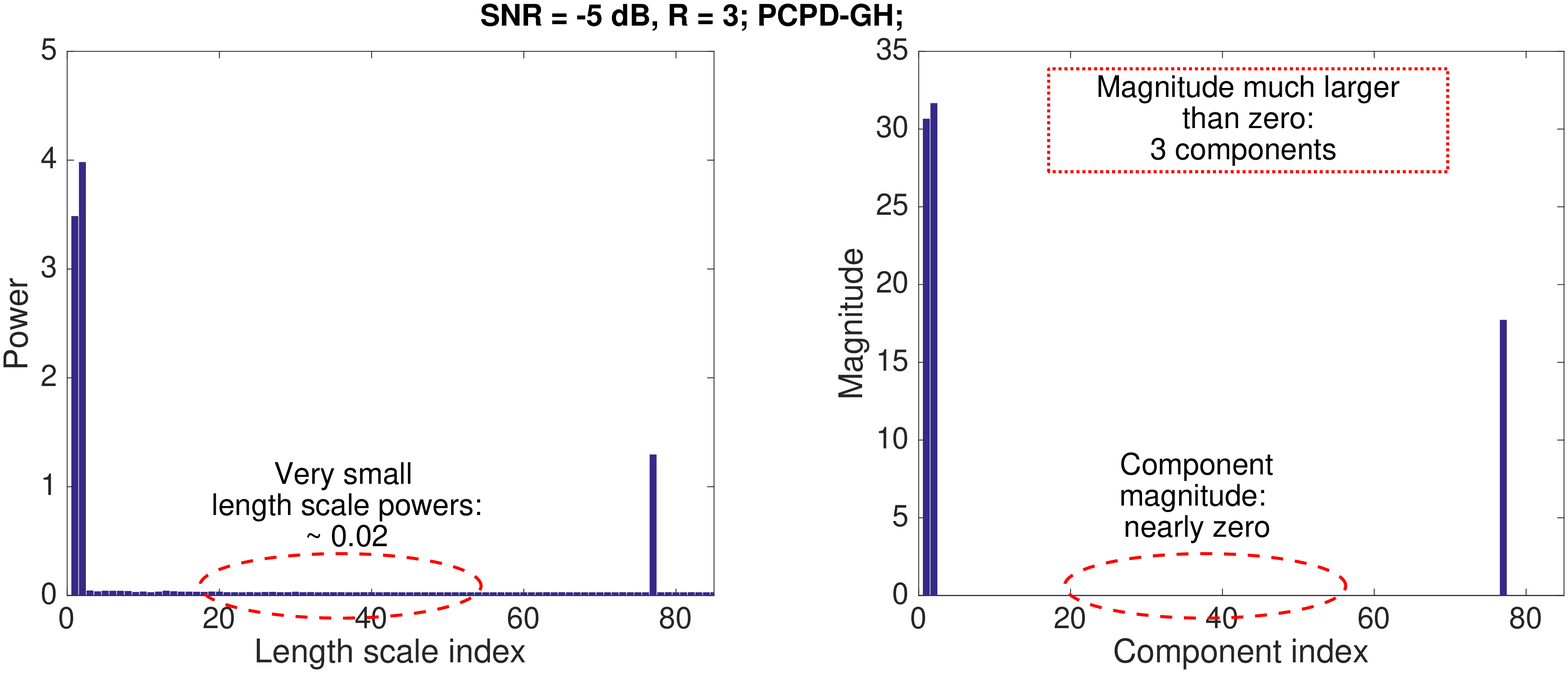}
}

\caption*{Figure H3: Amino acids  fluorescence data analysis. (a) The powers of learnt length scales (i.e., $\{ \gamma_l^{-1}\}_l$) and the magnitudes of associated components for PCPD-GG;  (b) The powers of learnt length scales (i.e., $\{ z_l\}_l$) and the magnitudes of associated components for PCPD-GH.  It can be seen that PCPD-GG recovers 4 components with non-negligible magnitudes, while PCPD-GH recovers 3 components. The two algorithms are with the same upper bound value: 201.  SNR = -5 dB. Since the x-axis is too long (containing 201 points), we only present partial results that include non-zero components. Those values not shown in the figures are all very close to zero.}
\end{figure*}

\begin{figure*}[!t]
\centering
\includegraphics[width= 7 in]{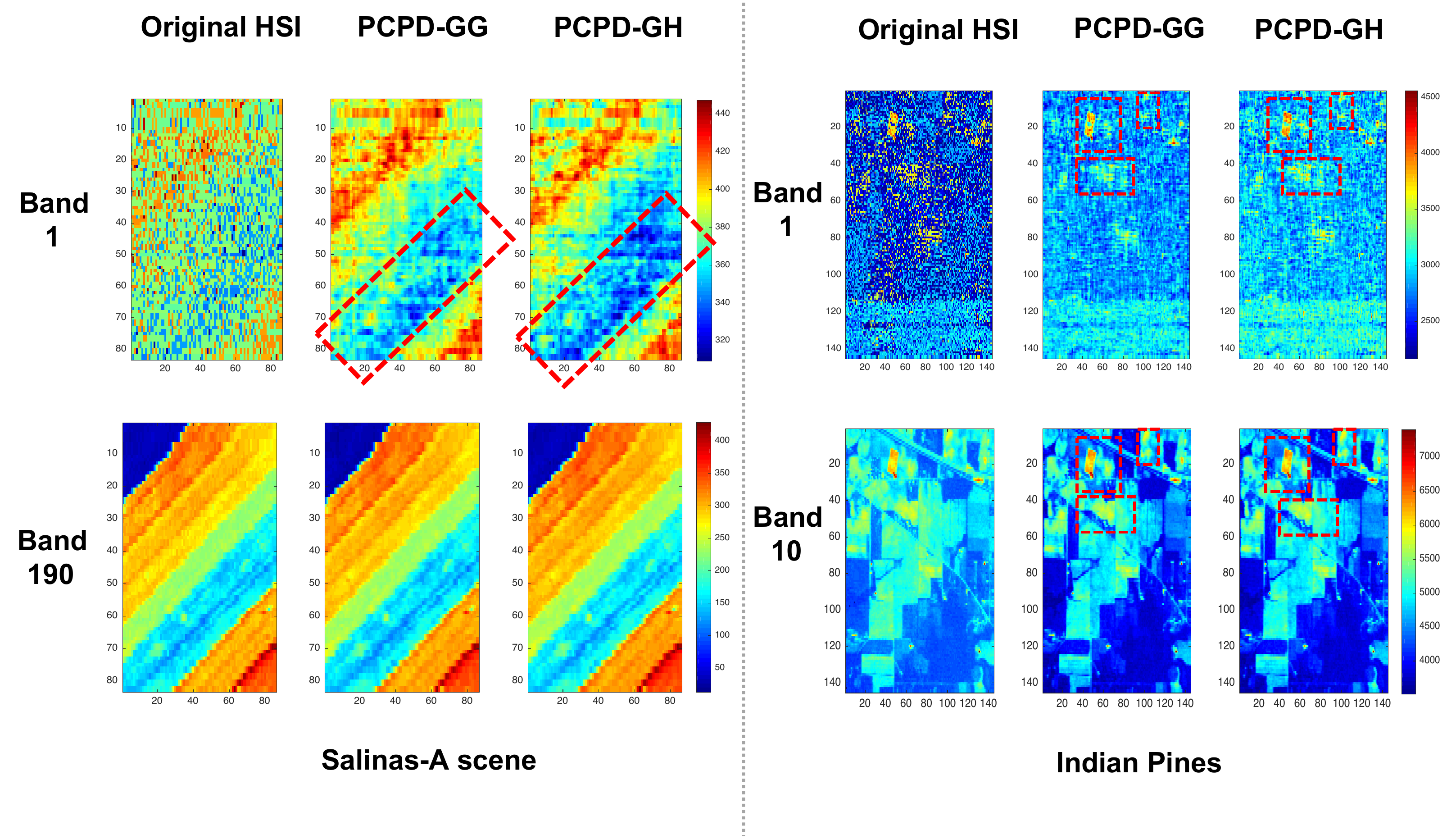}
\caption*{Figure I1: The hyper-spectral image denoising results.}
\label{fig_topology}
\end{figure*}

\section{More Results on Fluorescence Data Analytics}

The benchmarking spectra are presented in Figure H1. The recovered spectra from the two methods are resented in Figure H2  under different SNRs (assuming the rank upper bound value $\max\{J_n\}_{n=1}^N$). When SNR = 10 dB, since the two methods both recover the true tensor rank, the recovered spectra are very similar to the benchmarking results in Figure 10. However, when SNR = 0 dB and -10 dB, the PCPD-GG method gives wrong estimates of the tensor rank. Therefore, its recovered spectra consist of ``ghost" components (in grey color) that has no physical meaning. In contrast, the proposed PCPD-GH method correctly estimate the tensor rank under these two low SNRs, and gives interpretable spectral learning results. 

In Table H1, Fit values and estimated tensor ranks of fluorescence data under different SNRs (with rank upper bound $2\max\{J_n\}_{n=1}^N$) are presented.  

In Figure H3, the learnt length scales from analyzing  amino acids fluorescence dataset are presented. Discussions are given in Section VI. A.

\section{Hyperspectral image denoising}

Samples of denoised HSIs are shown in Figure I1. On the left side of Figure I1, the relatively clean Salinas-A HSI in band 190 is presented to serve as a reference, from which it can be observed that the landscape exhibits ``stripe" pattern. For the noisy HSI in band 1, the denoising results from the two methods using the rank upper bound $\max\{J_n\}_{n=1}^N$ are presented. It is clear that the proposed method recovers better ``stripe" pattern than the PCPD-GG method. Similarly, the results from Indian Pines dataset are presented in the right side of Figure I1. For noisy HSI in band 1,  with the relatively clean image in band 10 serving as the reference, it can be observed that the proposed PCPD-GH method recovers more details than the PCPD-GG method, when both using rank upper bound  $2\max\{J_n\}_{n=1}^N$.  Since the HSIs in band 1 are quite noisy, inspecting the performance difference of the two methods requires a closer look. In Figure I1,  we have used red boxes to highlight those differences. Note that although HSIs in different frequency bands have different pixel intensities (different color bars), they share the same ``clustering'' structure.  The goal of HSI denoising is  to reconstruct the ``clustering'' structure in each band  in order to facilitate the downstream segmentation task [12],[13]. Therefore,  the assessment is based on whether the recovered HSI exhibits correct ``clustering'' patterns.  Specifically, for Salinas-A scene data, the recovered images are supposed to render explicit ``stripe'' patterns, in each of which the intensities (colors) are almost the same. As indicated by the red boxes, it can be observed that  PCPD-GH recovers better ``stripe'' pattern than PCPD-GG, since much more pixels in the red box of PCPD-GH have the same blue color. Similarly, for Indian Pines dataset, as indicated by each red box, the area supposed to be identified as a cluster (with warmer colors than nearby areas) is more accurately captured by PCPD-GH.

{\color{black}
\section{The convergence behavior of the proposed algorithm}
The KL divergence defined in (13) cannot be directly evaluated since the true posterior is unknown. Instead, the VI research usually assess an equivalent measure called evidence lower bound (ELBO) [50], [51] as follows:
\begin{align}
\mathcal L \left( Q\left(\boldsymbol \Theta \right) \right) = \mathbb E \left[\ln p(\mathcal Y, \boldsymbol \Theta)\right] - \mathbb E \left( \ln Q(\boldsymbol \Theta)\right).
\end{align}
We derive its expression in Appendix K.  Minimizing the KL divergence is equivalent to maximizing the ELBO [50], [51].

\begin{figure*} [!t]
\setcounter{subfigure}{0}
\centering
\subfigure[] {
\includegraphics[width=3 in]{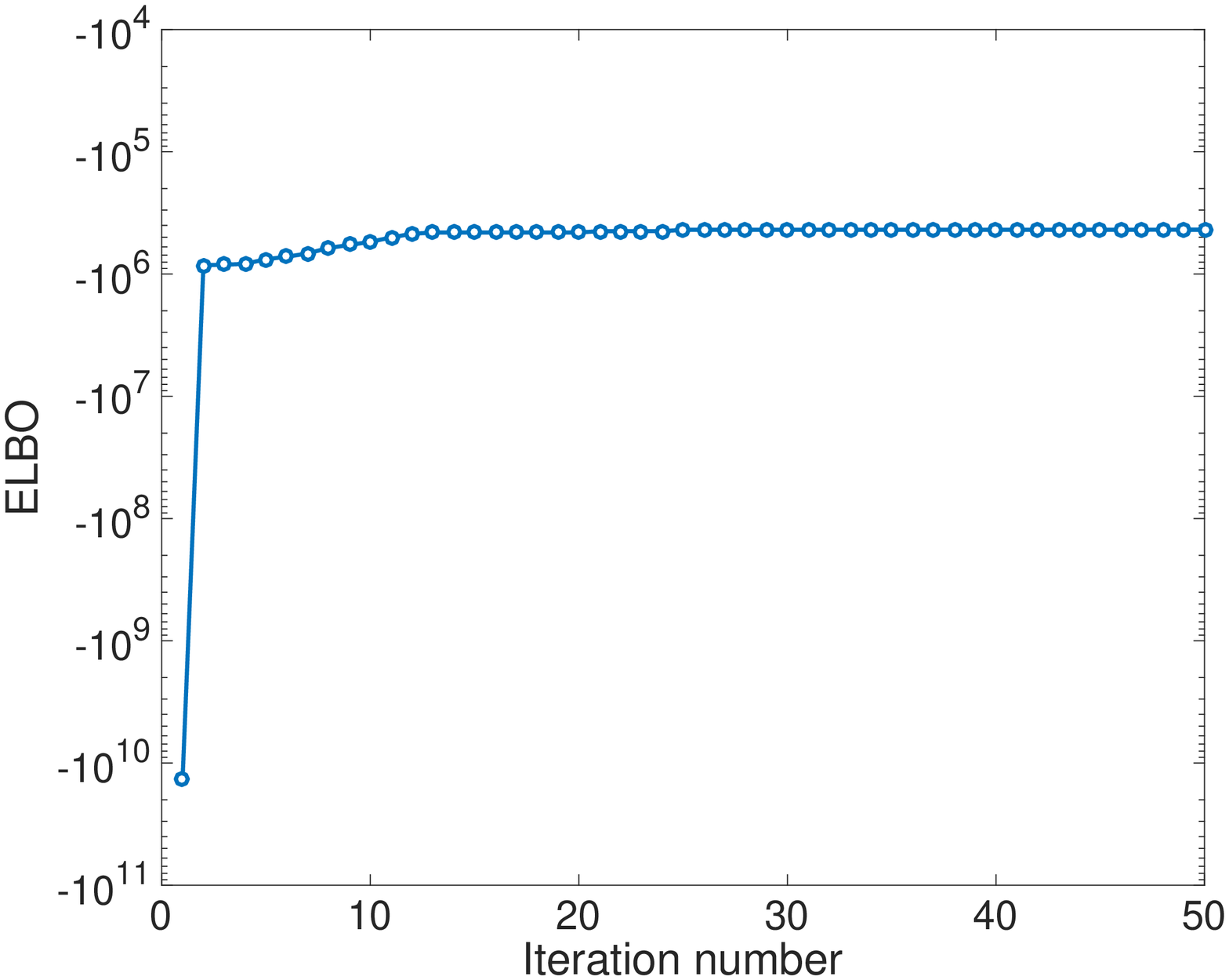}
}
\subfigure[] {
\includegraphics[width=3 in]{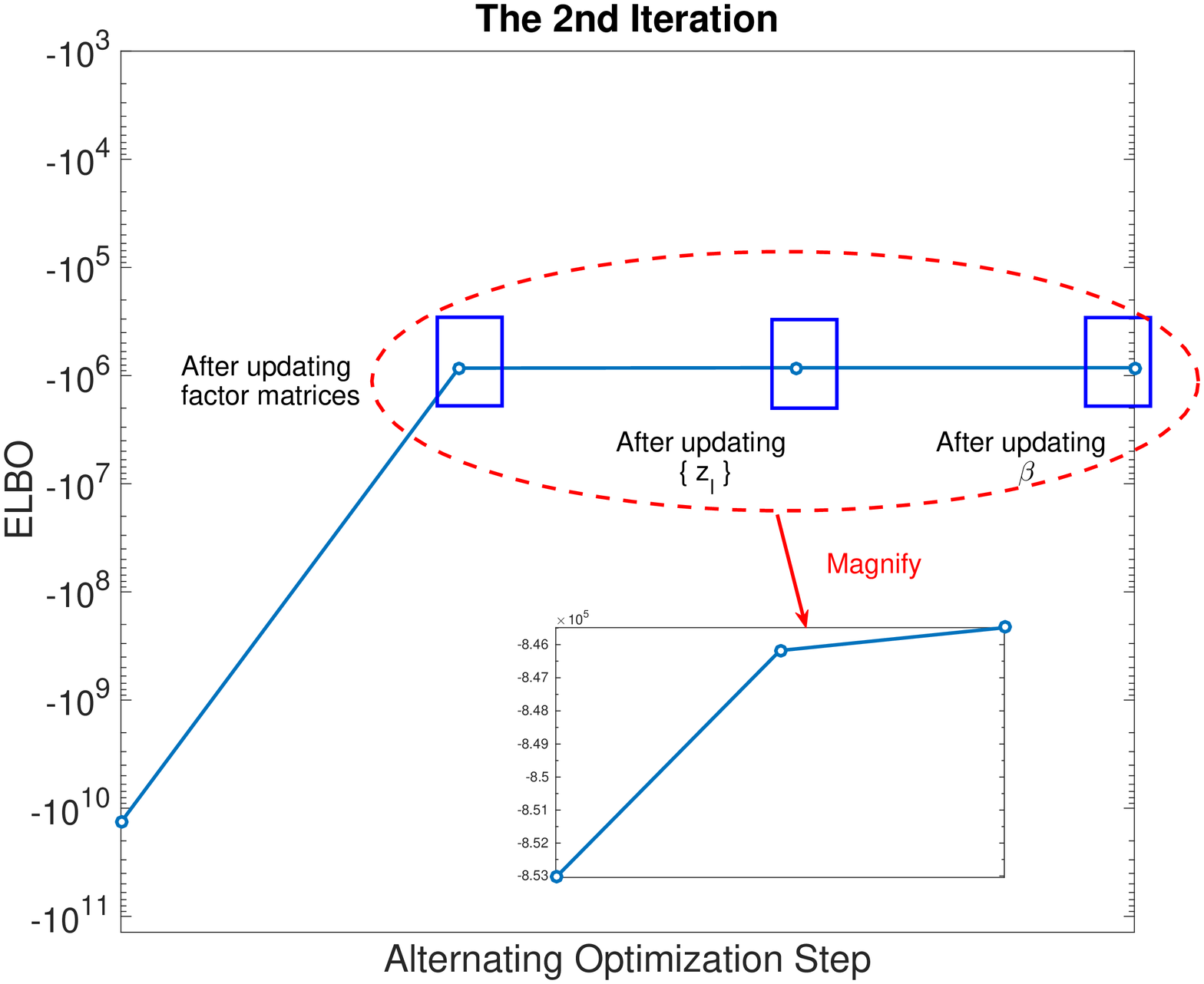}
}
\subfigure[] {
\includegraphics[width=3 in]{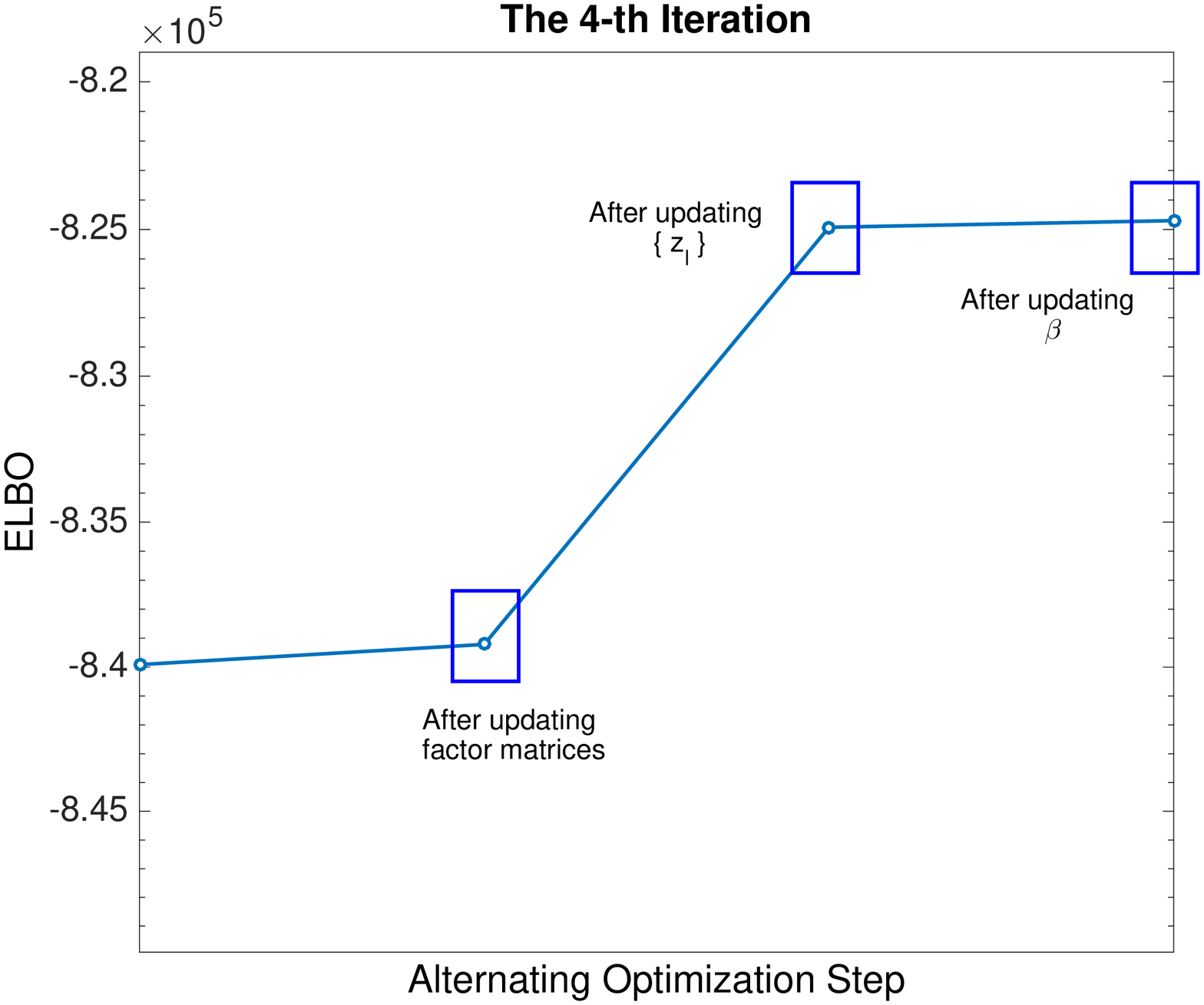}
}
\subfigure[] {
\includegraphics[width=3 in]{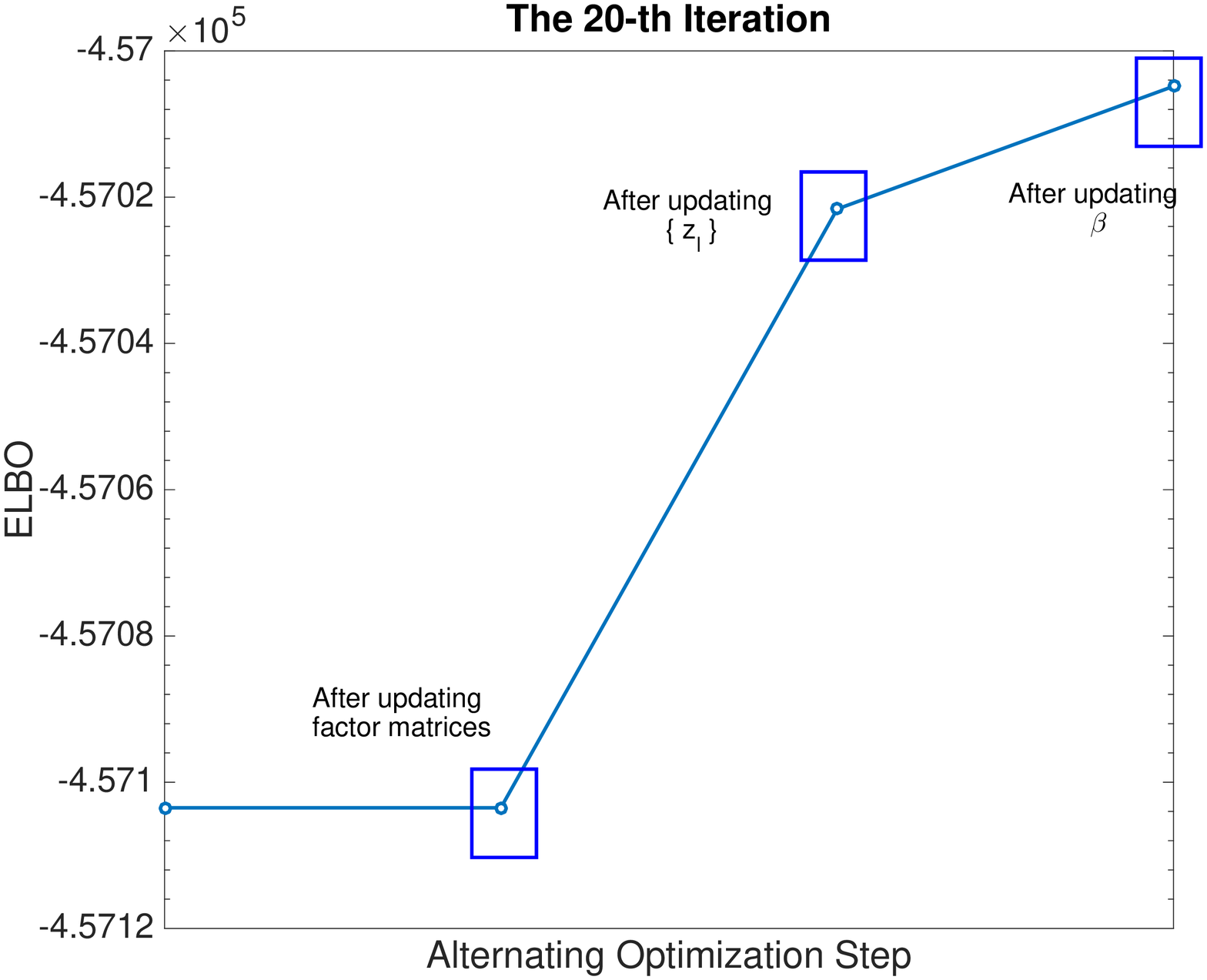}
}
\caption*{Figure J1: (a) The evidence lower bound (ELBO) versus iteration number. The improvement of ELBO after three alternating  update steps in the $2$-nd iteration, the $4$-th iteration, and the $20$-th iteration, see (b)-(d) respectively.}
\label{fig}
\end{figure*}

To show the performance of the proposed algorithm in terms of ELBO, we conduct an experiment following the setting of Section V. A. The true tensor rank $R = 24$ and SNR is  $5$ dB. As demonstrated in Figure J1 (a), the proposed algorithm monotonically increases the ELBO as the iteration number increases. It can be observed that the algorithm significantly improves the ELBO in the first $5$ iterations, and gradually converges after $20$ iterations.  Particularly, we show how the three alternating update steps improve the ELBO in the $2$-nd, the $4$-th, and the $20$-th iteration, as seen in Figure J1 (b)-(d). It can be seen that in the very first iteration, e.g., the $2$-nd iteration, the improvement brought by updating factor matrices dominates those from updating other variational pdfs. Later, when the factor matrices tend to converge (see Figure J1 (c)-(d)), the update  of column variances $\{z_l\}$ contributes the most. }

{\color{black}
\section{Evidence Lower Bound (ELBO)}

According to the definition of ELBO (46), it is computed by
\begin{align}
&\mathcal L (Q(\boldsymbol \Theta)) =  \mathbb{E} \left[ \ln p(\mathcal{Y}, \boldsymbol{\Theta})\right] - \mathbb{E} \left[ \ln Q\left( \boldsymbol{\Theta} \right) \right]  \nonumber \\
&= \mathbb{E} \left[ \ln p( \mathcal{Y} | \left\{ \boldsymbol{ U}^{(n)} \right\}_{n=1}^{N}, \beta ) \right] \nonumber \\
& + \mathbb{E} \left[ \ln p( \left\{ \boldsymbol{ U}^{(n)} \right\}_{n=1}^{N} | \left\{ z_l \right\}_{l=1}^{L} ) \right]  \nonumber \\
&+ \mathbb{E} \left[ \ln p( \left\{ z_l \right\}_{l=1}^{L}) \right]   + \mathbb{E} \left[ \ln p( \beta ) \right] - \mathbb{E} \left[ \ln Q\left( \left\{ \boldsymbol{ U}^{(n)} \right\}_{n=1}^{N} \right) \right]  \nonumber \\
& - \mathbb{E} \left[ \ln Q\left( \left\{ z_l \right\}_{l=1}^{L} \right) \right] - \mathbb{E} \left[ \ln Q\left( \beta \right) \right]. 
\end{align}
By taking expectations with respect to the corresponding variational pdfs and organizing the resulting terms,  the ELBO takes the following expression: 
\begin{align}
&\mathcal L (Q(\boldsymbol \Theta))  \nonumber \\
&= - \frac{\prod_{n=1}^N J_n}{2} \ln(2 \pi) +  \frac{\prod_{n=1}^N J_n}{2} \mathbb{E} \left[ \ln \beta \right] \nonumber \\
&- \frac{\mathbb{E} \left[ \beta \right]}{2} \mathbb{E} \left[ \parallel \mathcal Y - \llbracket  \boldsymbol U^{(1)},  \boldsymbol U^{(2)},..., \boldsymbol U^{(N)}  \rrbracket   \parallel_F^2 \right] \nonumber \\
&- \frac{L \sum_{n=1}^N J_n}{2} \ln(2 \pi) + \sum_{n=1}^{N} \Bigg\{ -\frac{J_n}{2} \sum_{l=1}^{L} \mathbb{E} \left[ \ln (z_l) \right] \nonumber \\
& -\frac{1}{2} \mathrm{Tr} \left( \mathbb{E} \left[ \boldsymbol{Z}^{-1} \right] \mathbb{E} \left[ \boldsymbol{ U}^{(n)^{T}} \boldsymbol{ U}^{(n)} \right] \right) \Bigg\} \nonumber \\
& + \sum_{l=1}^{L} \Bigg\{ \frac{\lambda_l^0}{2} \ln \left( \frac{a_l^0}{b_l^0} \right) - \ln \left( 2K_{\lambda_l^0} \left(\sqrt{a_l^0 b_l^0} \right) \right) \nonumber \\
&+ (\lambda_l^0-1) \mathbb{E} \left[ \ln (z_l) \right] - \frac{a_l^0}{2} \mathbb{E} \left[ z_l \right] - \frac{b_l^0}{2} \mathbb{E} \left[ z_l^{-1} \right] \Bigg\} \nonumber \\
& - \ln (\Gamma (\epsilon) ) + \epsilon \ln (\epsilon) + (\epsilon-1) \mathbb{E} \left[ \ln(\beta) \right] - \epsilon \mathbb{E} \left[ \beta \right] \nonumber \\
&+ \sum_{n=1}^{N} \Bigg\{ \frac{J_n}{2} \ln \det (\boldsymbol{ U}^{(n)}) + \frac{J_n R}{2} (1+\ln(2 \pi))\Bigg\} \nonumber\\
&  +\sum_{l=1}^{L} \Bigg\{ \frac{1}{2} \ln \left( \frac{b_l}{a_l} \right) + \ln \left( 2K_{\lambda_l} \left(\sqrt{a_l b_l} \right) \right) \nonumber \\
& - (\lambda_l - 1) \frac{ \frac{d}{d\nu}\left[ K_{\nu} \left(\sqrt{a_l b_l} \right) \right]_{\nu=\lambda_l} }{K_{\lambda_l} \left(\sqrt{a_l b_l} \right)}  \nonumber \\
& + \frac{\sqrt{a_l b_l}}{2K_{\lambda_l} \left(\sqrt{a_l b_l} \right)}  K_{\lambda_l+1} \left(\sqrt{a_l b_l} \right) \nonumber\\
& + \frac{\sqrt{a_l b_l}}{2K_{\lambda_l} \left(\sqrt{a_l b_l} \right)} K_{\lambda_l-1} \left(\sqrt{a_l b_l} \right) \Bigg\} \nonumber \\
& + \ln (\Gamma (e)) - (e-1) \psi(e) - \ln (f) + e. 
\end{align}
In (48), the expressions of  the expectation  $\mathbb{E} \left[ z_l^{-1} \right]$,  $\mathbb{E} \left[ z_l \right]$, and $ \mathbb{E} \left[ \beta\right]$ can be found in Table II of the manuscript. For other expectation terms, we provide their expressions in the following:
\begin{align}
&\mathbb E \left[  \ln \beta \right] = \psi(e) -\ln (f)  \\
&\mathbb E \left[  \ln z_l \right] =   \ln \frac{\sqrt{b}}{\sqrt{a}}  \frac{d}{d\nu}\left[ K_{\nu} \left(\sqrt{a_l b_l} \right) \right]_{\nu=\lambda_l} . 
\end{align}
In (47)-(50), $\psi(\cdot)$ denotes the digamma function and $\Gamma(\cdot)$ denotes the Gamma function.}

{\color{black} 

\section{Tensor Rank Estimation Versus Different Sparsity Levels}
\begin{figure*} [!t]
\setcounter{subfigure}{0}
\centering
\subfigure[] {
\includegraphics[width=2.9  in]{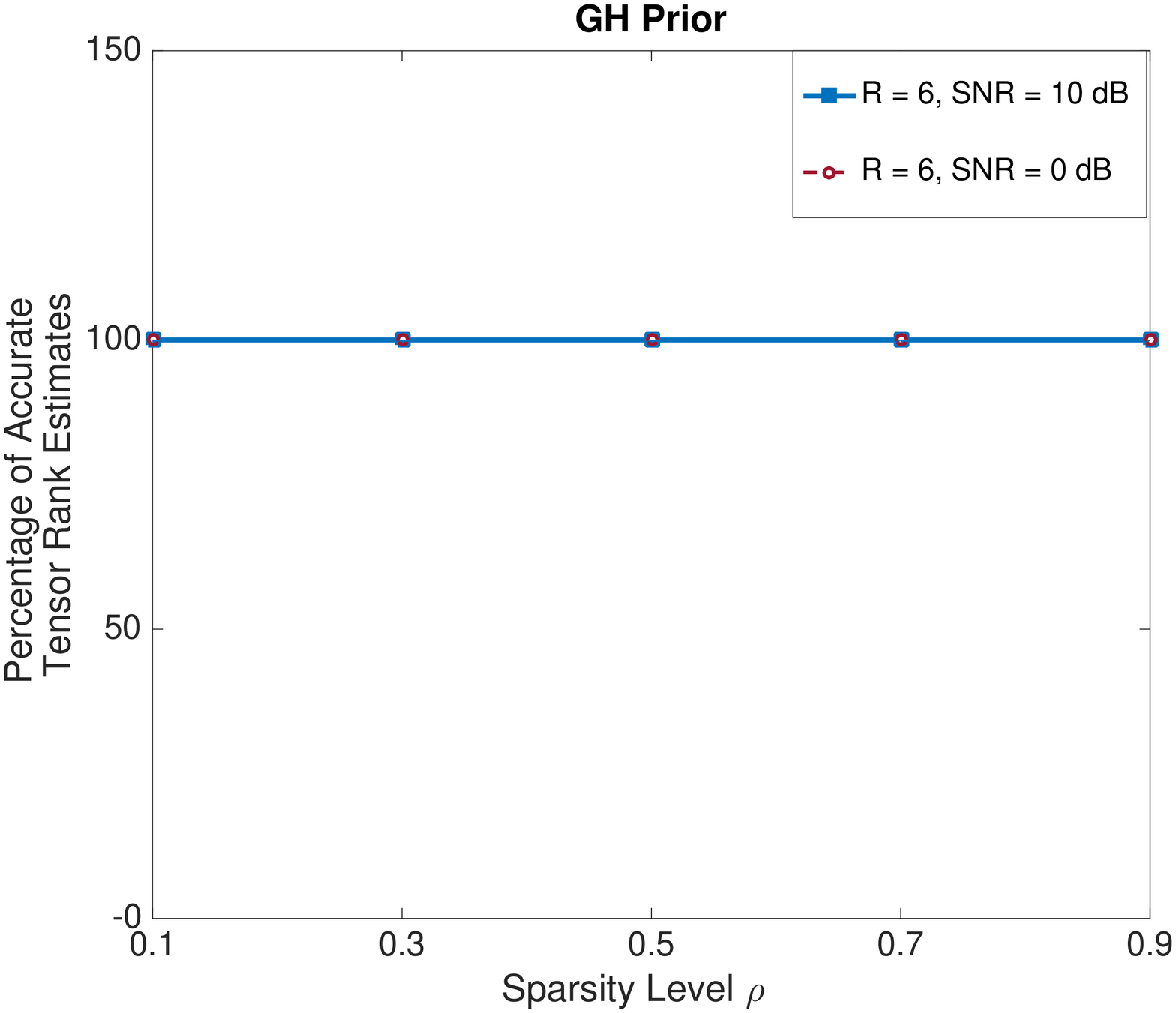}
}
\subfigure[] {
\includegraphics[width=2.9 in]{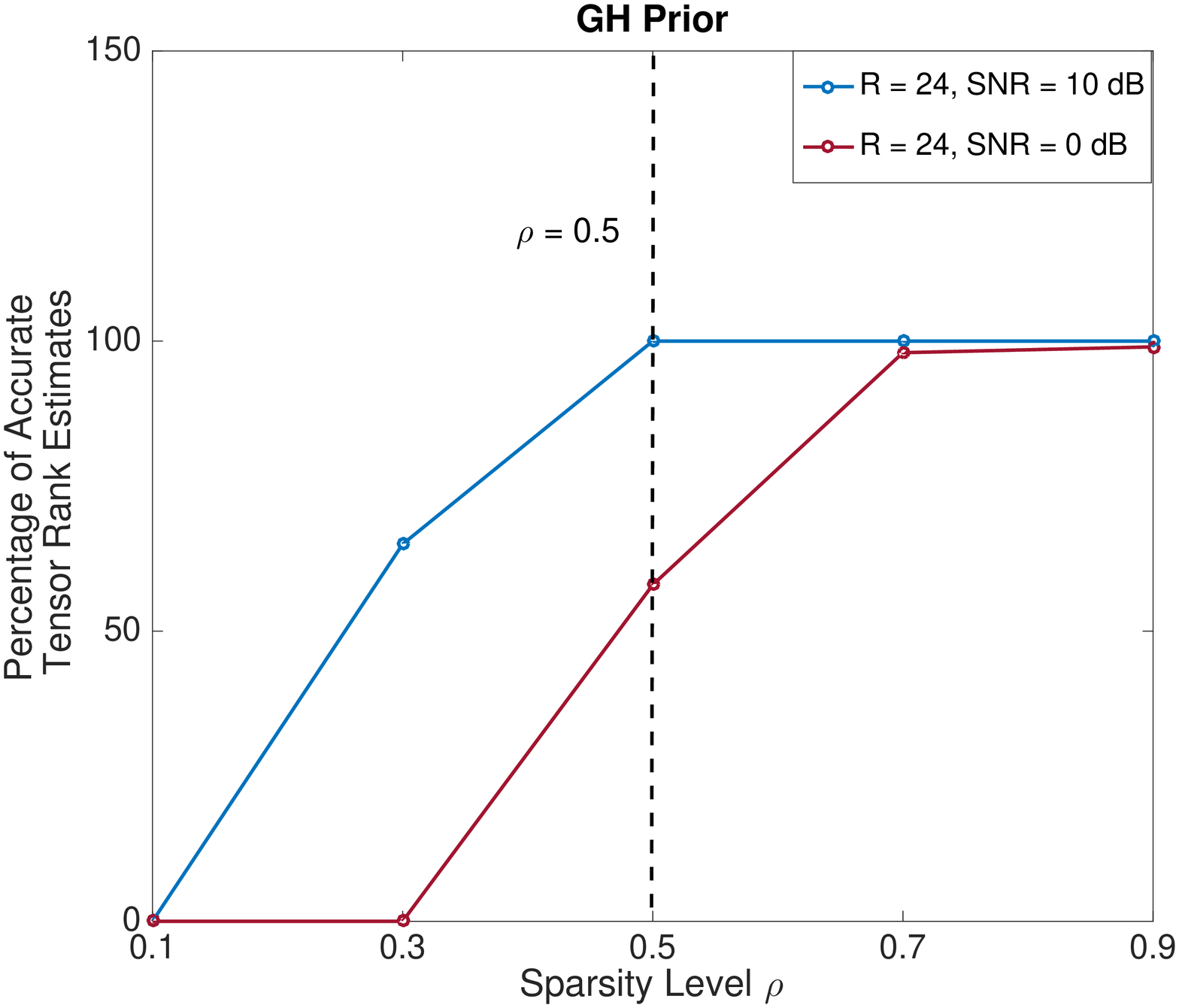}
}
\subfigure[] {
\includegraphics[width=2.9 in]{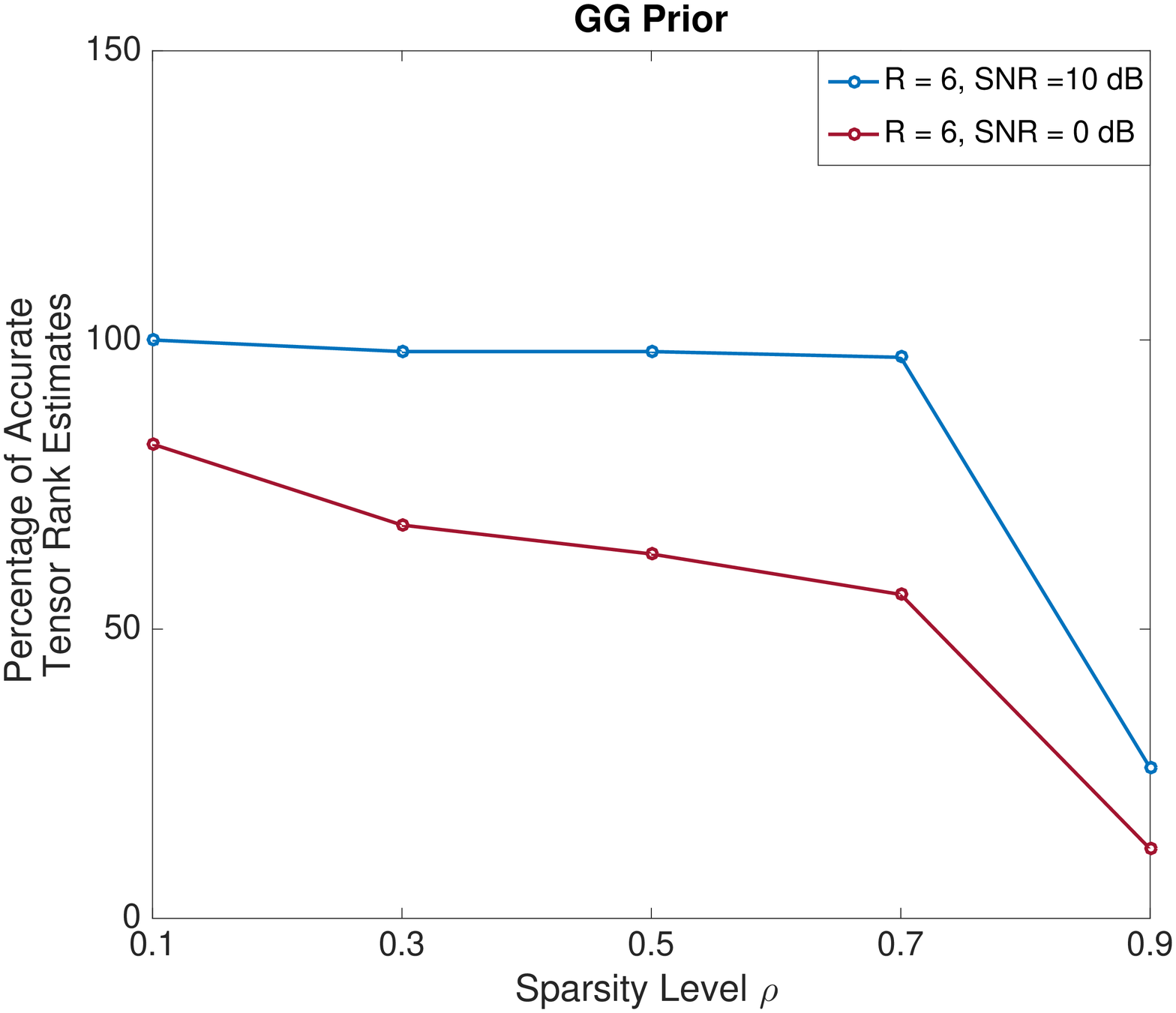}
}
\subfigure[] {
\includegraphics[width=2.9 in]{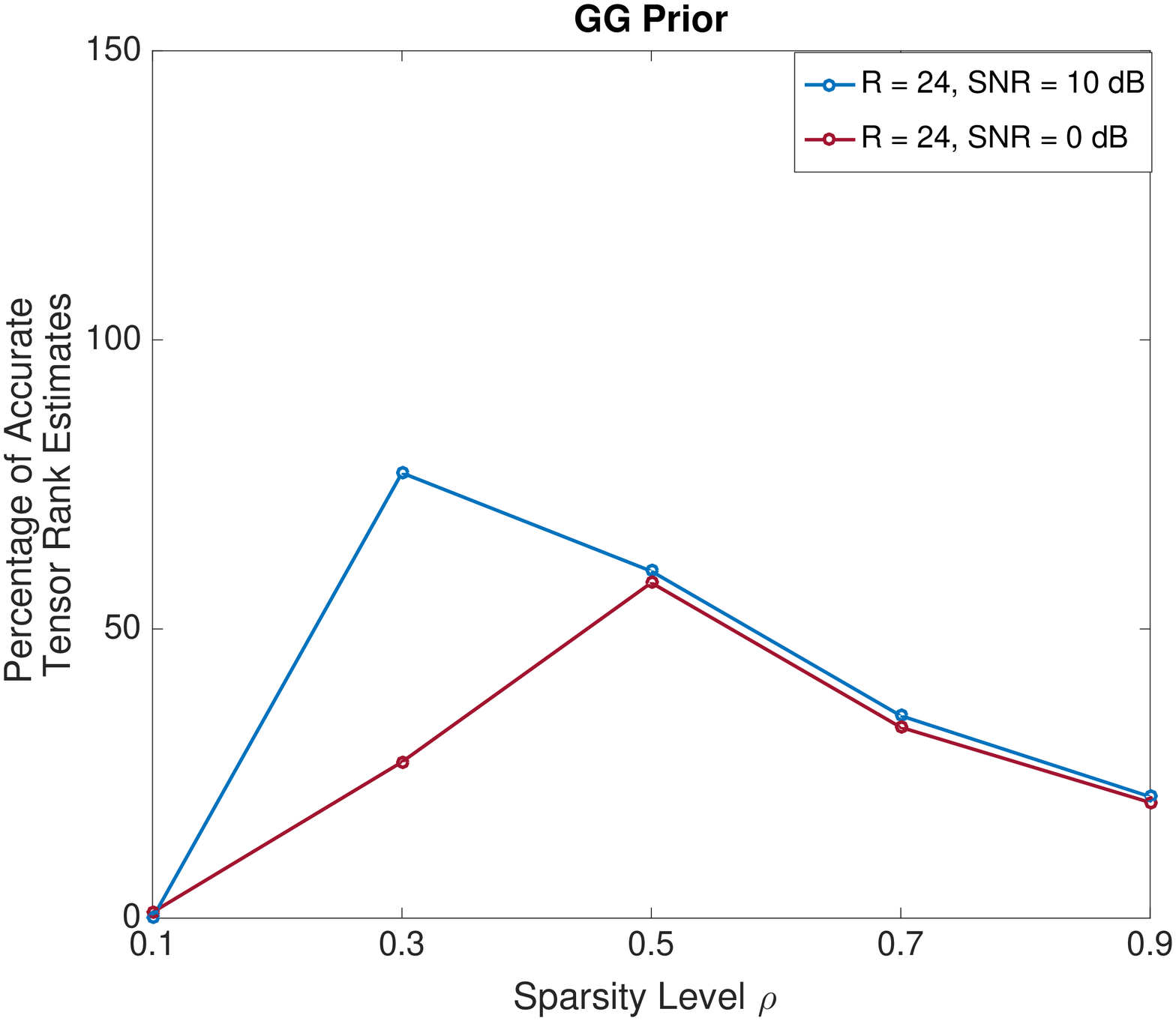}
}
\caption*{Figure L1: Performance of tensor rank learning versus different sparsity levels for GH prior (a)-(b)  and GG prior (c)-(d).   }
\label{fig}
\end{figure*}

To quantify what sparsity level the GH prior can handle, we run the experiment of Section V. A with different sparsity level $\rho = \frac{L-R}{L}$.  In Figure L1, we present the tensor rank estimation accuracies of the proposed PCPD-GH algorithm and the PCPD-GG algorithm [22] versus different sparsity levels. Particularly, given a tensor rank $R$, the upper bound value is set as $ L = \text{ceil}(\frac{R}{1-\rho})$ according to the definition of sparsity level, where $\text{ceil}(\cdot)$ denotes the ceil function. We consider  the low-rank case ($R=6$) and the high rank case ($R=24$) at both low SNR  (SNR = 0 dB) and high SNR (SNR = 10 dB). For the proposed PCPD-GH algorithm, when the true tensor rank is low, e.g., $R=6$, the tensor rank learning performance achieves $100 \%$ accuracy among different values of $\rho = \{ 0.1, 0.3, 0.5, 0.7, 0.9\}$, under both SNR = 0 dB and SNR = 10 dB, see Figure L1(a). Therefore, GH prior can lead to successful tensor rank learning no matter the sparsity level is if the true tensor rank is low.  On the other hand, when the true tensor rank is high, e.g., $R=24$, it can be observed in Figure L1 (b) that the accuracy of tensor rank learning degenerates significantly when $\rho < 0.5$. Therefore, a reasonable bound on the sparsity level for the GH-prior to work well is $0.5$ when the true rank is high. The theoretical justification for the value of lower bound $\rho = 0.5$ for different tensor ranks and/or SNRs is an interesting future work. For comparison, in Figure L1 (c)-(d), we also present the tensor rank learning accuracies of the PCPD-GG algorithm. It can be seen that the proposed algorithm performs better in a wider range of sparsity level $\rho$, showing its flexibility to adapt to various sparsity levels. }

{\color{black}
\section{More Discussions on Mean-field VI}

The derivation of mean-field VI algorithm is a little bit tedious, and sometimes has no closed-form, see, e.g., [64].  But once the closed-form VI updates can be successfully derived, the resulting inference algorithm is known to be computationally efficient and convergence guaranteed. The accuracy of inference is also satisfactory in various data analytic tasks [48]-[51]. Theses advantages make mean-field VI still a popular choice for inference algorithm development. Other recent advances can be found in the review paper [50].

Before deriving mean-field VI algorithm, we should have a close look at the priors and the likelihood of the probabilistic model. If each probability density function (pdf) pair in the probabilistic model forms a conjugate pair in the exponential distribution family (see detailed rules in [51]), we can conclude that each mean-field VI update has a closed-form expression. For the tensorized neural networks in [64], due to the non-linearities introduced by the deep neural network model, the prior is not conjugate to the likelihood. Therefore, there is no need to try the mean-field VI. Instead, one can resort to some recent VI advances [50],  such as stein VI [52] or black-box VI [69], that address non-conjugate probabilistic models.}

{\color{black}
\section{Discussions on the pruning of the proposed algorithm}
Note that the pruning of the proposed algorithm would be permanent. If the algorithm fortunately jumps out from one inferior local minima, the columns once deemed ``irrelevance'' might recover its importance. To address this, the birth process, which is opposite to the pruning process (called death process) can be resorted to \cite{beal, RJMCMC}. Exploiting such  schemes might further improve the tensor rank learning capability, especially in very low SNR an/or very high tensor rank regimes. However, from the extensive experiments conducted in this paper, the issue does not frequently appear in a wide range of SNRs and tensor ranks.}

{\color{black}
\section{More discussions on future research}

To have more rigorous  analysis on the behavior of the Bayesian tensor CPD algorithm with various priors, one can analyze the properties of local minima or global minima of an equivalent optimization problem (with GH prior or Gaussian-gamma prior serving as a regularizer). Similar attempts have been achieved in the research of linear regression \cite{F1} and matrix decomposition \cite{F2}. However, generalizing their analysis to the tensor decomposition is non-trivial and has not been reported so far, which is an interesting future direction to investigate. }

\section{The learnt length scales of PCPD-GG and PCPD-GH}
In Figure P1-P3, the learnt length scales of PCPD-GG and PCPD-GH under different cases are presented.  Discussions are in Section V. B.

\begin{figure*} [!t]
\setcounter{subfigure}{0}
\centering
\subfigure[] {
\includegraphics[width=7 in]{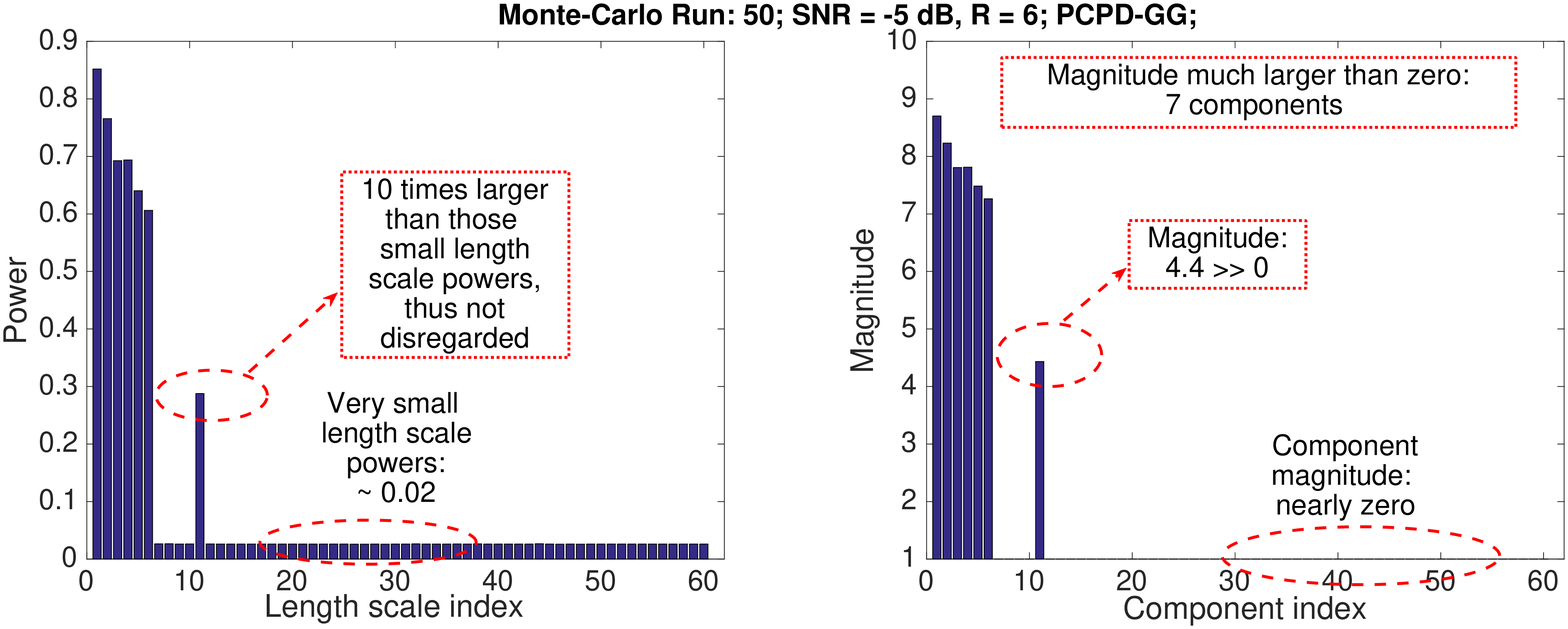}
}
\subfigure[] {
\includegraphics[width=7 in]{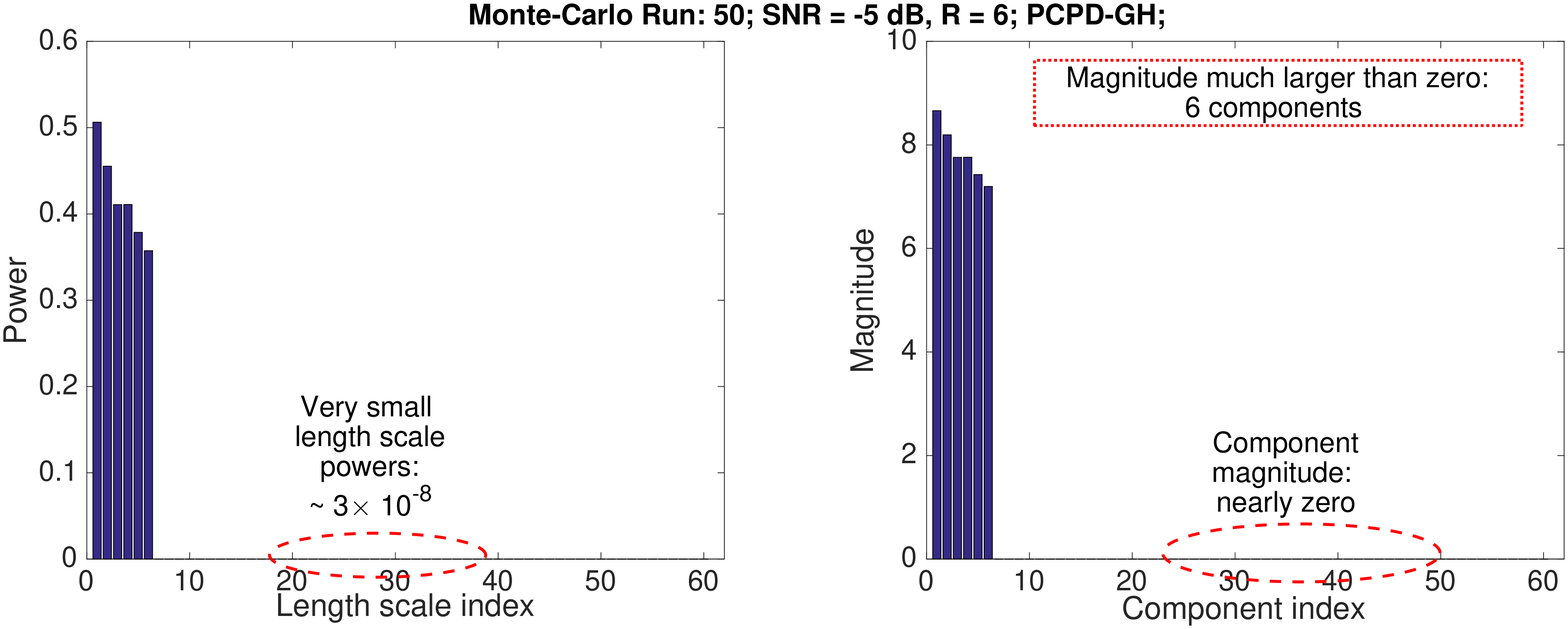}
}

\caption*{Figure P1:  (a)  The powers of learnt length scales (i.e., $\{ \gamma_l^{-1}\}_l$) and the magnitudes of  associated components for PCPD-GG;  (b) The powers of learnt length scales (i.e., $\{ z_l\}_l$)  and the magnitudes of  associated components for PCPD-GH.   It can be seen that PCPD-GG recovers 7 components with non-negligible magnitudes, while  PCPD-GH recovers 6 components. The two algorithms are with the same upper bound value: 60. SNR = -5 dB, R = 6; Monte-Carlo Run: 50.}
\end{figure*}

\begin{figure*} [!t]
\setcounter{subfigure}{0}
\centering
\subfigure[] {
\includegraphics[width=7  in]{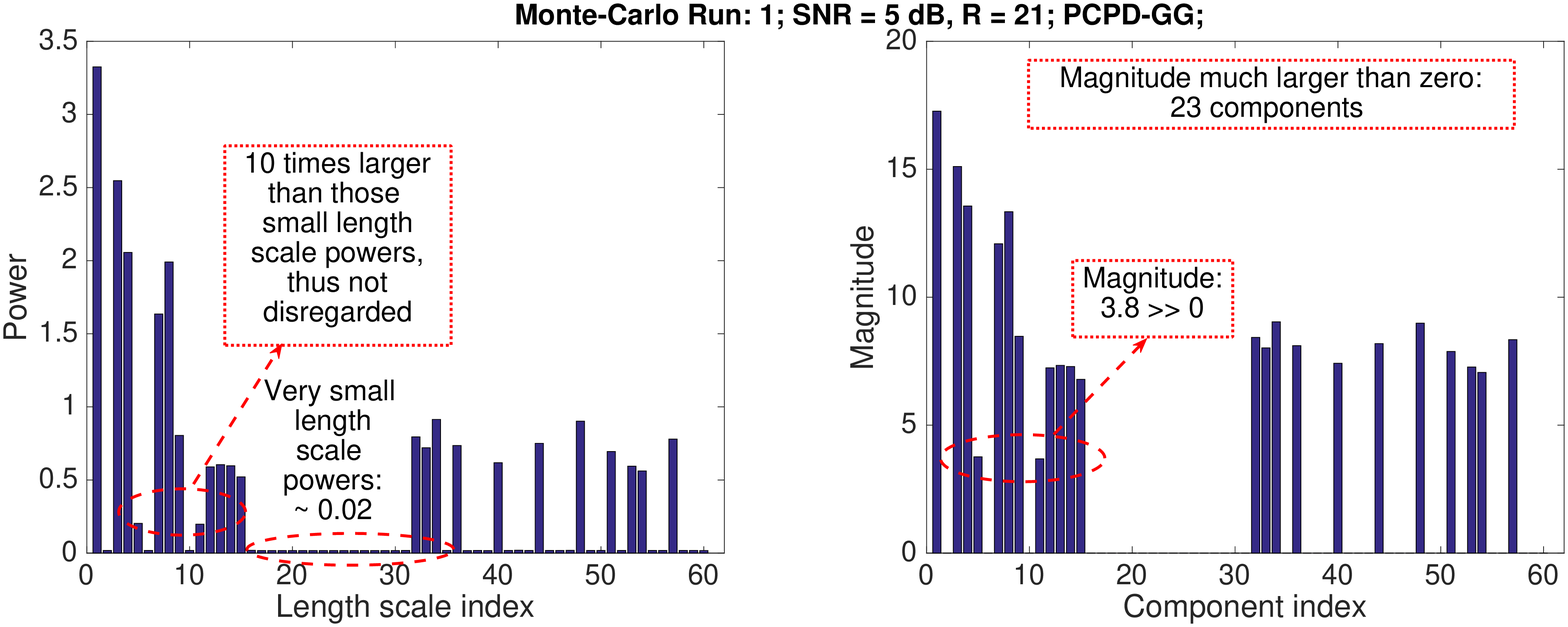}
}
\subfigure[] {
\includegraphics[width=7 in]{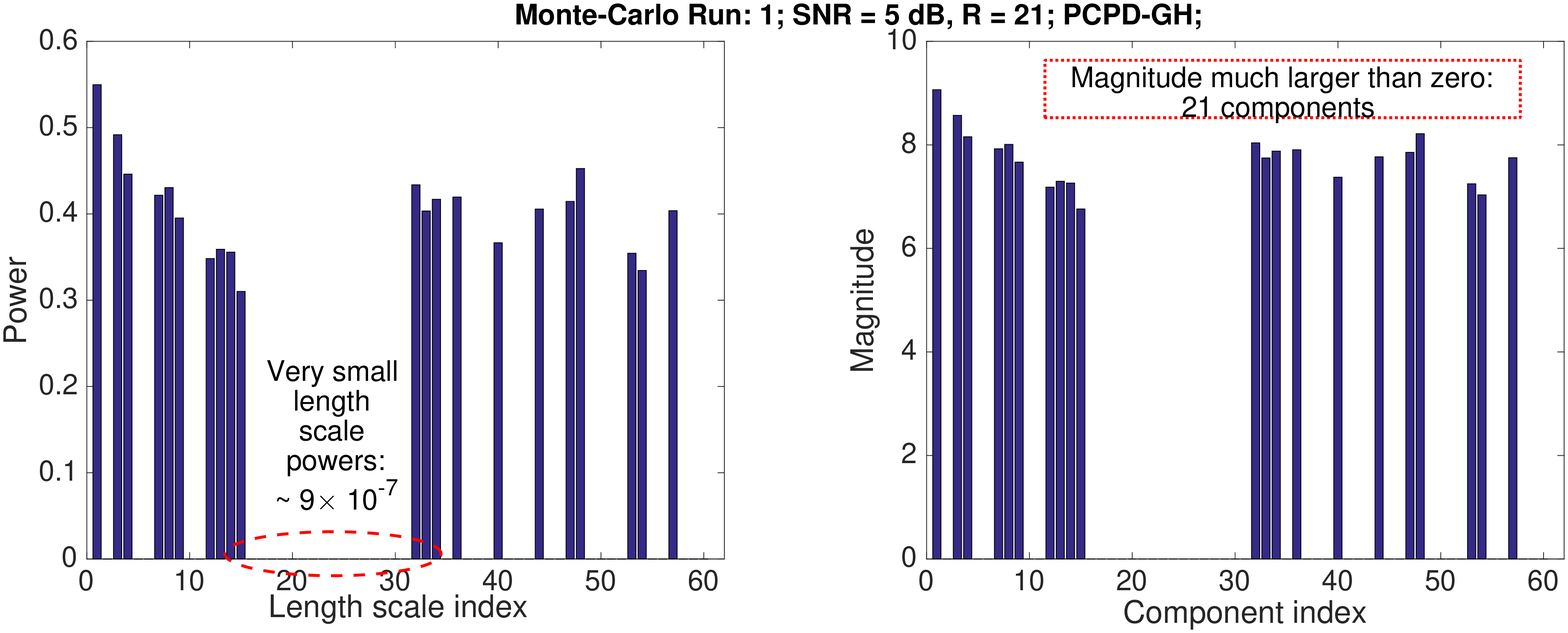}
}

\caption*{Figure P2: (a) The powers of learnt length scales (i.e., $\{ \gamma_l^{-1}\}_l$) and the magnitudes of  associated components for PCPD-GG;  (b) The powers of learnt length scales (i.e., $\{ z_l\}_l$)  and the magnitudes of    associated components for PCPD-GH.   It can be seen that  PCPD-GG recovers 23 components with non-negligible magnitudes, while  PCPD-GH recovers 21 components. The two algorithms are with the same upper bound value: 60. SNR = 5 dB, R = 21; Monte-Carlo Run: 1.}
\end{figure*}

\begin{figure*} [!t]
\setcounter{subfigure}{0}
\centering
\subfigure[] {
\includegraphics[width=7  in]{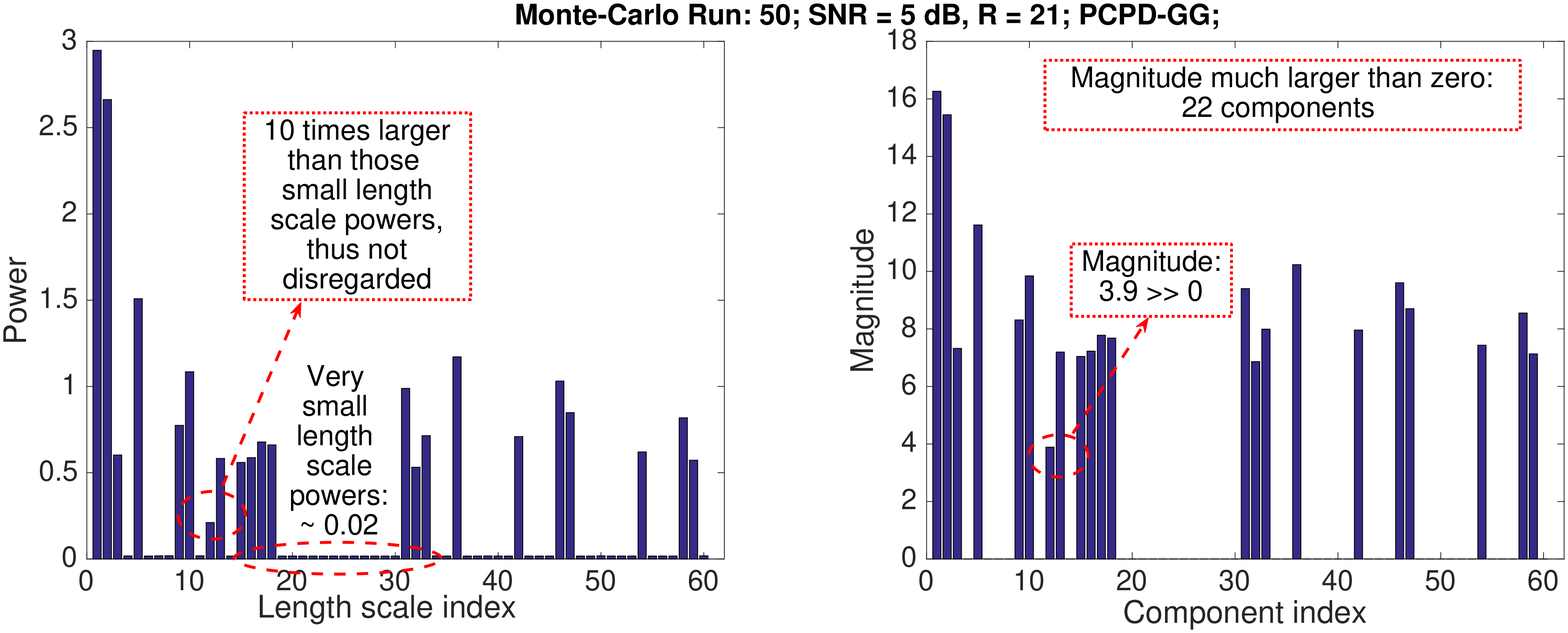}
}
\subfigure[] {
\includegraphics[width=7 in]{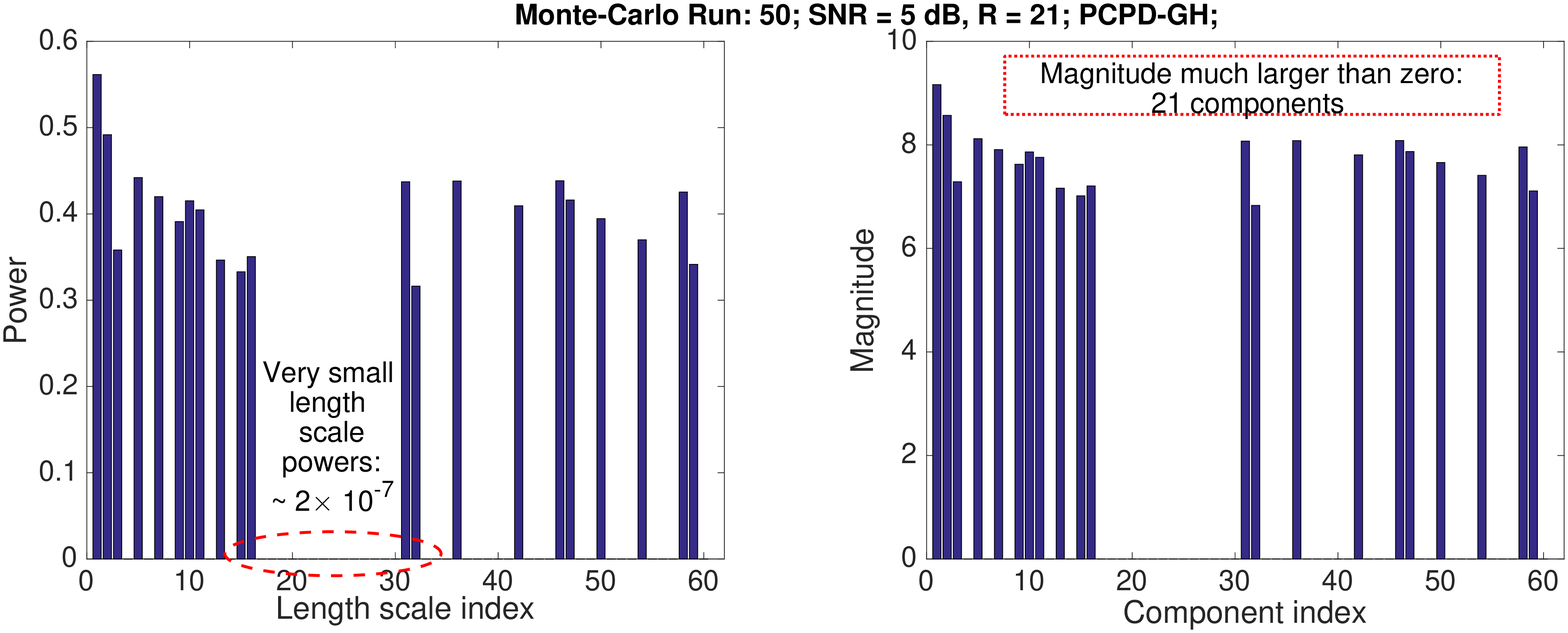}
}

\caption*{Figure P3:  (a)  The powers of learnt length scales (i.e., $\{ \gamma_l^{-1}\}_l$) and the magnitudes of  associated components for PCPD-GG;  (b) The powers of learnt length scales (i.e., $\{ z_l\}_l$)  and the magnitudes of  associated components for PCPD-GH.   It can be seen that PCPD-GG recovers 22 components with non-negligible magnitudes, while PCPD-GH recovers 21 components. The two algorithms are with the same upper bound value: 60. SNR = 5 dB, R = 21; Monte-Carlo Run: 50.}
\end{figure*}

\section{The effect of noise precision learning}

\begin{figure*} [!t]
\setcounter{subfigure}{0}
\centering
\subfigure[] {
\includegraphics[width=7  in]{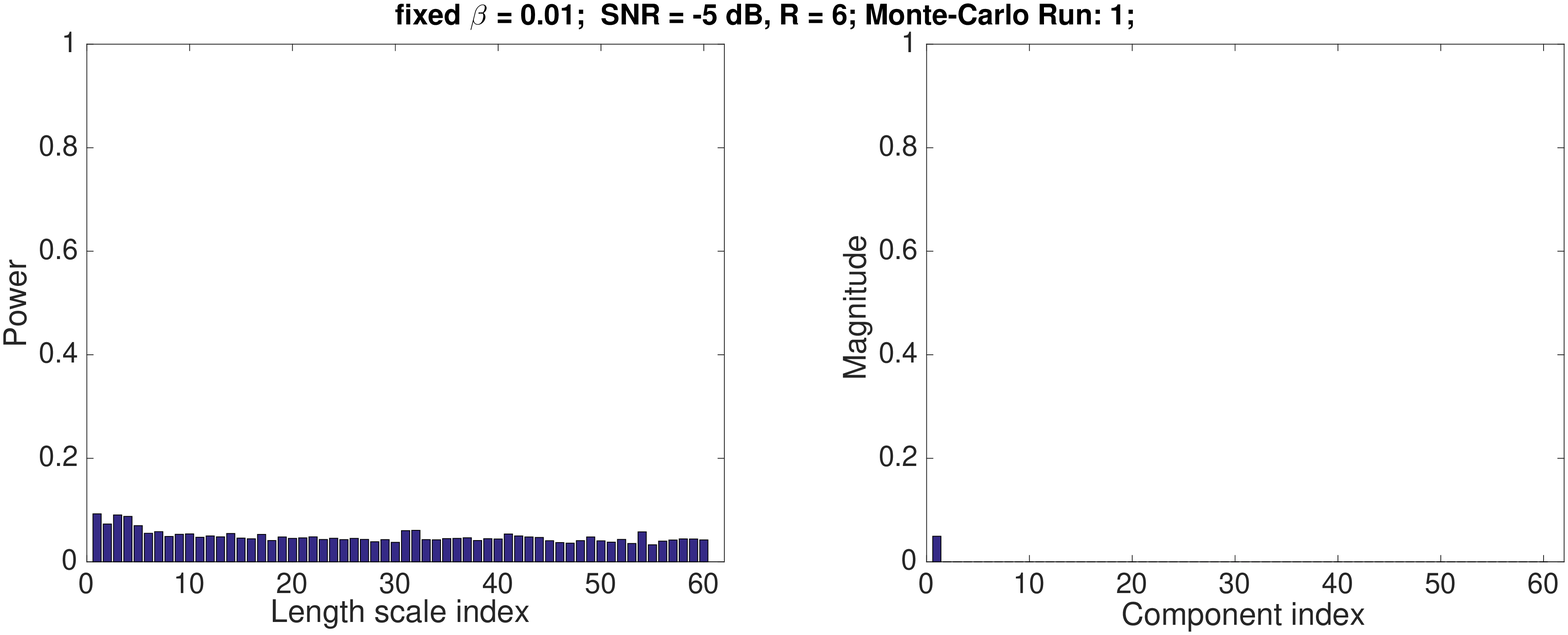}
}
\subfigure[] {
\includegraphics[width=7 in]{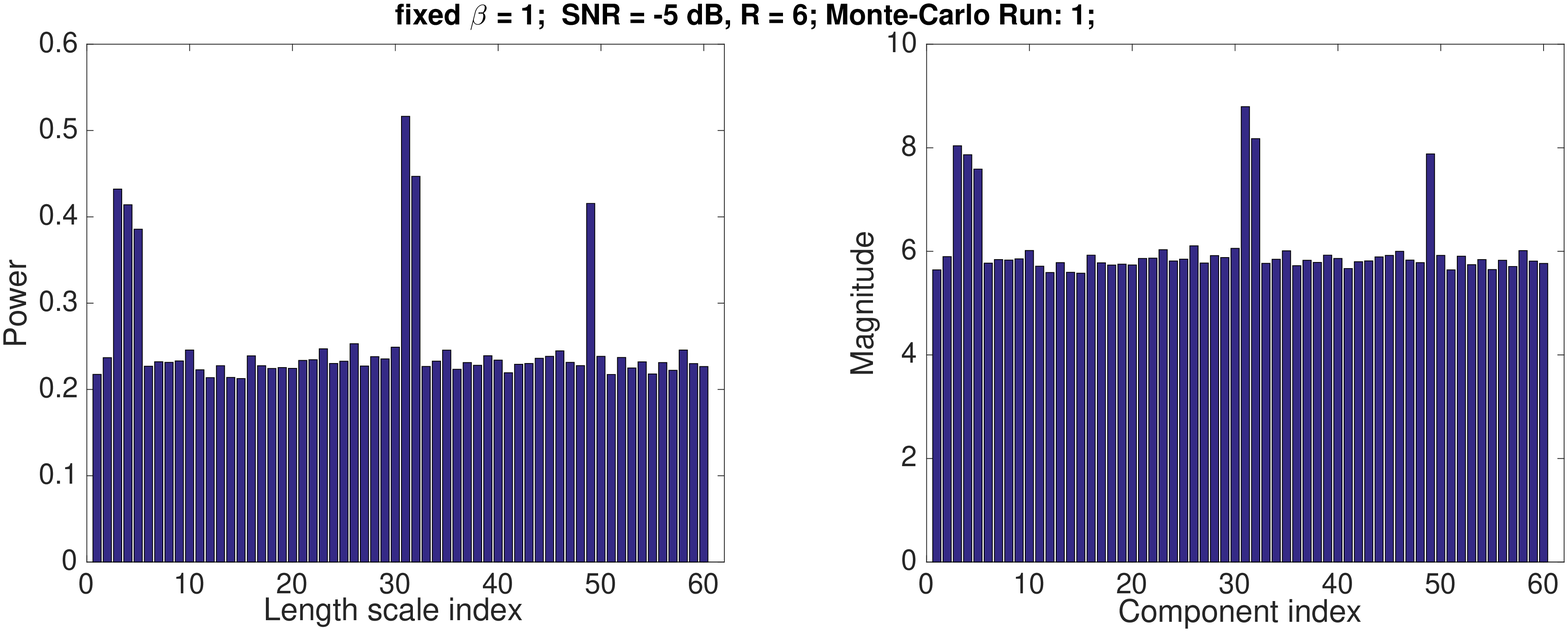}
}
\subfigure[] {
\includegraphics[width=7 in]{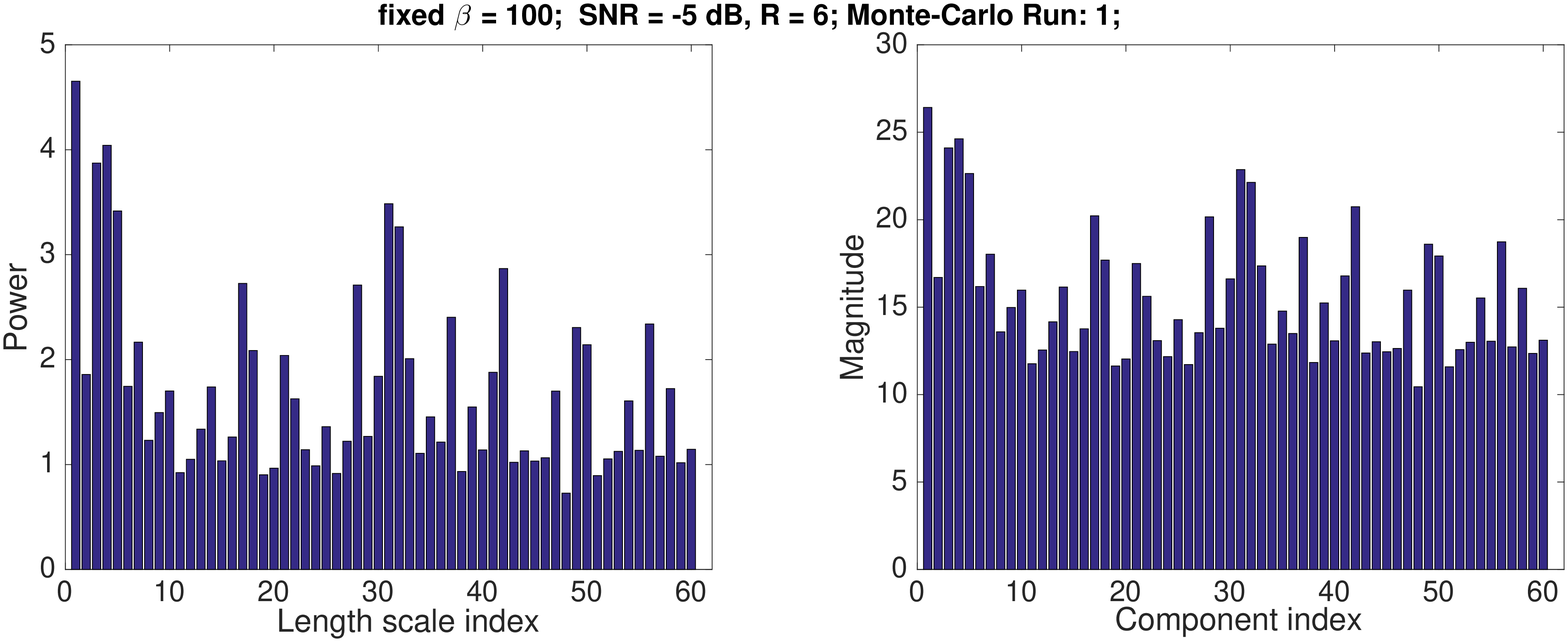}
}
\caption*{Figure Q1: Learnt length scale powers and the associated component magnitudes  under a fixed noise precision $\beta$. (a). $\beta = 0.01$;   (b). $\beta = 1$;  (c). $\beta = 100$. Algorithm: PCPD-GH; SNR = -5 dB; R = 6; Monte-Carlo Run: 1. }
\end{figure*}

To see the effect of noise precision learning, we turn off the noise learning process and set noise precision $\beta$ to fixed values $\{0.01, 1, 100 \}$.   The learnt length scales (without pruning) of  PCPD-GH are shown in Figure Q1 (under low SNR). Particularly, a small value of $\beta$ (e.g., 0.01) leads to over-regularization, thus causing under-estimation of non-zero components; a large value of $\beta$  (e.g., 100) causes under-regularization, thus inducing over-estimation of non-zero components.

\section{Optimizing hyper-parameter of PCPD-GG}
In PCPD-GG \cite{PI1},  since the parameters controlling the shape of the posterior $Q(\lambda_l)$ is updated by the tensor data during the learning process (see Eq. (25) in  \cite{PI1}), further optimizations of their hyper-parameters will not help too much when the tensor data is large. To verify this, following the notations in  \cite{PI1},  we impose a hyper-prior for the parameter $d^0_r$ as $p(d^0_r) = \mathrm{gamma}(d^0_r | \epsilon, \epsilon )$, where $\epsilon = 10^{-6}$ indicates the non-informative nature. Since this prior satisfies the conjugacy property, it is straightforward to derive the optimal variational pdf $Q(d^0_r) = \mathrm{gamma} (d^0_r |  c^0_r + \epsilon, \lambda_r + \epsilon) $, and use this equation to update $d^0_r$. We name this variant as PCPD-GG-HO. In Figure R1, we turn off the pruning and show the learnt length scales in low SNR regime. It is seen that incorporating the hyper-parameter optimization does not mitigate the redundant component with non-negligible magnitude. Then, we turn on the pruning procedure, and it shows that in 100 Monte-Carlo runs,  PCPD-GG-HO method only correctly estimates the tensor rank in 10 runs (i.e., accuracy 10\%), which is similar to that of PCPD-GG. 

\begin{figure*}[!t]
\includegraphics[width= 6.5in]{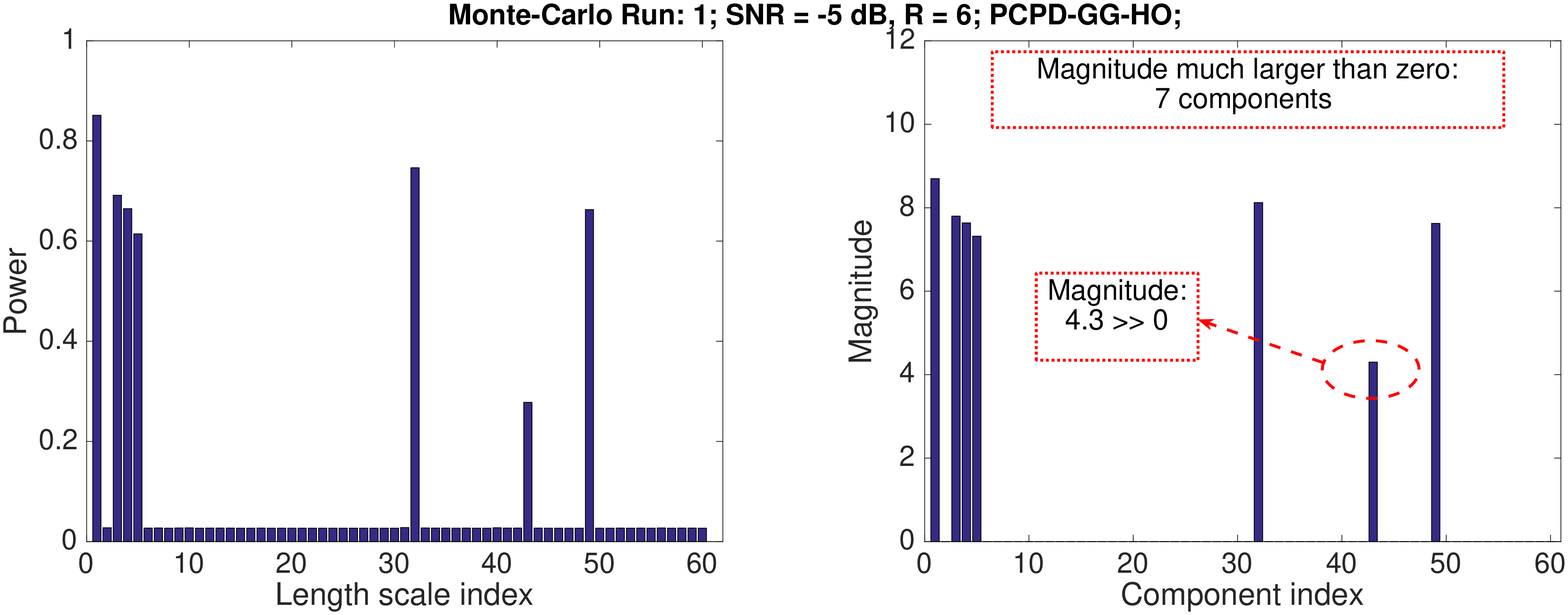}
\caption*{Figure R1.  Synthetic data analysis for PCPD-GG-HO. It shows the powers of learnt length scales (i.e., $\{ \gamma_l^{-1}\}_l$) and the magnitudes of associated components. It can be seen that  PCPD-GG-HO recovers 7 components with non-negligible magnitudes. SNR = -5 dB, R = 6; Monte-Carlo Run: 1.}
\end{figure*}

\section{Learnt length scales of PCPD-MGP}
In Figure S1-S5, the learnt length scales of PCPD-MGP and PCPD-GH under different cases are presented. Discussions on these results are in Section V. C..

\begin{figure*} [!t]
\setcounter{subfigure}{0}
\centering
\subfigure[] {
\includegraphics[width=7  in]{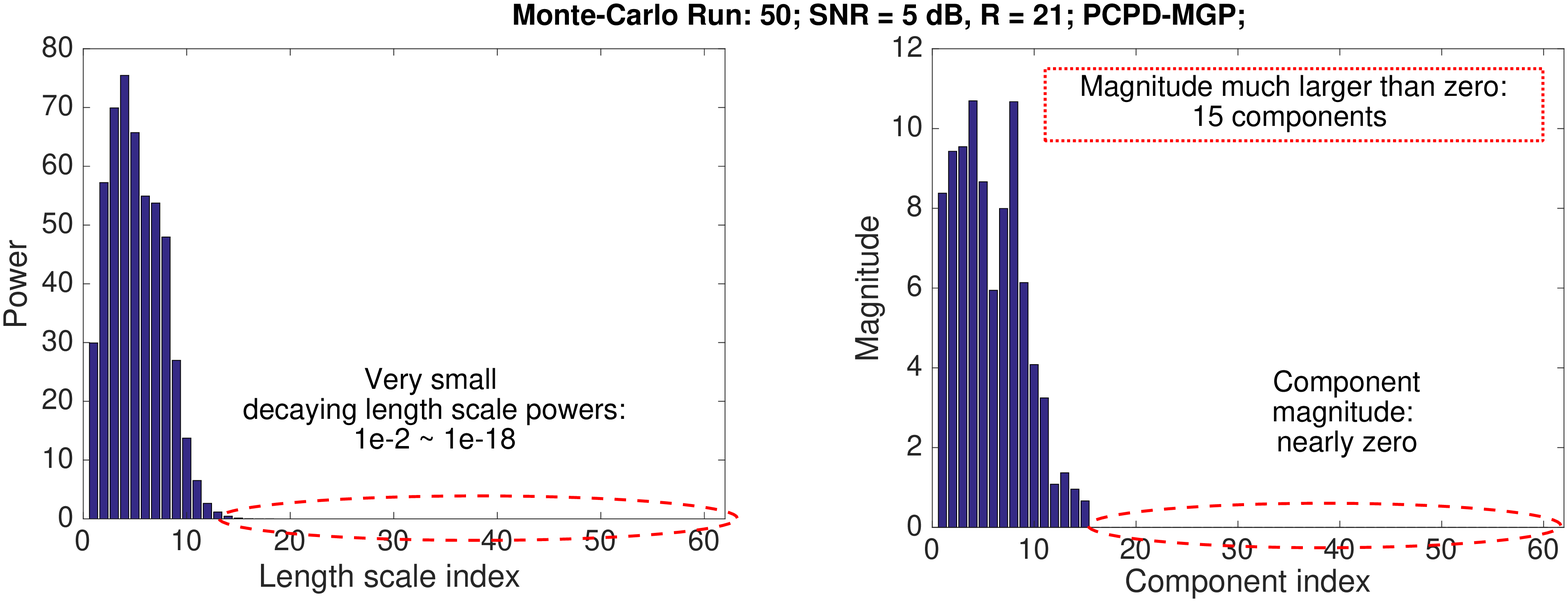}
}
\subfigure[] {
\includegraphics[width=7 in]{mont_50_gh_21.eps}
}
\caption*{Figure S1:   (a) The powers of learnt length scales (i.e., $\{ \tau_l^{-1}\}_l$) and the magnitudes of associated components for PCPD-MGP;  (b) The powers  of learnt length scales (i.e., $\{ z_l\}_l$)  and the magnitudes of  associated components for PCPD-GH.  It can be seen that  PCPD-MGP recovers 15 components with non-negligible magnitudes, while  PCPD-GH recovers 21 components. The two algorithms are with the same upper bound value: 60.  SNR = 5 dB, R = 21; Monte-Carlo Run: 50. }
\end{figure*}

\begin{figure*} [!t]
\setcounter{subfigure}{0}
\centering
\subfigure[] {
\includegraphics[width=7  in]{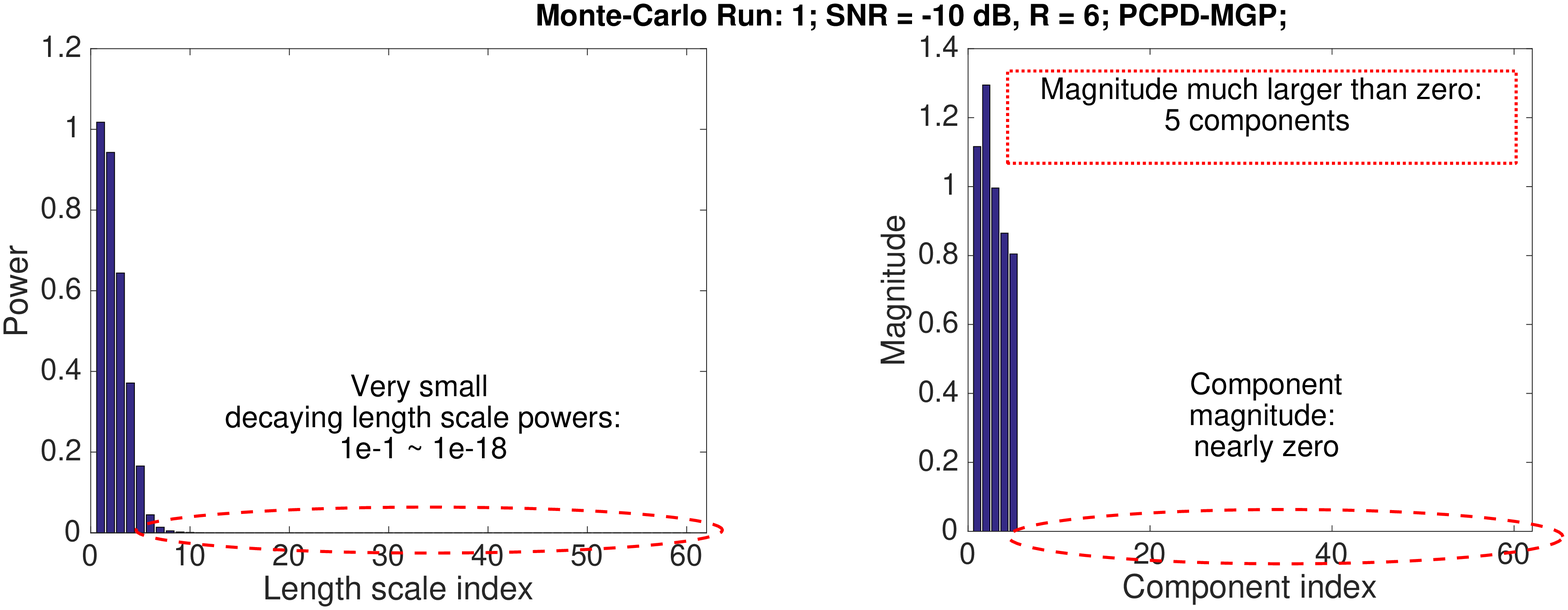}
}
\subfigure[] {
\includegraphics[width=7 in]{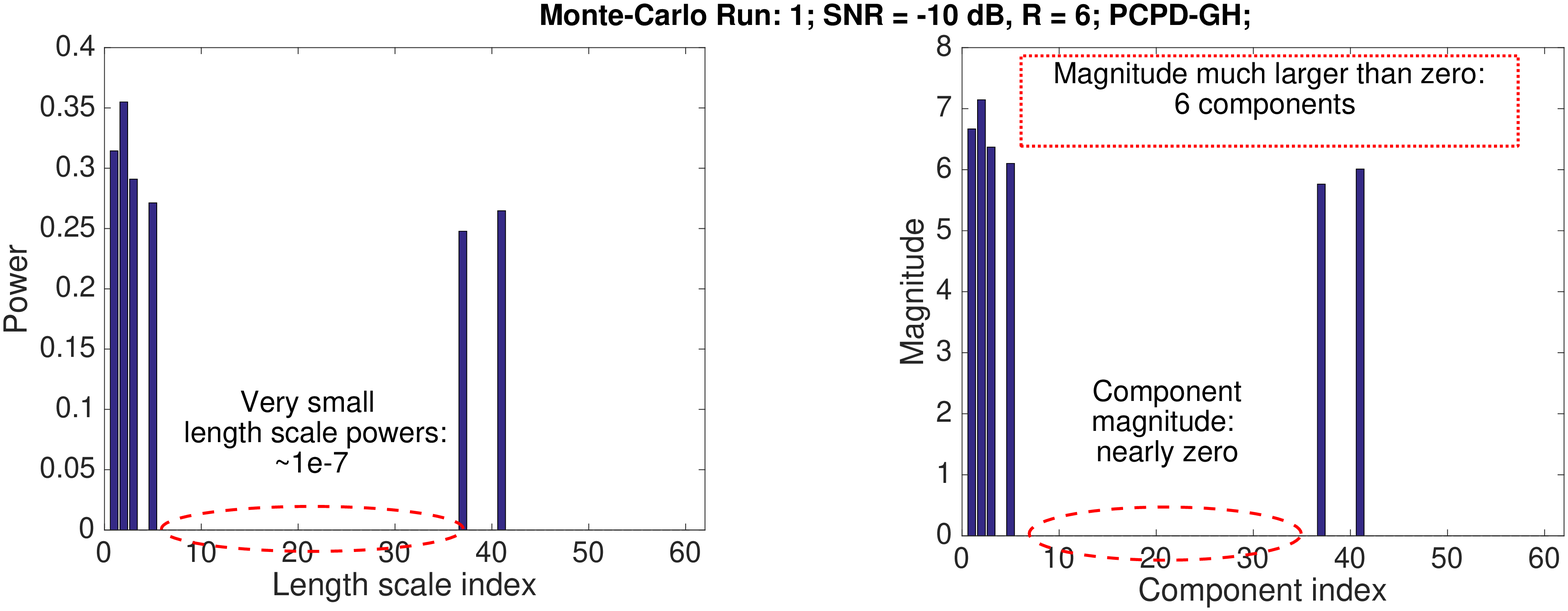}
}
\caption*{Figure S2:   (a) The powers of learnt length scales (i.e., $\{ \tau_l^{-1}\}_l$) and the magnitudes of associated components for PCPD-MGP;  (b) The powers  of learnt length scales (i.e., $\{ z_l\}_l$)  and the magnitudes of  associated components for PCPD-GH.   It can be seen that  PCPD-MGP recovers 5 components with non-negligible magnitudes, while PCPD-GH recovers 6 components. The two algorithms are with the same upper bound value: 60.  SNR = -10 dB, R = 6; Monte-Carlo Run: 1.  }
\end{figure*}

\begin{figure*} [!t]
\setcounter{subfigure}{0}
\centering
\subfigure[] {
\includegraphics[width=7  in]{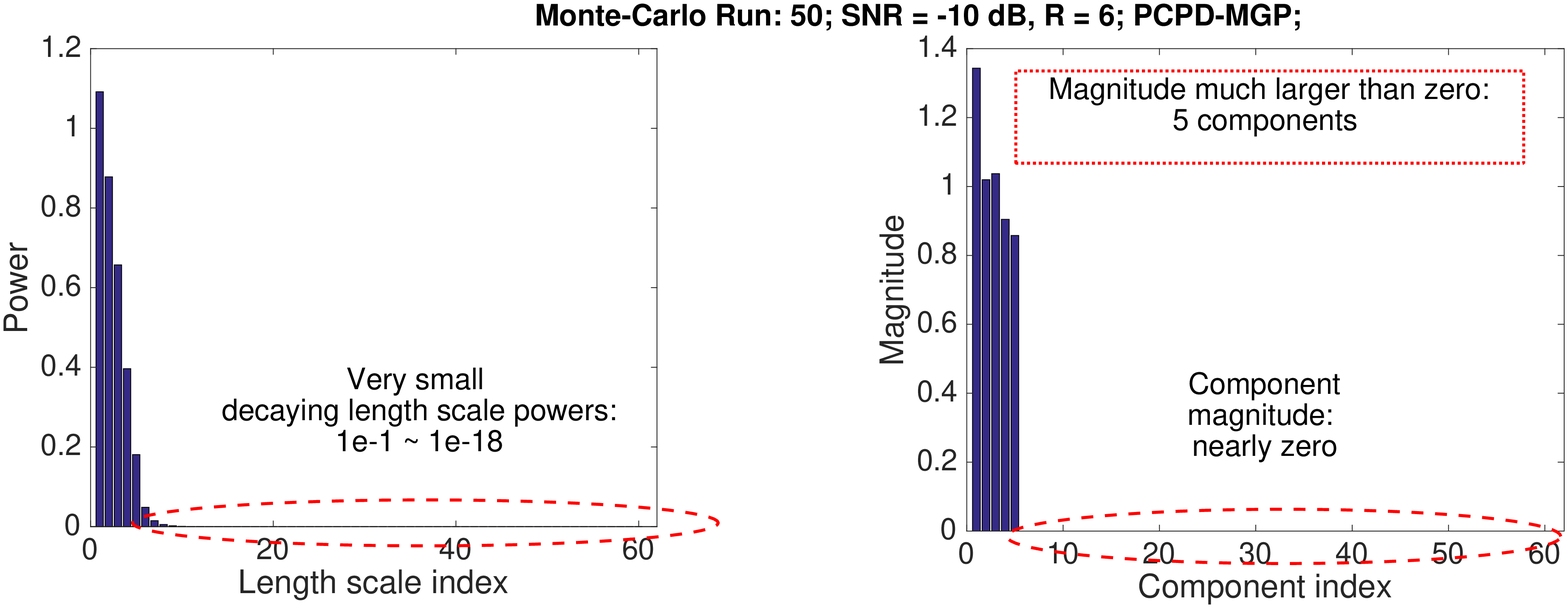}
}
\subfigure[] {
\includegraphics[width=7 in]{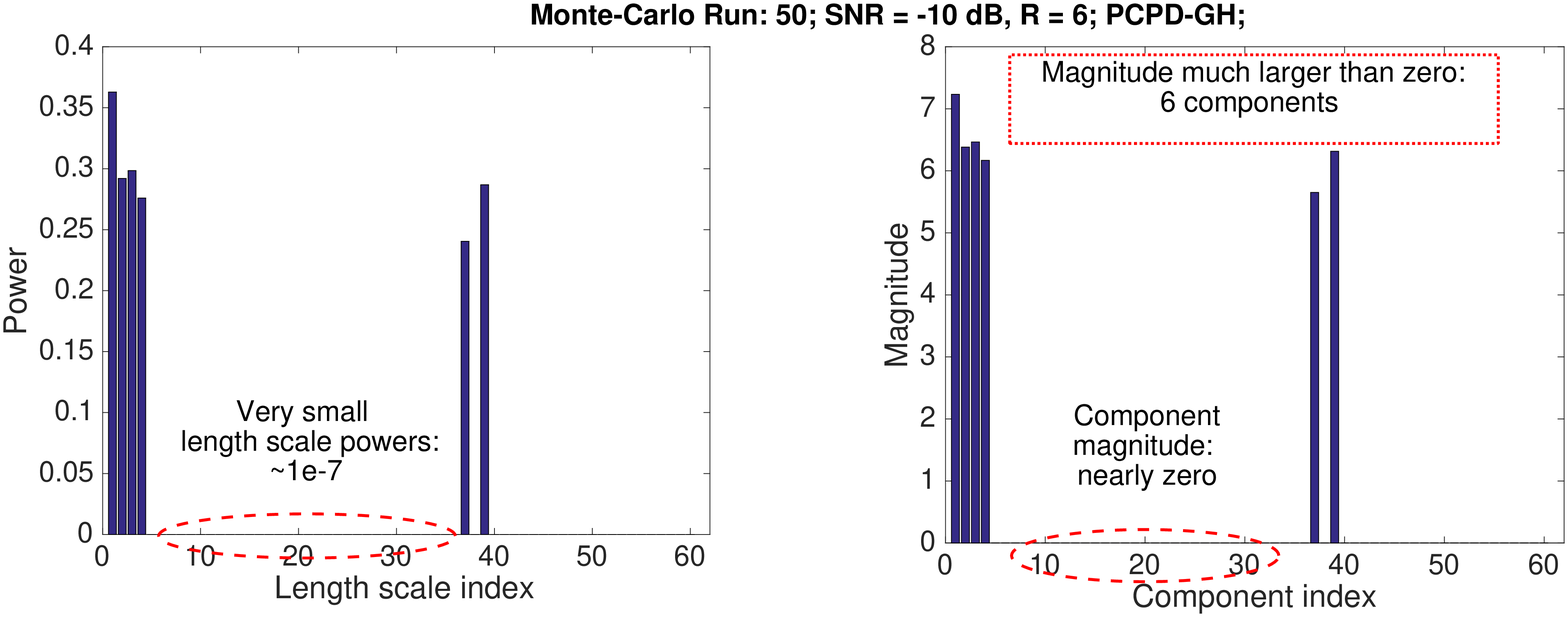}
}
\caption*{Figure S3:    (a) The powers of learnt length scales (i.e., $\{ \tau_l^{-1}\}_l$) and the magnitudes of associated components for PCPD-MGP;  (b) The powers  of learnt length scales (i.e., $\{ z_l\}_l$)  and the magnitudes of  associated components for PCPD-GH.   It can be seen that PCPD-MGP recovers 5 components with non-negligible magnitudes, while  PCPD-GH recovers 6 components. The two algorithms are with the same upper bound value: 60.  SNR = -10 dB, R = 6; Monte-Carlo Run: 50. }
\end{figure*}

\begin{figure*} [!t]
\setcounter{subfigure}{0}
\centering
\subfigure[] {
\includegraphics[width=7  in]{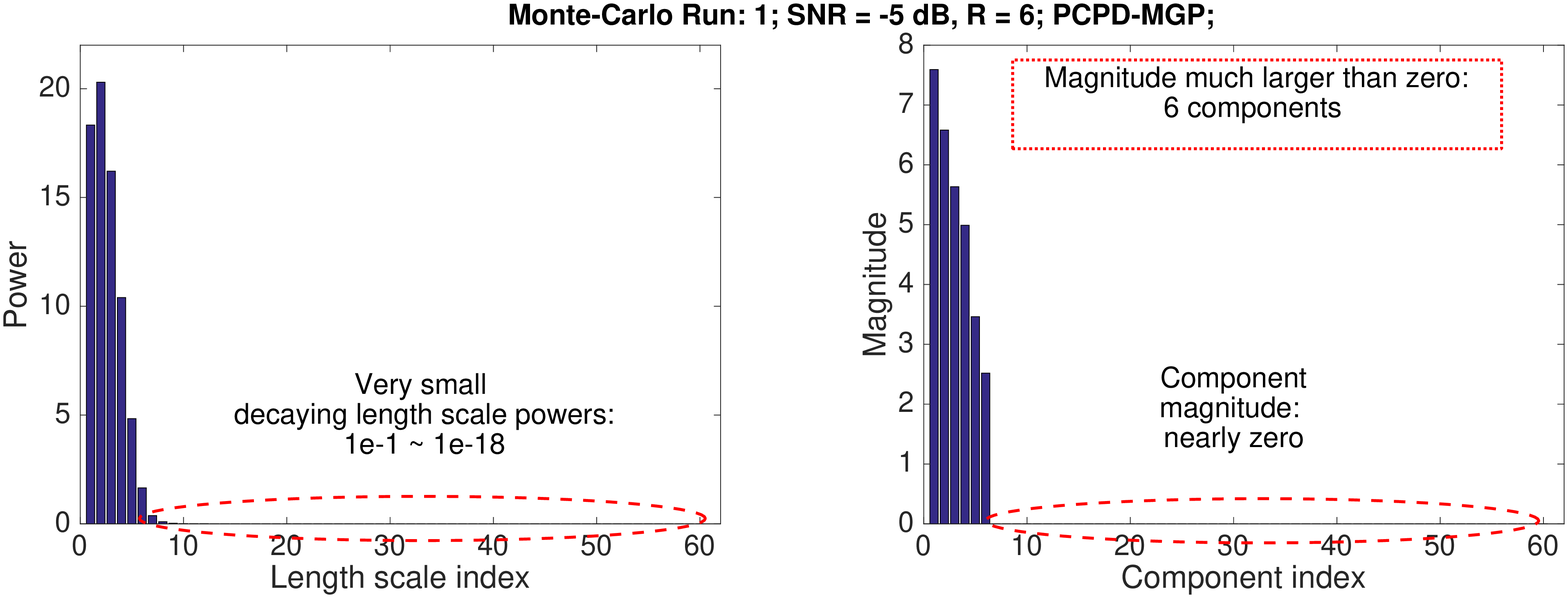}
}
\subfigure[] {
\includegraphics[width=7 in]{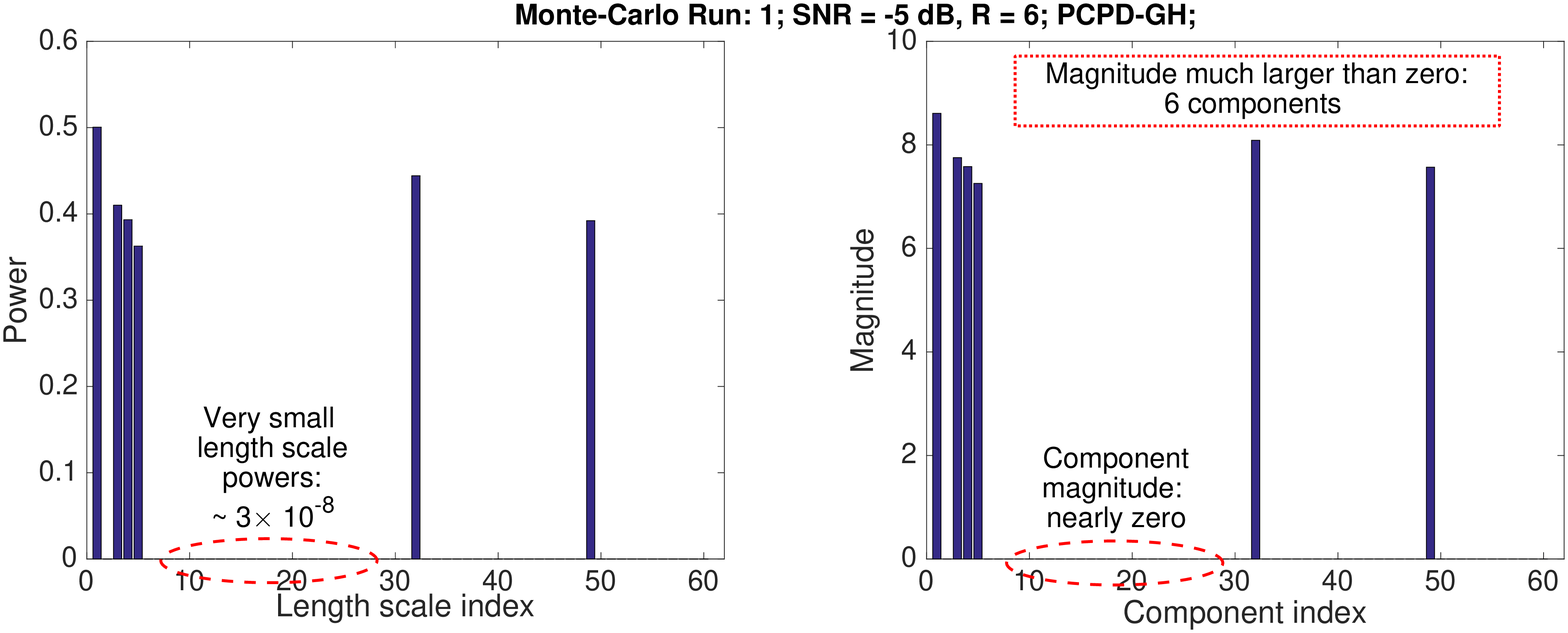}
}
\caption*{Figure S4:   (a) The powers of learnt length scales (i.e., $\{ \tau_l^{-1}\}_l$) and the magnitudes of associated components for PCPD-MGP;  (b) The powers  of learnt length scales (i.e., $\{ z_l\}_l$)  and the magnitudes of  associated components for PCPD-GH.  It can be seen that PCPD-MGP and PCPD-GH both recover 6 components with non-negligible magnitudes. The two algorithms are with the same upper bound value: 60.  SNR = -5 dB, R = 6; Monte-Carlo Run: 1. }
\end{figure*}

\begin{figure*} [!t]
\setcounter{subfigure}{0}
\centering
\subfigure[] {
\includegraphics[width=7  in]{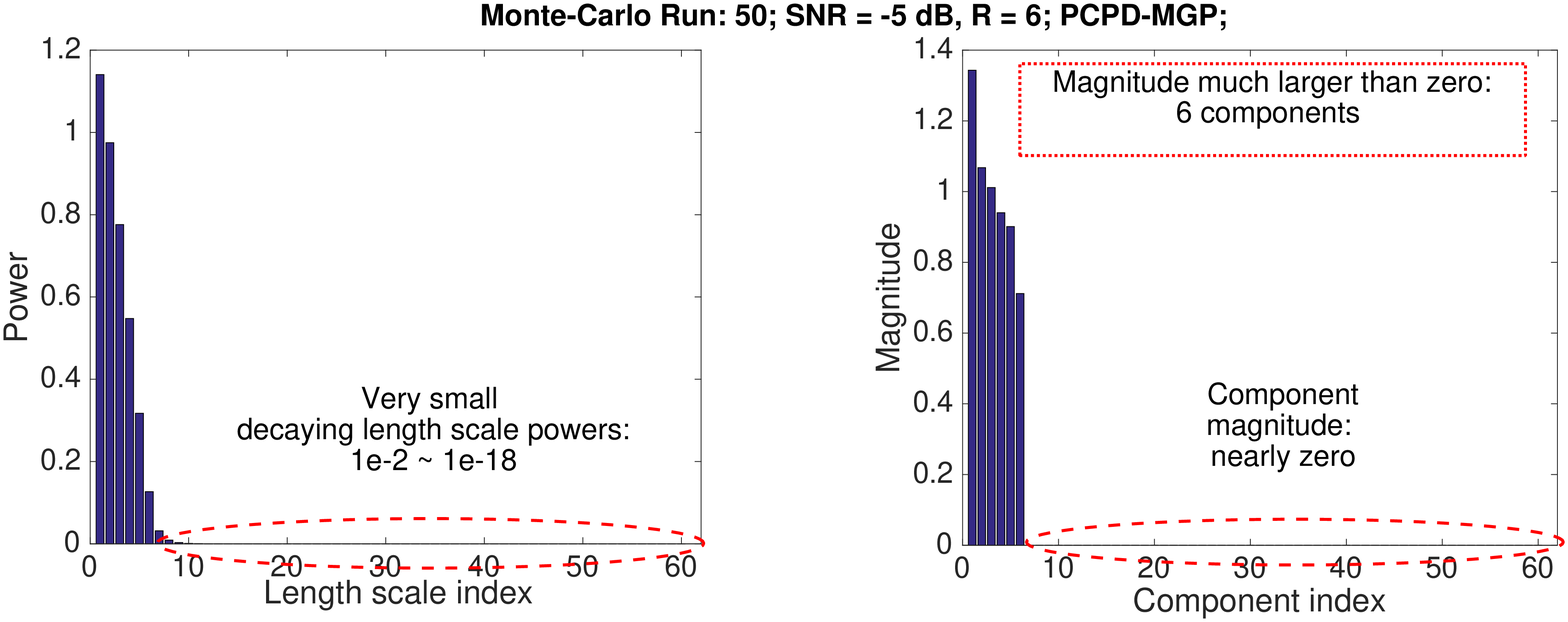}
}
\subfigure[] {
\includegraphics[width=7 in]{mont_50_gh_6.eps}
}
\caption*{Figure S5: (a) The powers of learnt length scales (i.e., $\{ \tau_l^{-1}\}_l$) and the magnitudes of associated components for PCPD-MGP;  (b) The powers  of learnt length scales (i.e., $\{ z_l\}_l$)  and the magnitudes of  associated components for PCPD-GH.   It can be seen that PCPD-MGP and PCPD-GH both recover 6 components with non-negligible magnitudes. The two algorithms are with the same upper bound value: 60.  SNR = -5 dB, R = 6; Monte-Carlo Run: 50. }
\end{figure*}

\end{document}